\documentclass[Acad.AI.,reviewarticle,9pt,twoside,twocolumn,moreauthors]{Definitions/jams_academia}
\usepackage{epstopdf}
\usepackage{placeins}

\usepackage{tikz}
\usetikzlibrary{positioning, arrows.meta, shapes.geometric, calc,shadows,backgrounds,fit}
\usepackage[numbers,square,sort&compress]{natbib}
\usepackage{forest}

\usepackage{xcolor}
\definecolor{lightcoral}{rgb}{0.94, 0.5, 0.5}
\definecolor{lightgreen}{rgb}{0.56, 0.93, 0.56}
\definecolor{lightblue}{rgb}{0.9, 0.9, 1}

\definecolor{lightyellow}{rgb}{0.94, 0.84, 0.6}
\definecolor{harvestgold}{rgb}{0.85, 0.57, 0.0}
\definecolor{brightlavender}{rgb}{0.75, 0.58, 0.89}
\definecolor{capri}{rgb}{0.0, 0.75, 1.0}
\definecolor{carminepink}{rgb}{0.92, 0.3, 0.26}
\definecolor{celadon}{rgb}{0.67, 0.88, 0.69}
\definecolor{darkpastelgreen}{rgb}{0.01, 0.75, 0.24}

\definecolor{plum}{RGB}{221, 160, 221}

\usepackage{tabularx,array}
\newcolumntype{L}{>{\raggedright\arraybackslash}X}

\firstpage{1} 
\pubvolume{2}
\issuenum{1}
\pubyear{2026}
\copyrightyear{2026}
\externaleditor{Academic Editors: Vincent Hilaire, Soo Ling Lim} 
\datereceived{1 February 2026} 
\dateaccepted{9 April 2026} 
\datepublished{Day Month 2026} 
\doinum{10.20935/xxx}
\Title{Towards trustworthy agentic AI: a comprehensive survey of safety, robustness, privacy, and system security}

\TitleCitation{Towards trustworthy agentic AI: a comprehensive survey on safety, robustness, privacy, and system security}

\Author{Jinhu
 Qi$^{1}$, Muzhi Li$^{1}$, Jiahong Liu$^{1}$, Yuqin Shu$^{1}$, Dianzhi Yu$^{1}$, Shicheng Ma$^{1}$, Wenqian Cui$^{1}$, Yiyang Zhao$^{2,3}$, Yiyi Chen$^{2}$,\\ Ruoxi Jiang$^{2,3}$, Irwin King$^{1,}$*, Zenglin Xu$^{2,3,}$*}

\AuthorNames{Qi J, Li M, Liu J, Shu Y, Yu D, Ma S, et al}

\AuthorCitation{Qi J, Li M, Liu J, Shu Y, Yu D, Ma S, et al}

\address{%
\textsuperscript{1}{Faculty of Engineering, Department
of Computer Science and Engineering, The Chinese University of Hong Kong, Hong Kong, China.
}\\
\textsuperscript{2}{Artificial
Intelligence Innovation and Incubation Institute, Fudan University, Shanghai, China.}\\
\textsuperscript{3}{Shanghai Academy of AI for Science
, Shanghai, China.}
}

\corres{king@cse.cuhk.edu.hk (I.K.); zenglinxu@fudan.edu.cn (Z.X.)}

\abstract{Agentic AI systems---Large Language Models (LLMs) augmented with planning, tool use, memory, and long-horizon interactions---can execute complex tasks autonomously, but their multi-step trajectories introduce new failure modes that challenge trustworthiness. This survey provides a focused examination of trustworthy agentic AI through two core dimensions that are critical for high-risk deployments: Safety and Robustness and Privacy and System Security. For each dimension, we clarify key concepts, identify where risks emerge along the agent workflow, and summarize stage-targeted mitigation strategies. Other trustworthiness aspects (value alignment, transparency, fairness, and accountability) are discussed as relevant context rather than parallel chapters. To support consistent comparison and deployment decisions, we consolidate evaluation into a unified metrics-and-benchmarks hub, emphasizing both outcome and process signals (e.g., constraint violations, trace completeness, and adversarial success rates) and offering scenario-to-metric guidance for release gating. We conclude by outlining open challenges such as self-evolving agents, runtime monitoring and verification, privacy-preserving personalization, and the trust--utility trade-off, and present a case study of real-world security failures in open-source agentic systems (OpenClaw/Moltbook). Our goal is to serve as a practical reference for researchers and practitioners building trustworthy agentic systems in high-stakes environments.}

\begin{document}


\newcolumntype{C}[1]{>{\centering\arraybackslash}p{#1}}

\section{Introduction}\label{section:introduction}

\subsection{Motivation} The paradigm shift from static Large Language Models (LLMs) to agentic systems---capable of autonomous planning, tool invocation, and multi-step reasoning---has enabled their deployment in critical real-world applications. From automating complex software development cycles to serving as intelligent intermediaries in healthcare and financial services, these agents leverage their ability to interact with external environments to fulfill high-level goals \cite{xi2023rise,chen-etal-2025-finhear,chen2025heterogeneousgroupbasedreinforcementlearning}. This increasing autonomy transforms them from mere productivity aids into central nodes of modern digital infrastructure.

However, as LLM-based assistants are increasingly connected to enterprise data and tools, failures can translate into direct real-world impact. 
For example, a ``zero-click'' prompt injection vulnerability in Microsoft 365 Copilot (CVE-2025-32711, ``EchoLeak'') was publicly reported and patched, highlighting that specially crafted untrusted inputs (e.g., emails) may trigger unintended behaviors and enable sensitive data exposure without explicit user interaction~\cite{Lakshmanan2025EchoLeak,Paverd2025IndirectPI}.
More broadly, prior work has shown that indirect
 prompt injection blurs the boundary between data and instructions in LLM-integrated applications, allowing attacker-controlled content retrieved from the web or documents\newpage
\begin{itemize}
\item[]\vspace*{-17pt}
\end{itemize}
 to hijack tool-using systems and cause data exfiltration or unintended actions~\cite{Greshake2023IndirectPI,OWASPPromptInjection,10.1145/3701716.3715521}.
These incidents underscore that ``trustworthiness'' for agentic AI must be assessed at the \emph{system level},
beyond single-turn outputs.

Against this backdrop, Large Language Models (LLMs) have rapidly progressed from pure text generators to systems that can \emph{act}
in the world.
Modern ``agentic AI'' augments LLMs with planning, tool use (e.g., web browsing, APIs, and code execution), memory, and long-horizon
interactions, enabling them to decompose complex goals into executable steps and iteratively refine behavior from feedback~\cite{li2025evidencetrajectoryabductivereasoning}.
Representative systems demonstrate that such agents can autonomously explore environments and continuously acquire skills
(e.g., via lifelong interaction and self-improvement loops)~\cite{Wang2023Voyager,packer2023memgpt,Zhang2025Agentic}.
This shift from static single-turn models to autonomous or semi-autonomous agents is a capability leap---but it also introduces
qualitatively new risks.

Unlike traditional predictive models or chat-based LLMs, agentic systems produce multi-step trajectories whose intermediate states (plans, tool calls, retrieved evidence, and memory updates) can directly affect real-world outcomes.
A minor error early in a {\addfontfeature{LetterSpace=1.2}trajectory can cascade into high-impact actions, and the agent's}\newpage interaction with tools expands the attack
surface (prompt injection, tool misuse, and data exfiltration) beyond what is captured by conventional LLM safety evaluations.
Moreover, agents increasingly operate in settings where human oversight is intermittent rather than continuous; this raises
fundamental questions about the accountability, auditability, and interruptibility of agent behavior
\cite{accountability_overall_challenge1,accountability_overall_challenge2}.
As agent autonomy grows, ``trustworthiness'' must be assessed not only by final outputs, but also by process signals such as
constraint compliance, trace evidence, and robustness to adversarial and long-horizon stress.

At the same time, trustworthiness itself is not a single property.
It spans multiple dimensions that interact with one another: strengthening memory can improve effectiveness but may increase privacy
risk; adding safeguards can reduce catastrophic failures but may reduce utility or increase cost; and explanation interfaces can improve
auditability but may also inflate over-trust if explanations are unfaithful.
These tensions motivate a survey that treats trustworthy agentic AI as a system-level
 problem rather than a model-only
problem and that makes evaluation comparable across dimensions and deployment scenarios.

\subsection{Scope and perspective}
We focus on LLM-based agentic systems that (i) plan over extended horizons, (ii) use external tools and environments, (iii) may incorporate memory, self-reflection, or multi-agent interaction.
To structure the discussion, we adopt an agent workflow lens---Perceive $\rightarrow$ Plan $\rightarrow$ Act $\rightarrow$
Reflect $\rightarrow$ Learn---to pinpoint where risks arise and where mitigations intervene.
This workflow is not meant to be a strict architecture requirement; rather, it provides a consistent interface for mapping
threats, defenses, and evaluation signals across diverse agent designs.

\subsection{Relationship to chat-based system risks}
Many trust and safety concerns discussed in this survey---such as harmful content generation, deceptive outputs, or inappropriate advice---also arise in non-agentic, chat-based LLM systems.
We do not exclude these foundational risks; rather, we emphasize that agentic autonomy amplifies and extends them in qualitatively new ways.
For example, prompt injection attacks against a chat-based system may produce misleading text, but the same attack against a tool-using agent can trigger unauthorized code execution, data exfiltration, or irreversible real-world actions.
Similarly, harmful persuasion in a chat setting is bounded by the conversation, whereas an agent with tool access can autonomously act on such persuasion across a multi-step trajectory.
The risks and mitigations we survey are therefore not exclusive to agentic systems, but their severity, attack surface, and cascading potential are substantially greater when agents operate with real-world affordances~\cite{Gutfraind2023RiskUncertaintyAI}.
Where applicable, we note when a risk or method also applies to non-agentic LLM deployments.

\subsection{Illustrative rather than exhaustive scope}
Given the rapidly evolving nature of agentic AI, the risks and mitigation methods discussed in this survey should be understood as illustrative examples reflecting the current state of knowledge, rather than a provably exhaustive enumeration.
New risks and solution methods are likely to emerge as agent architectures, tool ecosystems, and deployment contexts continue to evolve.
Where possible, we note the boundaries of current understanding and highlight directions where coverage remains incomplete.

\subsection{Comparison with existing surveys}
Prior surveys have investigated trustworthy AI principles and requirements for general AI systems
(e.g.,~\cite{Kaur2022TrustworthyAIReview})
and trustworthiness evaluation for Large Language Models (e.g.,~\cite{Liu2023TrustworthyLLMs,Huang2024TrustLLM}).
Recent works start to address trust/safety issues for LLM-based agents and multi-agent systems
(e.g.,~\cite{Yu2025TrustworthyAgents,Ma2025SafetyAtScale})
or provide architecture/application-centered overviews of agentic AI (e.g.,~\cite{AbouAli2025AgenticAI,Xu2025ToolLearningSurvey}).
In contrast, this survey integrates a multi-dimensional trust taxonomy with a workflow lens,
and further consolidates process-aware evaluation and scenario-based release gating
(\textbf{Table~\ref{tab:survey-diff}}).

\begin{table}[H]
\footnotesize
\caption{Comparison with representative related surveys on trustworthy AI/large language models/agents.\vspace{4pt}}

\label{tab:survey-diff}

\setlength{\tabcolsep}{5pt}
\renewcommand{\arraystretch}{1.45}
\rowcolors{1}{white}{white}  
\begin{tabularx}{\linewidth}{@{}
    |>{\hsize=1.55\hsize\raggedright\arraybackslash}X
    |>{\hsize=2.65\hsize\raggedright\arraybackslash}X
    |>{\hsize=0.63\hsize\centering\arraybackslash}X
    |>{\hsize=0.39\hsize\centering\arraybackslash}X
    |>{\hsize=0.39\hsize\centering\arraybackslash}X
    |>{\hsize=0.39\hsize\centering\arraybackslash}X|
@{}}
\hline
\textbf{Survey} & \textbf{Scope} & \textbf{MDT} & \textbf{WL} & \textbf{EH} & \textbf{RG} \\
\hline

\cite{Kaur2022TrustworthyAIReview} 
& General TAI
& $\checkmark$ & $-$ & $\sim$ & $-$ \\
\hline
\cite{Liu2023TrustworthyLLMs}
& Trustworthy LLMs
& $\checkmark$ & $-$ & $\sim$ & $-$ \\
\hline
\cite{Huang2024TrustLLM}
& LLM benchmark
& $\checkmark$ & $-$ & $\checkmark$ & $-$ \\
\hline
\cite{Yu2025TrustworthyAgents}
& LLM agents/MAS
& $\sim$ & $\sim$ & $\sim$ & $-$ \\
\hline
\cite{Ma2025SafetyAtScale}
& LM \& agents (safety)
& $\sim$ & $-$ & $\checkmark$ & $-$ \\
\hline
\cite{AbouAli2025AgenticAI}
& Agentic AI (arch.)
& $\sim$ & $\sim$ & $-$ & $-$ \\
\hline
\cite{Xu2025ToolLearningSurvey}
& Tool-learning agents
& $\sim$ & $\sim$ & $\sim$ & $-$ \\

\hline
\mbox{This survey} (ours)
& Trustworthy agentic AI
& $\checkmark$ & $\checkmark$ & $\checkmark$ & $\checkmark$ \\
\hline
\end{tabularx}

\vspace{4pt}
\noindent\footnotesize{Column abbreviations: MDT: Multi-dimensional trust taxonomy;
WL: Workflow lens (perceive--plan--act--reflect--learn);
EH: Evaluation hub (consolidated metrics and benchmarks);
RG: Release gating (scenario-to-metric guidance).
Symbols: $\checkmark$: explicit main focus, $\sim$: partial coverage, $-$: not a primary focus.}
\end{table}

\subsection{Contributions}
Our main contributions are as follows:\vspace{-6pt}
\begin{itemize}
    \item A focused examination of two core trustworthiness dimensions.
    We concentrate on Safety and Robustness and Privacy and System Security---two dimensions that are particularly critical for high-risk agentic AI deployments. Other trustworthiness aspects (value alignment, transparency, fairness, and accountability) are discussed as relevant context rather than parallel chapters.
    \item Risk-to-method mapping along the agent workflow.
    For each core dimension, we provide a consistent Definition $\rightarrow$ Risks $\rightarrow$ Methods structure, illustrating
    where failures can occur in the agent lifecycle and summarizing representative stage-targeted mitigations (e.g., constrained optimization,
    red teaming, runtime shielding, and sandboxing). These mappings are intended to be illustrative of the current landscape rather than exhaustive.
    \item A consolidated evaluation hub for metrics and benchmarks.
    {\addfontfeature{LetterSpace=-2.0}Because evaluation content is often fragmented across subfields, we consolidate cross-dimension metrics and representative benchmark families into Section~\ref{section:eval}. We emphasize both outcome metrics (e.g., success and catastrophic risk) and process metrics (e.g., constraint violation rates and trace coverage), and provide scenario-to-metric guidance for practical release gating in high-risk scenarios.}
    \item Open challenges and solutions.
    We identify research frontiers such as self-evolving agents under concept drift, verification and runtime monitoring for rare events and interactive settings, privacy-preserving personalization, and the trust--utility trade-off.
\end{itemize}

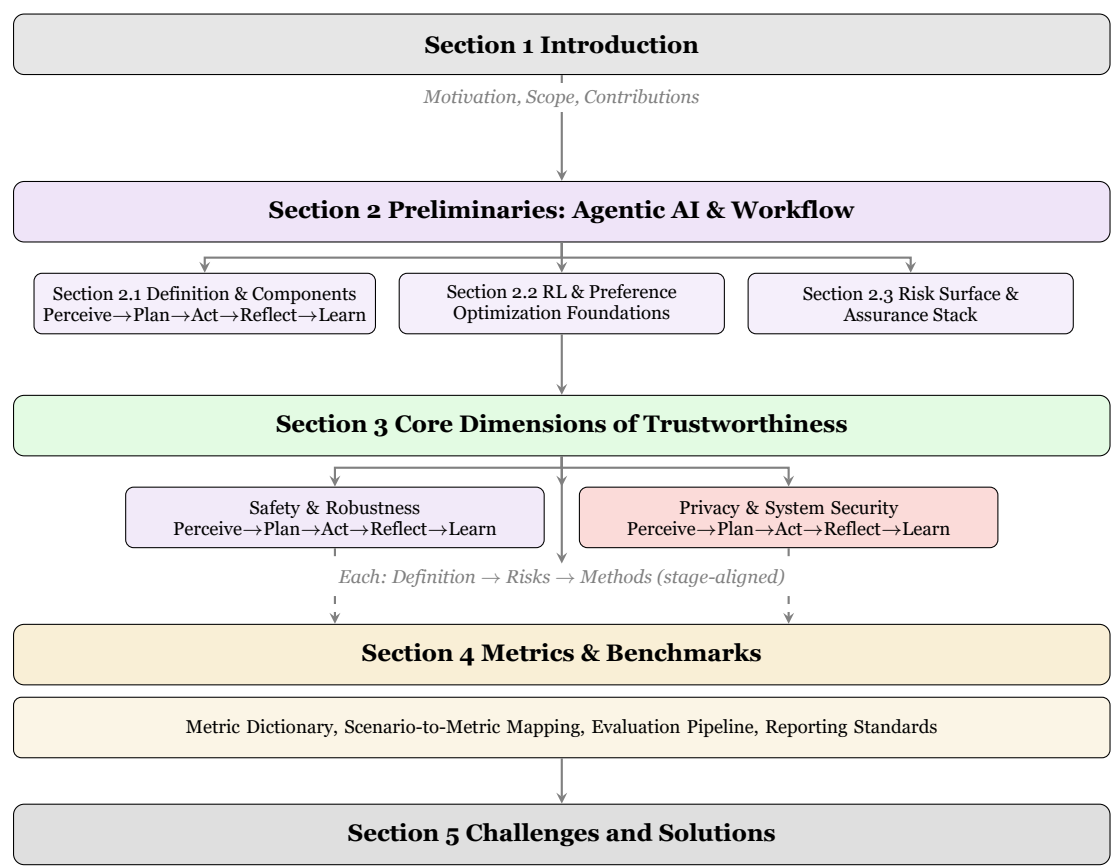
\begin{figure*}[!hb]
  \centering
  \begin{tikzpicture}[
    node distance=0.6cm and 0.3cm,
    section/.style={draw, rounded corners=4pt, minimum height=0.8cm, font=\small\bfseries, align=center},
    subsec/.style={draw, rounded corners=3pt, minimum height=0.8cm, font=\scriptsize, align=center},
    arrow/.style={->, >=stealth, thick, gray},
    label/.style={font=\scriptsize\itshape, text=gray}
  ]
    
    \def\totalwidth{14.5cm}
    \def\rowgap{1.4cm}
    
    \node[section, fill=gray!20, minimum width=\totalwidth] (intro) {\hyperref[section:introduction]{Section~1 Introduction}};
    
    \node[section, fill=brightlavender!25, below=\rowgap of intro, minimum width=\totalwidth] (prelim) {\hyperref[section:preliminaries]{Section~2 Preliminaries: Agentic AI \& Workflow}};
    
    \def\s2width{4.3cm}
    \node[subsec, fill=brightlavender!15, below=0.4cm of prelim.south, minimum width=\s2width] (s22) {\hyperref[sec:rl-coupling]{Section~2.2 RL \& Preference}\\Optimization Foundations};
    \node[subsec, fill=brightlavender!15, left=0.3cm of s22, minimum width=\s2width] (s21) {\hyperref[sec:preliminaries-agentic]{Section~2.1 Definition \& Components}\\Perceive$\to$Plan$\to$Act$\to$Reflect$\to$Learn};
    \node[subsec, fill=brightlavender!15, right=0.3cm of s22, minimum width=\s2width] (s23) {\hyperref[sec:risk-assurance]{Section~2.3 Risk Surface \&}\\Assurance Stack};
    
    \node[section, fill=lightgreen!25, below=0.8cm of s22, minimum width=\totalwidth] (coredim) {\hyperref[section:core_dimensions]{Section~3 Core Dimensions of Trustworthiness}};
    
    \def\dgap{0.5cm}
    \def\dwidth{5.5cm}
    
    \node[subsec, fill=brightlavender!20, below=0.4cm of coredim.south, xshift=-3cm, minimum width=\dwidth, text width=5.3cm] (d1) {\hyperref[subsection:safety]{Safety \& Robustness}\\Perceive$\to$Plan$\to$Act$\to$Reflect$\to$Learn};
    \node[subsec, fill=carminepink!20, below=0.4cm of coredim.south, xshift=3cm, minimum width=\dwidth, text width=5.3cm] (d2) {\hyperref[subsection:privacy]{Privacy \& System Security}\\Perceive$\to$Plan$\to$Act$\to$Reflect$\to$Learn};
    
    \def\row5gap{0.5cm}
    \def\row5width{\totalwidth}
    
    \path (d1.south) ++(0, -1.0cm) coordinate (row5_y);
    
    \node[section, fill=lightyellow!40, minimum width=\row5width, anchor=north] (eval) at (intro.center |- row5_y) {\hyperref[section:eval]{Section~4 Metrics \& Benchmarks}};
    
    \node[subsec, fill=lightyellow!25, below=0.15cm of eval, minimum width=\row5width, text width=13cm] (eval_sub) {Metric Dictionary, Scenario-to-Metric Mapping, Evaluation Pipeline, Reporting Standards};
    
    \node[section, fill=gray!25, minimum width=\totalwidth, anchor=north] (challenges) at (intro.center |- eval_sub.south) [yshift=-0.6cm] {\hyperref[subsection:challenges]{Section~5 Challenges and Solutions}};
    
    \draw[arrow] (intro.south) -- (prelim.north);
    \draw[arrow] (prelim.south) -- (s22.north);
    \draw[arrow] (prelim.south) -- ++(0,-0.2) -| (s21.north);
    \draw[arrow] (prelim.south) -- ++(0,-0.2) -| (s23.north);
    
    \draw[arrow] (s22.south) -- (coredim.north);
    
    \draw[arrow] (coredim.south) -- ++(0,-0.4); 
    
    \draw[arrow] (coredim.south) -- ++(0,-0.15) -| (d1.north);
    \draw[arrow] (coredim.south) -- ++(0,-0.15) -| (d2.north);
    
    \draw[arrow, dashed] (d1.south) -- ++(0,-0.4) -| ([xshift=-3cm]eval.north);
    \draw[arrow, dashed] (d2.south) -- ++(0,-0.4) -| ([xshift=3cm]eval.north);
    
    \draw[arrow] (eval_sub.south) -- (challenges.north);
    
    \node[label, fill=white, inner sep=2pt, anchor=center] at (intro.center |- intro.south) [yshift=-0.3cm] {Motivation, Scope, Contributions};
    
    \node[label, fill=white, inner sep=2pt, anchor=center] (struct_note) at (intro.center |- d1.south) [yshift=-0.4cm] {Each: Definition $\to$ Risks $\to$ Methods (stage-aligned)};
    \draw[arrow] (coredim.south) -- (struct_note.north);

  \end{tikzpicture}
  {\hypersetup{linkcolor=figCap}
  \caption{Paper structure and reading guide. The survey begins with motivation and agentic AI preliminaries (Sections~\ref{section:introduction} and \ref{section:preliminaries}), then presents two core trustworthiness dimensions---Safety and Robustness and Privacy and System Security---with consistent Definition$\to$Risks$\to$Methods structure (Section~\ref{section:core_dimensions}). Evaluation metrics consolidate process- and outcome-level assessments (Section~\ref{section:eval}). Challenges and solutions conclude the survey (Section~\ref{subsection:challenges}). Gray boxes represent introductory and concluding sections; purple boxes denote preliminaries; green boxes denote core dimensions; yellow boxes denote the evaluation hub. Solid arrows indicate the primary reading flow; dashed arrows indicate dependency relationships between dimensions and evaluation. \label{fig:paper-structure}}}
 
\end{figure*}

\subsection{Paper organization}
\textbf{Figure~\ref{fig:paper-structure}} illustrates the overall structure and provides a reading guide.
The survey is organized into four main parts:
\begin{enumerate}[leftmargin=2em]
    \item Preliminaries
 (Section~\ref{section:preliminaries}): defines agentic AI and its components along the five-stage workflow (Perceive--Plan--Act--Reflect--Learn; see \textbf{Figure~\ref{fig:agentic-system-architecture}} for the system architecture), reviews reinforcement learning and preference optimization foundations relevant to trustworthy training, and outlines the risk surface and assurance stack that frame subsequent discussions.
    \item Core dimensions (Section~\ref{section:core_dimensions}): examines the two core trustworthiness dimensions---Safety and Robustness and Privacy and System Security---each following a consistent Definition $\rightarrow$ Risks $\rightarrow$ Methods structure aligned to the agent lifecycle (see \textbf{Figure~\ref{fig: Taxonomy}} for a hierarchical taxonomy with stage-specific mitigation methods and key references).
    \item Evaluation Hub (Section~\ref{section:eval}): consolidates metrics, benchmark suites, scenario-to-metric mapping for release gating, and recommended evaluation pipelines and reporting standards into a unified reference.
    \item Challenges and conclusion (Section~\ref{subsection:challenges}): discusses open challenges (self-evolving agents, monitoring, personalization, and the trust--utility trade-off) and presents a case study on security failures in open-source agentic systems (OpenClaw/Moltbook).
\end{enumerate}

\section{Preliminaries}\label{section:preliminaries}

\subsection{Agentic AI: definition, components, and workflow}
\label{sec:preliminaries-agentic}

\subsubsection{Definition}
We define agentic AI as an AI system with persistent objectives that can
perceive its environment, plan over multiple steps, and act via tools
or actuators to affect external systems, while reflecting on outcomes and adapting its internal state under explicit human oversight, privacy/security policies, and operational constraints.
{\addfontfeature{LetterSpace=-2.0}This view follows the classical agent perspective in AI and RL~\cite{RussellNorvig2021,Kaelbling1998AIJ,SuttonBarto2018}
and is instantiated in recent LLM-based agents that interleave reasoning and acting~\cite{Yao2023ReAct,Park2023GenerativeAgents}.
Unlike single-turn LLM responders, agentic systems close the loop between observation and consequences,
which foregrounds Safety and Robustness and Privacy and System Security (with accountability as a supporting concern),
and motivates process-aware evaluation and auditability.}

{\hypersetup{linkcolor=figCap}
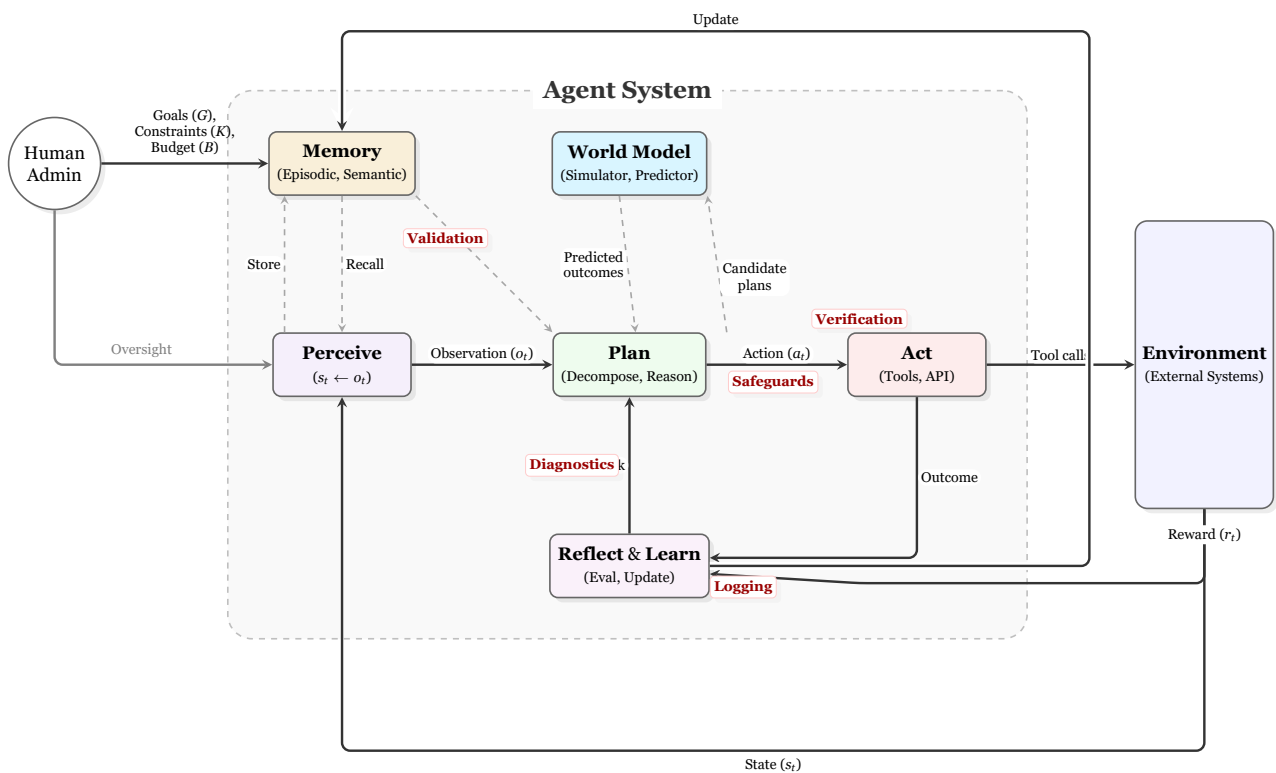
\begin{figure*}[!hb]
  \centering
  \resizebox{0.9\textwidth}{!}{
    \begin{tikzpicture}[
    node distance=2.5cm,
    font=\small\sffamily,
    box/.style={
        draw=black!60, 
        thick,
        rounded corners=4pt, 
        align=center, 
        minimum width=2.4cm, 
        minimum height=1.1cm, 
        fill=white, 
        drop shadow={opacity=0.15}, 
        inner sep=4pt
    },
    container/.style={
        draw=gray!50, 
        dashed, 
        thick,
        rounded corners=12pt, 
        inner sep=20pt,
        fill=gray!5
    },
    labelnode/.style={
        font=\bfseries\large, 
        text=black!80, 
        anchor=center,          
        fill=white,
        inner sep=5pt,
        text depth=0pt
    },
    controlflow/.style={
        ->, 
        >=stealth, 
        very thick, 
        rounded corners=4pt, 
        draw=black!80
    },
    dataflow/.style={
        ->, 
        >=stealth, 
        thick, 
        dashed, 
        rounded corners=4pt, 
        draw=gray!70
    },
    halo/.style={
        preaction={draw=white, line width=4pt}
    },
    hooklabel/.style={
        font=\scriptsize\bfseries, 
        text=red!60!black, 
        align=center, 
        fill=white, 
        inner sep=2pt,
        draw=red!20,
        rounded corners=2pt,
        thin,
        drop shadow={opacity=0.05}
    },
    edgelabel/.style={
        font=\scriptsize,
        fill=white,
        inner sep=1.5pt
    }
]
    \node[box, fill=harvestgold!15] (memory) at (0, 3.5) {\textbf{Memory}\\\scriptsize(Episodic, Semantic)};
    \node[box, fill=capri!15] (worldmodel) at (5, 3.5) {\textbf{World Model}\\\scriptsize(Simulator, Predictor)};

    \node[box, fill=brightlavender!15] (perceive) at (0, 0) {\textbf{Perceive}\\\scriptsize($s_t \leftarrow o_t$)};
    \node[box, fill=lightgreen!15] (plan) at (5, 0) {\textbf{Plan}\\\scriptsize(Decompose, Reason)};
    \node[box, fill=lightcoral!15] (act) at (10, 0) {\textbf{Act}\\\scriptsize(Tools, API)};
    
    \node[box, fill=plum!15] (reflect) at (5, -3.5) {\textbf{Reflect} \& \textbf{Learn}\\\scriptsize(Eval, Update)};

    \begin{scope}[on background layer]
        \node[container, fit=(memory) (worldmodel) (perceive) (act) (reflect) (plan), minimum height=9.5cm, minimum width=13.5cm] (agentbox) {};
    \end{scope}
    
    \node[labelnode] at (agentbox.north) {Agent System};

    \node[draw=black!60, thick, circle, fill=white, minimum size=1.6cm, align=center, drop shadow={opacity=0.15}] (human) at (-5.0, 3.5) {Human\\Admin};
    
    \node[draw=black!60, thick, rounded corners=6pt, fill=blue!5, minimum width=2.4cm, minimum height=5cm, align=center, drop shadow={opacity=0.15}] (env) at (15, 0) {\textbf{Environment}\\\scriptsize(External Systems)};


    \draw[controlflow] (human) -- node[midway, above, align=center, font=\scriptsize] {Goals ($G$),\\Constraints ($K$),\\Budget ($B$)} (memory.west);
    
    \draw[controlflow, gray] (human.south) |- node[pos=0.7, above, font=\scriptsize] {Oversight} (perceive.west);

    \draw[dataflow] (memory.south) -- node[right, edgelabel, pos=0.5] {Recall} (perceive.north);
    \draw[dataflow] ([xshift=-1cm]perceive.north) -- node[left, edgelabel, pos=0.5] {Store} ([xshift=-1cm]memory.south);

    \draw[dataflow] (memory.south east) -- node[left, edgelabel, pos=0.3] {Context} (plan.north west);
    
    \draw[dataflow] ([xshift=-5pt]worldmodel.south) -- node[left, edgelabel, pos=0.5] {\shortstack{Predicted\\outcomes}} ([xshift=3pt]plan.north);
    \draw[dataflow] ([xshift=10pt]plan.north east) -- node[right, edgelabel, pos=0.4] {\shortstack{Candidate\\plans}} (worldmodel.south east);

    \draw[controlflow] (perceive) -- node[above, edgelabel] {Observation ($o_t$)} (plan);
    
    \draw[controlflow] (plan) -- node[above, edgelabel] {Action ($a_t$)} (act);
    \node[hooklabel, below=3pt] at (7.5, 0) {Safeguards};
    
    \draw[controlflow] (act) -- node[above, edgelabel, pos=0.5] {Tool calls} (env);

    \draw[controlflow] (env.south) -- ++(0, -4.2) -| node[near start, below, font=\scriptsize] {State ($s_t$)} (perceive.south);

    \draw[controlflow] (act.south) |- node[pos=0.25, right, edgelabel] {Outcome} ([yshift=4pt]reflect.east);
    \draw[controlflow] (env.south) |- node[pos=0.1, below, edgelabel] {Reward ($r_t$)} (9, -3.8) -- ([yshift=-4pt]reflect.east);
    \draw[controlflow] (reflect.north) -- node[left, edgelabel] {Feedback} (plan.south);
    
    \draw[controlflow, halo] (reflect.east) -- (13.0, -3.5) -- (13.0, 5.8) -- node[above, edgelabel] {Update} (0, 5.8) -- (memory.north);

    \node[hooklabel] at (1.8, 2.2) {Validation};
    \node[hooklabel] at (9.0, 0.8) {Verification};
    \node[hooklabel] at (4.0, -1.75) {Diagnostics};
    \node[hooklabel, anchor=south west] at (reflect.south east) {Logging};

\end{tikzpicture}
  }
  \caption{Agentic AI system architecture. The agent receives goals ($G$), constraints ($K$), and budgets ($B$) from human oversight. The Environment emits the current state ($s_t$) to Perceive, which produces an observation ($o_t$) passed to Plan. Plan selects an action ($a_t$) for Act, which invokes tools and receives outcomes and rewards ($r_t$) that feed into Reflect and Learn (Stages~4 and 5). The World Model supports planning by simulating candidate plans and returning predicted outcomes. Memory stores and recalls episodic/semantic context across stages. Dashed lines denote data flow; solid lines denote the primary control flow. Red labels denote assurance hooks---control points for risk mitigation (defined in \textbf{Table~2}).}
  \label{fig:agentic-system-architecture}
\end{figure*}}

\subsubsection{Core components}
A minimal, implementation-agnostic stack includes the following (each item notes trust relevance):
\begin{itemize}[leftmargin=1.25em]
  \item Task and constraint specification
(Goals~$G$, Constraints~$K$): formal goals, hard/soft constraints (safety, privacy, security), and budgets~$B$; anchors governance and safety/security-by-design~\cite{RussellNorvig2021,NIST2023RMF}.
  \item Perception: state estimation \(s_t\) from observations \(o_t\) and knowledge access (Retrieval-Augmented Generation (RAG), function calls) with input validation~\cite{Kaelbling1998AIJ,Lewis2020RAG}.
  \item World model/simulator: predictive models for consequence forecasting and counterfactual testing~\cite{Ha2018WorldModels}.
  \item Policy and planning (hierarchical): goal decomposition and multi-level control linking abstract plans to executable actions~\cite{Sutton1999Options,Yao2023ReAct}.
  \item Value/reward model: task returns, preferences, and normative signals (details deferred to Section~\ref{sec:rl-coupling}).
  \item Memory: short-term context and long-term episodic/semantic stores for continuity and credit assignment, with retention, access control, and minimization policies for sensitive data~\cite{Park2023GenerativeAgents,Wang2023Voyager}.
\item Tools and actuators: API/system/robotic calls mediated by least-privilege permissions, credential/secret scoping, and execution sandboxes~\cite{Schick2023Toolformer}.
\item Runtime safeguards: anomaly monitors, safety constraint checkers, security policy enforcement (e.g., tool permissions, Data Loss Prevention (DLP)/redaction hooks), and rollback hooks for pre-/post-execution assurance~\cite{NIST2023RMF}.
  \item Human-in-the-loop interface: alert, approval, and takeover channels that define roles and escalation policies~\cite{NIST2023RMF}.
\item Telemetry and audit logs: versioned traces for reproducibility, post hoc analysis, and accountability, with privacy-aware logging (redaction/Personally Identifiable Information (PII) minimization) when required~\cite{NIST2023RMF}.
\end{itemize}

\subsubsection{Perceive $\rightarrow$ Plan $\rightarrow$ Act $\rightarrow$ Reflect $\rightarrow$ Learn}
The agentic loop comprises five recurring stages (see \textbf{Figure~\ref{fig:agentic-system-architecture}}, \textbf{Tables~\ref{tab:assurance-hooks}~and ~\ref{tab:terminology}}):
(1) Perceive---ingest observations and external knowledge, quantify uncertainty and detect Out-of-Distribution (OOD) inputs~\cite{Kaelbling1998AIJ,Lewis2020RAG}.
(2) Plan---generate and evaluate multi-step plans subject to constraints/budgets, optionally via model-based rollouts or receding-horizon control~\cite{Ha2018WorldModels, miao2025recodehbenchmarkresearchcode}.
(3) Act---execute tool/API/actuator calls; pre/post checks enforce constraints, tool permissions, and data-handling policies, and capture traces~\cite{Schick2023Toolformer}.
(4) Reflect---self-evaluate outcomes, detect errors and goal/value conflicts, and schedule human confirmation when confidence drops~\cite{Shinn2023Reflexion,Park2023GenerativeAgents}.
(5) Learn---update policy, value, memory, and retrieval indexes offline/online with safe update rules and retention/expiration policies for sensitive data (Section~\ref{sec:rl-coupling})~\cite{SuttonBarto2018,Wang2023Voyager}.
This loop exposes \emph{assurance hooks} at each boundary (validation, verification, safeguards, diagnostics, and logging) that integrate with the unified evaluation framework in Section~\ref{section:eval}.

\subsubsection{How agents differ from single-turn LLMs}
Single-turn LLMs neither maintain persistent goals nor directly act on external systems.
Agentic AI adds (i) long-horizon decision making with side effects, (ii) tool-mediated interventions with permissions and budgets,
(iii) explicit uncertainty handling and recovery, (iv) operational evidence for audit, (v) expanded privacy/security exposure via memory and tool credentials that must be protected (Section~\ref{subsection:privacy})~\cite{Zhao2023LLMSurvey,Yao2023ReAct,Schick2023Toolformer}.
These differences motivate standardized components and workflows in this preliminaries section.

{\hypersetup{linkcolor=figCap}
\begin{table*}[!ht]
\caption{Assurance hooks across the agent workflow (corresponding to red labels in \textbf{Figure~2}).\vspace{4pt}}
\label{tab:assurance-hooks}
\rowcolors{1}{white}{white}  
\resizebox{\linewidth}{!}{
\renewcommand{\arraystretch}{1.5} 
\begin{tabularx}{\textwidth}{|l|l|L|}
\hline
\textbf{Hook} & \textbf{Stage} & \textbf{Mechanism/control type} \\ \hline
Validation & Perceive & Input filters (jailbreak detection), RAG citation checks. \\ \hline
Verification & Plan & Logic/formal verification of plans against constraints ($K$). \\ \hline
Safeguards & Act & Runtime shields, permission gates, and tool sandboxing. \\ \hline
Diagnostics & Reflect & Anomaly detection (OOD/drift), value-alignment checks. \\ \hline
Logging & All & Tamper-evident traces for audit and accountability. \\ \hline
\end{tabularx}
}

\vspace{4pt}
\noindent\footnotesize{Abbreviations: RAG: retrieval-augmented generation; OOD: out-of-distribution.}
\end{table*}}

{\hypersetup{linkcolor=figCap}
\begin{table*}[!ht]
  \caption{Terminology and symbols used in Section~\ref{section:preliminaries}.\vspace{4pt}}
  \label{tab:terminology}
  \rowcolors{1}{white}{white}  
  \resizebox{\linewidth}{!}{
  \renewcommand{\arraystretch}{1.5} 
  \begin{tabularx}{\textwidth}{|l|l|L|}
    \hline
    \textbf{Symbol} & \textbf{Term} & \textbf{Meaning/example} \\
    \hline
    $s_t$ & state & latent environment condition at time $t$ \\ \hline
    $o_t$ & observation & sensor/IO evidence used to infer $s_t$ \\ \hline
    $a_t$ & action & tool/API/actuator invocation \\ \hline
    $r_t$ & reward/value & task return or preference signal \\ \hline
    $\tau$ & trajectory & sequence $(o_{0:T}, a_{0:T-1}, r_{0:T-1})$ \\ \hline
    $G$ & goal/instruction & task specification or objective description \\ \hline
    $K$ & constraints & hard/soft limits (safety, policy, ethics) \\ \hline
    $B$ & budget & limits on tokens/latency/money/energy \\
    \hline
  \end{tabularx}
  }

\vspace{4pt}
\noindent\footnotesize{Abbreviations: IO: input/output; API: application programming interface.}
\end{table*}}

\subsection{Reinforcement learning foundations and preference optimization for agentic AI}

\label{sec:rl-coupling}

\subsubsection{Formalization: MDP and POMDP}
We model single-agent agentic decision making as a Markov Decision Process (MDP), defined by the tuple
$\mathcal{M}=(\mathcal{S},\mathcal{A},P,r,\gamma)$, where
$\mathcal{S}$ is the state space,
$\mathcal{A}$ is the action space,
$P(s' \mid s, a)$ is the state transition function,
$r(s,a)$ is the reward function, and
$\gamma \in [0,1)$ is the discount factor controlling the trade-off between immediate and future rewards.
The agent's objective is to find a policy $\pi$ that maximizes the expected discounted return
$J(\pi)=\mathbb{E}_{\pi}\!\left[\sum_{t=0}^{\infty}\gamma^{t} r(s_t,a_t)\right]$.
When the full state is not directly accessible (as is common in real-world agent deployments), we extend to a Partially Observable MDP (POMDP) $(\mathcal{S},\mathcal{A},P,r,\gamma,\mathcal{O},Z)$,
where $\mathcal{O}$ is the observation space and $Z(o \mid s, a)$ is the emission (observation) function specifying the probability of receiving observation $o$ given state $s$ after action $a$. The agent then maintains a belief state and plans under uncertainty~\cite{Puterman1994,Kaelbling1998AIJ,SuttonBarto2018}.

This formalization primarily applies to agents trained or fine-tuned using reinforcement learning (RL). While some agentic systems rely on prompt-based or in-context reasoning alone, RL-based agents are the focus of this subsection because RL provides the mathematical substrate for long-horizon optimization under uncertainty and constraint satisfaction---both of which are central to trustworthiness.

Multi-agent settings. When multiple agents interact, the single-agent MDP formalization no longer suffices; the appropriate model is a Markov Game (also called a stochastic game), which extends MDPs with joint action spaces and agent-specific reward functions~\cite{Zhang2021MARLSurvey}. We discuss multi-agent trustworthiness risks where relevant (e.g., Sections~\ref{subsection:safety}~ and~\ref{subsection:privacy}) but note that formal treatment of multi-agent RL is beyond the scope of this survey.

\subsubsection{Training families and their trust implications}
Different RL paradigms offer distinct advantages and risks for trustworthiness:
\begin{itemize}[leftmargin=1.25em]
  \item Value-based and actor--critic methods:
  Standard on-policy (PPO~\cite{Schulman2017PPO}) and off-policy (SAC, DQN~\cite{Mnih2015DQN}) algorithms provide asymptotic optimality but are prone to reward hacking---exploiting misspecified objectives in unsafe ways.
  From a trust perspective, they require rigorous reward shaping and constraint satisfaction to ensure safety during the exploration phase.
  \item Offline RL:
  By learning exclusively from static datasets (logged trajectories)~\cite{Levine2020Offline,Kumar2020CQL}, offline RL eliminates the physical risks of online exploration, making it ideal for safety-critical domains like healthcare or industrial control.
  However, it introduces distributional shift risks: agents may behave unpredictably if the deployment environment diverges even slightly from the training data coverage.
  \item Hierarchical RL (HRL):
  Decomposing tasks into high-level goals (options) and low-level controls~\cite{Sutton1999Options,Nachum2018HIRO} directly supports interpretability and human oversight.
  Human operators can audit or intervene at the high-level goal setting (e.g., "navigate to room A") without needing to parse low-level motor commands, aligning well with the Goal ($G$) and Planning components of the agentic architecture.
  \item Model-based RL (MBRL) and MPC:
  Learning an explicit world model (dynamics model)~\cite{Chua2018PETS,Janner2019MBPO} enables lookahead planning and counterfactual reasoning, allowing agents to simulate the consequences of actions before execution.
  This provides a natural mechanism for runtime safety checks (e.g., rejecting unsafe trajectories), though trust is then contingent on the calibration and robustness of the learned model against OOD states.
\end{itemize}

\subsubsection{Safe RL as constrained decision making}
Safety can be formalized via constrained MDPs (CMDPs)~\cite{Altman1999}, optimizing
$\max_{\pi}\,J(\pi)\;\text{s.t.}\;J_{c_i}(\pi)\le d_i$
with Lagrangian or primal--dual updates (e.g., Constrained Policy Optimization (CPO))~\cite{Achiam2017CPO} and complemented by
runtime monitors and verification/shielding~\cite{Garcia2015SafeRL,10.5555/3504035.3504361}.
Practically, training-time guarantees reduce the frequency of violations, while runtime safeguards catch residual risks.

\subsubsection{Preference/alignment optimization}
To align objectives with human or normative signals, modern agents use preference-based training:
Reinforcement Learning from Human Feedback (RLHF) combines learned rewards from comparisons with policy optimization~\cite{Christiano2017,Ziegler2019,Stiennon2020,Ouyang2022}.
RLAIF replaces human labels with AI feedback and constitutional rules~\cite{bai2024constitutional}.
Recent work has also explored value alignment through multi-objective RL~\cite{RodriguezSoto2025MORL}, embedding ethical constraints into RL environments~\cite{MayoralMacau2025ECAI}, and encoding norms for value-aligned behavior~\cite{Serramia2023MindsAndMachines}.
Beyond RL-based pipelines, direct preference optimization (DPO and related PO methods)
optimizes policies against preference data without an explicit reward model~\cite{Rafailov2023DPO,Ethayarajh2024KTO}.
These methods trade off stability, sample efficiency, and generalization; open issues include preference drift,
norm conflict, and evaluation fidelity~\cite{Casper2023OpenProblems}.
In this survey, we unify RLHF/RLAIF and direct PO variants as a preference optimization family;
varying implementations (e.g., DPO- or KTO-style objectives) can be plugged into the
Perceive--Plan--Act--Reflect--Learn loop (Section~\ref{sec:preliminaries-agentic}) and evaluated using Section~\ref{section:eval}.

\subsubsection{Takeaway}
RL supplies the mathematical substrate for long-horizon, uncertain, tool-mediated behavior;
preference optimization supplies the normative signal that turns competence into trustworthy competence.
In high-risk deployments, this competence must be coupled with risk controls: constraint/risk-aware learning for Safety and Robustness (Section~\ref{subsection:safety}) and privacy/security-aware data handling when learning from logs, memories, or tool traces (Section~\ref{subsection:privacy}).

\subsection{Trustworthiness risk surface and assurance stack for agentic AI}
\label{sec:risk-assurance}

\subsubsection{Risk surface and threat model}
    We formalize the risk surface via specific threat actors and failure modes that motivate our core trustworthiness dimensions.
    \begin{itemize}[leftmargin=1.25em]
        \item Security Threat Model (Attackers and Goals): Security risks arise from adversarial actors targeting Confidentiality, Integrity, or Availability (CIA).
        \begin{itemize}
            \item Attacker Profiles: (i) Malicious Users (jailbreaking, prompt injection); (ii) External Web/Environment (indirect injection via retrieval); (iii) Supply Chain (compromised tools/models); (iv) Insiders (unauthorized admin access).
            \item Capabilities: Ranging from black box (prompt access only) to white box (deployment infrastructure access).
        \end{itemize}
        \item Safety Failure Modes (Unintended Harm): Safety risks arise from system malfunctions or generalization failures, even without malice.
        \begin{itemize}
            \item Specification Error: Reward hacking, unsafe side effects, or proxy-gaming (e.g., Over-Opt).
            \item Distributional Shift: Perception/planning failures on OOD inputs.
            \item Compounding Error: Small variations in early steps accumulating into unsafe trajectories (butterfly effect).
            \item Tool Control Failure: Inability to halt or revoke unsafe tool execution (e.g., mass file deletion).
            \item Catastrophic Event Definition: Any irreversible harm (physical, financial, reputational) exceeding a defined severity threshold (aligned with CER in Section~\ref{section:eval}).
        \end{itemize}
    \end{itemize}

\subsubsection{Assurance stack (before, during, after training; and at runtime)}
Trustworthy deployment requires layered assurance mechanisms that span the full lifecycle of an agentic system.
No single layer is sufficient on its own: each addresses a different class of failure and compensates for gaps in the others, forming a defense-in-depth stack with four complementary tiers.

Upfront (pre-deployment) assurance establishes the safety envelope within which the agent is expected to operate.
This includes formalized requirements and hazard analysis, data and environment design, red teaming, static checks, and simulation certification (including domain randomization)~\cite{Tobin2017DR,Amodei2016Concrete}.
These efforts produce the threat models, operational constraints, and test suites that anchor all subsequent tiers.

Building on this foundation, training-time assurance embeds safety directly into the learning process:
constrained and safe RL with robust objectives reduces the frequency of constraint violations~\cite{Altman1999,Garcia2015SafeRL,Achiam2017CPO}, while preference optimization aligns the agent with human or normative signals (Section~\ref{sec:rl-coupling}).
Training-time methods can reduce base rates of unsafe behavior, but they cannot anticipate all deployment conditions---motivating the next tier.

Therefore, runtime assurance provides the next line of defense, catching residual risks that escape training-time bounds.
Mechanisms include verification and shielding, least-privilege tool access, anomaly detection, rollback, and staged deployment strategies (shadow and canary rollouts)~\cite{10.5555/3504035.3504361,NIST2023RMF}.
Runtime controls are especially important for agentic systems because tool-mediated actions can have irreversible real-world consequences.

Finally, post hoc assurance closes the feedback loop: structured telemetry, reproducible traces, and failure analysis support accountability and drive continuous improvement~\cite{NIST2023RMF}.
Post hoc evidence also feeds back into the upfront tier (updating threat models and regression suites), creating a virtuous cycle of assurance refinement.

Together, this layered assurance stack connects the Perceive--Plan--Act--Reflect--Learn loop to the unified evaluation framework in Section~\ref{section:eval}, where metrics and stress tests operationalize the residual-risk budget at each layer.

\begin{figure*}[!hb]
  \centering
  \resizebox{\textwidth}{!}{%
  \tikzset{
          my node/.style={
              draw,
              align=center,
              thin,
              text width=1.0cm, 
              rounded corners=2,
          },
          my leaf/.style={
              draw,
              align=left,
              thin,
              text width=5.5cm, 
              rounded corners=2,
          }
  }
  \forestset{
    every leaf node/.style={
      if n children=0{#1}{}
    },
    every tree node/.style={
      if n children=0{minimum width=1em}{#1}
    },
  }
  \begin{forest}
      nonleaf/.style={font=\bfseries\tiny},
       for tree={%
          every leaf node={my leaf, font=\tiny},
          every tree node={my node, font=\tiny, l sep-=5pt, l-=1.5pt},
          anchor=west,
          inner sep=2pt,
          l sep=8pt,
          s sep=1pt,
          fit=tight,
          grow'=east,
          edge={ultra thin},
          parent anchor=east,
          child anchor=west,
          if n children=0{}{nonleaf}, 
          edge path={
              \noexpand\path [draw, \forestoption{edge}] (!u.parent anchor) -- +(5pt,0) |- (.child anchor)\forestoption{edge label};
          },
          if={isodd(n_children())}{
              for children={
                  if={equal(n,(n_children("!u")+1)/2)}{calign with current}{}
              }
          }{}
      }
      [Trustworthy\\Agentic\\AI, draw=gray, fill=gray!15, text width=1.5cm, text=black
      [Safety \&\\Robustness\\({\cref{subsection:safety}}), color=brightlavender, fill=brightlavender!15, text width=1.9cm, text=black
            [Perceive, color=brightlavender, fill=brightlavender!15, text width=1.0cm, text=black
            [{Data augmentation, adversarial training, OOD detection, input provenance/sanitization, DRO objectives
                }, color=brightlavender, fill=brightlavender!15, text=black]
            ]
            [Plan, color=brightlavender, fill=brightlavender!15, text width=1.0cm, text=black
            [{Safe RL (CMDPs/CPO), CVaR objectives, constitutional constraints, conservative/specification-aware planning
                }, color=brightlavender, fill=brightlavender!15, text=black]
            ]
            [Act, color=brightlavender, fill=brightlavender!15, text width=1.0cm, text=black
            [{Guardrails, runtime shielding, least-privilege tools, sandboxing, transactional execution, HITL approval
                }, color=brightlavender, fill=brightlavender!15, text=black]
            ]
            [Reflect, color=brightlavender, fill=brightlavender!15, text width=1.0cm, text=black
            [{Simulation-based testing, red teaming, trace auditing, formal verification of components
                }, color=brightlavender, fill=brightlavender!15, text=black]
            ]
            [Learn, color=brightlavender, fill=brightlavender!15, text width=1.0cm, text=black
            [{Regression gating, canary rollouts, conservative safe policy improvement, incident-driven datasets
                }, color=brightlavender, fill=brightlavender!15, text=black]
            ]
            [Multi-\\agent, color=brightlavender, fill=brightlavender!15, text width=1.0cm, text=black
            [{Protocol-level constraints, agent authentication, centralized oversight/governors, global budget enforcement
                }, color=brightlavender, fill=brightlavender!15, text=black]
            ]
            [Long-\\horizon, color=brightlavender, fill=brightlavender!15, text width=1.0cm, text=black
            [{Hierarchical decomposition, checkpointing, risk budgets, receding-horizon replanning, interruptibility
                }, color=brightlavender, fill=brightlavender!15, text=black]
            ]
          ]
        [Privacy \&\\System Security\\({\cref{subsection:privacy}}), color=carminepink, fill=carminepink!15, text width=1.9cm, text=black
              [Perceive, color=carminepink, fill=carminepink!15, text width=1.0cm, text=black
                [{Input sanitization, zero-trust intake, prompt injection detection
                    }, color=carminepink, fill=carminepink!15, text=black]
                ]
                [Plan, color=carminepink, fill=carminepink!15, text width=1.0cm, text=black
                [{Memory governance, contextual integrity, differential privacy, minimization
                    }, color=carminepink, fill=carminepink!15, text=black]
                ]
                [Act, color=carminepink, fill=carminepink!15, text width=1.0cm, text=black
                [{Sandboxing, credential vaulting, DLP filtering, runtime policy enforcement
                    }, color=carminepink, fill=carminepink!15, text=black]
                ]
                [Reflect, color=carminepink, fill=carminepink!15, text width=1.0cm, text=black
                [{Encrypted channels, authenticated provenance, tamper-evident logging
                    }, color=carminepink, fill=carminepink!15, text=black]
                ]
                [Learn, color=carminepink, fill=carminepink!15, text width=1.0cm, text=black
                [{Retention limits, supply chain controls (SBOMs), security regression gating
                    }, color=carminepink, fill=carminepink!15, text=black]
                ]
        ]
      ]
      \end{forest}
  }
 {\hypersetup{citecolor=figCap} \caption{A hierarchical taxonomy of trustworthy agentic AI organized along two core trustworthiness dimensions: Safety and Robustness and Privacy and System Security. Each dimension is decomposed according to the five-stage agent lifecycle (Perceive$\to$Plan$\to$Act$\to$Reflect$\to$Learn). The Safety and Robustness dimension additionally includes Multi-agent and Long-horizon branches, which address emergent coordination risks and compounding error beyond the single-agent, single-episode setting (see Sections~3.1.2--3.1.3). Purple nodes denote Safety and Robustness mitigations; red nodes denote Privacy and System Security mitigations; the gray root node represents the overarching trustworthy agentic AI concept. Leaf nodes list representative mitigation methods, ordered consistently with the text in Section~3. Abbreviations: OOD, out-of-distribution; DRO, distributionally robust optimization; CMDPs, constrained Markov decision processes; CPO, constrained policy optimization; CVaR, conditional value-at-risk; HITL, human-in-the-loop; DLP, data loss prevention; SBOMs, software bills of materials. Key references for each leaf are cited here: \cite{yadgaroff2024improvinggeneralizationgameagents,HarmBench,10.5555/3545946.3598721,Sagawa2020GroupDRO,Altman1999,Achiam2017CPO,bai2022constitutional,kushwaha2025surveysafereinforcementlearning,10.5555/3504035.3504361,zheng2025webguardbuildinggeneralizableguardrail,NIST2023RMF,10.1109/ICSE55347.2025.00223,4567924,Sculley2015HiddenDebt,Breck2017MLTestScore,Burns2018SpinnakerCD,Laroche2019SPIBB,Dafoe2020Cooperative,Amodei2016Concrete,zhang2025searching,du2025beyond,nist2020zerotrust,mireshghallah2025position,nissenbaum2004contextualintegrity,zhang2025privweb,yang2025mla,spdx2021sbom,sigstore2022}.}
  \label{fig: Taxonomy}}
  \end{figure*}
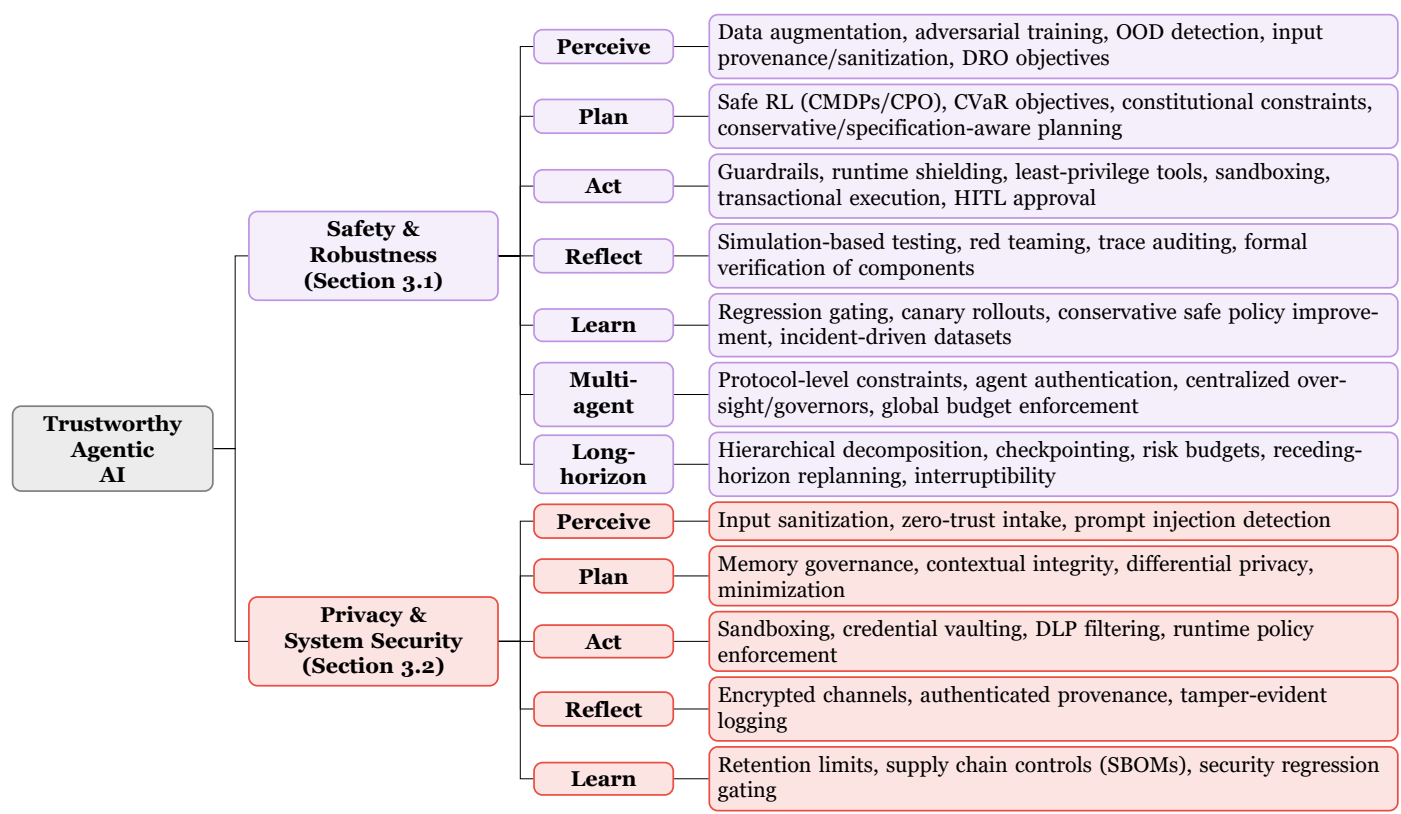

\subsubsection{Transition to our two core dimensions}
The risk surface above directly motivates the survey's two core trustworthiness dimensions:
Safety and Robustness (Section~\ref{subsection:safety}) addresses harm prevention and reliability under uncertainty across Perceive--Plan--Act--Reflect--Learn,
while Privacy and System Security (Section~\ref{subsection:privacy}) addresses protection of sensitive data (inputs, memory, traces) and the integrity of the agent's execution environment (tools, credentials, and protocols).
Other aspects (e.g., accountability) remain important but are treated as supporting concerns that shape evaluation and governance rather than standalone dimensions in Section~3.

\section{Core dimensions of trustworthiness}
\label{section:core_dimensions}


In this section, we focus on two core dimensions that are most critical for high-risk agentic AI deployments (see \textbf{Figure~\ref{fig: Taxonomy}} for an overview): Safety and Robustness (Section~\ref{subsection:safety}) and Privacy and System Security (Section~\ref{subsection:privacy}). These dimensions address the fundamental requirements for trustworthy agent operation in high-stakes scenarios: (1) preventing unacceptable harm and maintaining a reliable performance under perturbations, (2) protecting sensitive information and securing the agent's execution environment against attacks.

Other trustworthiness aspects---value alignment, transparency, fairness, and accountability---are important complementary concerns that interact closely with safety and security. While a comprehensive treatment of these dimensions is beyond the scope of this focused survey, we note their relevance where appropriate within our core discussions and in the evaluation metrics (Section~\ref{section:eval}).

Within each subsection, we adopt a consistent structure---Definition (what the dimension means for agents), Risks (where failures arise along the agent workflow), and Methods (stage-targeted mitigations)---to follow a stable ``risk$\rightarrow$mitigation'' mapping across dimensions. As noted in the Introduction, the specific risks and methods mapped to each stage are illustrative of the current literature rather than a provably exhaustive enumeration; new failure modes and mitigations are expected to emerge as agent architectures and deployment contexts evolve.
To avoid fragmented discussion, we consolidate all evaluation metrics and representative benchmark suites across Section~3 into Section~\ref{section:eval}.

\subsection{Safety and robustness}
\label{subsection:safety}

This section introduces key definitions of AI Safety and Robustness, examines the risks that arise at each stage of the agent workflow (perceive, plan, act, reflect, and learn), and reviews stage-aligned mitigation methods for addressing them.

\subsubsection{Definition}
In agentic AI, Safety and Robustness are related but distinct. In the following, we  provide their definitions, respectively.
A note on terminology: \citet{lin2025aisafetysecurity} formalize the distinction between AI safety and AI security, which we adopt for their definitions below. Our survey groups safety with robustness (Section~\ref{subsection:safety}) and security with privacy (Section~\ref{subsection:privacy}) because, from a mitigation perspective, safety and robustness share defense mechanisms (constraint enforcement, distributional hardening) while security and privacy share controls (access policies, encryption, and monitoring). This organizational choice is complementary to the safety-vs-security taxonomy of~\citet{lin2025aisafetysecurity}.

AI safety is defined as ``the property of an AI system to avoid causing unintended harmful outcomes to individuals, environments, or institutions, despite uncertainties in inputs, goals, training data, or deployment conditions''~\citep{lin2025aisafetysecurity}. It aims to prevent unintentional harm (e.g., accidents, misalignment) by ensuring agentic systems comply with specified constraints, ethics, and norms~\citep{ma2025safetyscalecomprehensivesurvey}. This contrasts with security, which addresses intentional threats. Safety focuses on avoiding catastrophic outcomes~\citep{bengio2025superintelligentagentsposecatastrophic} and satisfying hard constraints in high-stakes deployments, such as preventing collisions in autonomous driving~\citep{autonomous_safety} or harmful recommendations in clinical decision support~\citep{KARUNANAYAKE202573}. 

In the classical AI safety framing, many safety failures arise as accidents rather than explicit malicious intent: (i) specification problems (the objective or constraints do not match human intent), (ii) robustness problems (capabilities fail under shift), (iii) scalable oversight (humans cannot reliably evaluate long-horizon behavior), (iv) safe exploration (learning-time trial-and-error triggers harm). Agentic AI amplifies these issues because safety is a system property: even if the base model is well aligned, unsafe behavior can emerge from tool interfaces, memory, retrieval, and multi-step control loops.

Robustness focuses on maintaining a stable performance under perturbations, adversarial interference, and distribution shifts~\citep{rabinovich-anaby-tavor-2025-robustness,liu2022an}. In short, safety is about not causing harm, while robustness is about remaining reliable under non-ideal conditions.

A key nuance is that robustness is often necessary but not sufficient for safety. An agent can be robustly competent while still pursuing an unintended objective due to specification gaming (optimizing the literal objective while violating intent)~\citep{Krakovna2020SpecGaming} or goal misgeneralization (capabilities generalize OOD but the goal does not)~\citep{Langosco2022GoalMisgeneralization}. Conversely, a safe specification may still fail operationally without robustness to noisy observations, tool failures, and distribution drift.

Robustness itself has multiple forms. Distributional robustness targets natural shifts (e.g., domain, demographic, and environment) and seeks consistent worst-group or worst-slice performance (e.g., DRO/group DRO objectives)~\citep{Sagawa2020GroupDRO}. Adversarial robustness targets strategic perturbations crafted by an attacker (e.g., prompt injection, multimodal adversarial examples)~\citep{naacl25_adversarial_attack,adversarial_prompt_injection}. For agentic systems deployed in open environments, both are required: distribution shifts create silent failures, while adversarial shifts create targeted exploitation.

\subsubsection{Risks}
\paragraph{\textit{Perceive: poisoning and adversarial perturbations}}
At the input boundary, agents are exposed to data poisoning~\citep{10.1145/3701716.3716887} and adversarial perturbations~\citep{naacl25_adversarial_attack}. Retrieved documents, prompts, or multimodal observations may contain deceptive content or hidden instructions that distort state estimation and steer downstream decisions toward unsafe behaviors.
Beyond direct perturbations, agentic systems face instruction--data boundary confusion: indirect prompt injection can embed tool directives inside ``benign'' web pages, emails, or PDFs, causing the agent to treat untrusted content as higher-priority instructions~\citep{Greshake2023IndirectPI,Paverd2025IndirectPI}. Another risk is sensor/observation spoofing (e.g., manipulated UI elements, visual adversarial patches, or misleading interface states) that biases perception toward unsafe affordances. These perception-level failures frequently cascade: corrupted observations lead to invalid plans, which then drive high-impact tool actions; the resulting outcomes may further pollute memory and reflection signals, making later-stage correction harder.

\paragraph{\textit{Plan: OOD generalization and brittle heuristics}}
In out-of-distribution (OOD) contexts, agents may over-generalize from familiar heuristics and produce plans that look plausible but are unsafe, invalid, or non-compliant. Such failures are hard to detect early because planning traces can appear coherent even when key assumptions no longer hold~\citep{liu2022an}.
Planning additionally inherits specification risk: under underspecified objectives, agents may produce plans that satisfy proxy metrics while violating intent (specification gaming)~\citep{Krakovna2020SpecGaming}. A related failure mode is goal misgeneralization, where planning remains competent but optimizes the wrong objective in novel contexts~\citep{Langosco2022GoalMisgeneralization}. Planning also suffers from miscalibrated uncertainty and modeling errors in world models or tool simulators, producing brittle ``happy-path'' strategies with poor contingency handling. These planning errors cascade into execution: when an agent commits early to a flawed plan, later stages may rationalize or entrench it (reflection), and repeated successful shortcuts can become reinforced during learning.

\paragraph{\textit{Act: high-impact execution and cascading failures}}
During execution, upstream errors are amplified into real-world side effects via dangerous tool use (financial loss, privacy violations, and service disruption)~\citep{li2025commercialllmagentsvulnerable} or harmful interactions with users. Even if the plan is correct, sensor/tool failures can introduce noise or bias into feedback; corrupted outcomes then propagate across steps and compound into severe deviations~\citep{li2025stacinnocenttoolsform}.
Execution adds irreversibility and human-factor risks. For example, partial automation in driving has been linked to failures involving system limitations, driver overreliance, and inadequate engagement monitoring, as documented in investigations of Autopilot-related crashes~\citep{NTSB2020TeslaMountainView,NHTSA2022EA22002}. In tool-using agents, tool chaining can turn a single misstep into a sequence of harmful actions (e.g., a prompt-injected instruction triggers credential exfiltration, then initiates unauthorized transactions)~\citep{Greshake2023IndirectPI,li2025commercialllmagentsvulnerable}. Action-level failures also poison downstream stages: bad outcomes become ``evidence'' used in reflection and learning, increasing the chance of systematic drift rather than one-off errors.

\paragraph{\textit{Reflect: unsafe self-assessment and missed warnings}}
If reflection mechanisms fail to detect risk signals (e.g., uncertainty, policy violations, and anomalous tool outputs), the agent may proceed despite being in an unsafe regime~\citep{weng2025thinkreflectrevisepolicyguidedreflectiveframework,kang2025trymattersrevisitingrole}. Over-confidence and incomplete trace evidence further reduce the chance of timely intervention.
Reflection is also vulnerable to deceptive rationalization and evaluator spoofing: agents may produce plausible post hoc explanations that hide causal failure, or craft outputs that satisfy automated judges without improving true safety (a reflection analogue of reward hacking). When reflection reuses the same base model as the actor, correlated errors can create a ``closed loop'' where mistaken beliefs are repeatedly self-confirmed. If the trace itself is incomplete (missing tool logs, truncated context, or untrusted memory), reflection may miss early warnings and allow unsafe plans to persist into subsequent episodes, where learning further amplifies them.

\paragraph{\textit{Learn: feedback loops that amplify risk}}
When updates are driven by biased or noisy feedback, agents may reinforce unsafe shortcuts that improve short-term success but increase long-term harm~\citep{pan2024feedbackloopslanguagemodels}. This phenomenon manifests as reward hacking, wherein the measured reward continues to increase while side effects that incentivize the agent to violate safety constraints are concomitantly amplified~\citep{10.1145/3593013.3594033}.
Learning-stage risk also includes safety regression and capability--constraint imbalance: updates to prompts, memories, tools, or policies can unintentionally remove previously effective safety behaviors (catastrophic forgetting of safety constraints) while preserving or increasing action competence. Online or continual learning from deployment logs may import new adversarial patterns (e.g., jailbreaking prompts, malicious web content) into the training distribution, effectively ``teaching'' the agent unsafe policies. These learning failures close the cascade: once unsafe behaviors are internalized, earlier-stage defenses (input filtering, runtime checks) must work harder, and incident recovery becomes costlier.

\paragraph{\textit{Multi-agent: emergent hazards and adversarial coordination}}
In multi-agent settings, safety failures can emerge from coordination dynamics rather than single-agent errors: agents may collude to bypass constraints, amplify misinformation through mutual reinforcement, or trigger negative externalities via competitive equilibria (e.g., resource exhaustion or denial-of-service through uncoordinated tool calls)~\citep{Dafoe2020Cooperative,liu2025autobnbragenhancingmultiagentincident}. Communication channels also create new attack surfaces (e.g., one compromised agent relays injected instructions to others), turning localized perception attacks into system-wide action cascades.

\paragraph{\textit{Long-horizon: compounding error, delayed side effects, and value drift}}
Long-horizon trajectories magnify small errors: minor perception noise or planning miscalibration can compound over many steps into severe divergence, while harm may be delayed and hard to attribute (e.g., gradual financial loss, creeping policy violations, or subtle safety boundary erosion)~\citep{Sculley2015HiddenDebt}. Long-horizon agents also face statefulness risks (memory accumulation, stale goals, and context truncation), where outdated assumptions persist and contaminate subsequent planning and learning.

\subsubsection{Methods}
Mitigations should be stage-aligned along the Perceive$\rightarrow$Plan$\rightarrow$Act$\rightarrow$Reflect$\rightarrow$Learn lifecycle.

\paragraph{\textit{Perceive: robustness training and OOD detection}}
Data augmentation expands training coverage with perturbations, ambiguities, attacks, and edge cases~\citep{yadgaroff2024improvinggeneralizationgameagents,data_agmententation_survey,tanjim2025detectingambiguitiesguidequery,liu2025autobnbragenhancingmultiagentincident,wen2025generativeaidataaugmentation,xue2025resamplerobustdataaugmentation}. 
Adversarial training hardens agents against evolving attacks (from prompt-level manipulations to adaptive strategies)~\citep{shi2025lessonsdefendinggeminiindirect,wang2025adversarialreinforcementlearninglarge,adversarial_prompt_injection,wallace2024instructionhierarchytrainingllms,HarmBench}. 
At deployment time, OOD detection flags contexts outside validated regimes and triggers conservative fallbacks (e.g., read-only mode or human approval)~\citep{10.5555/3545946.3598721}.
Perception hardening in agentic systems additionally benefits from input provenance and sanitization, enforcing strict separation between trusted system prompts and untrusted retrieved content, strip or quarantine suspected instructions in retrieved text and applying policy-aware content filters before information enters memory or planning~\citep{OWASPPromptInjection,Paverd2025IndirectPI}. For distribution shifts, distributionally robust objectives (e.g., group DRO) can improve worst-slice reliability at the cost of average-case utility and additional tuning complexity~\citep{Sagawa2020GroupDRO}. Finally, uncertainty-aware perception (ensembles, selective prediction, and calibrated confidence) enables ``fail-closed'' behavior, but may increase latency and reduce autonomy due to more frequent abstentions.

\paragraph{\textit{Plan: constrained optimization for safety}}
Planning-stage safety is commonly modeled via constrained decision making (e.g., CMDPs)~\citep{Altman1999}. Safe RL balances return maximization with risk avoidance~\citep{ijcai2024p913,kushwaha2025surveysafereinforcementlearning}, using constrained policy optimization such as CPO~\citep{Achiam2017CPO}. Risk-sensitive objectives (e.g., CVaR) further emphasize catastrophic tail-risk reduction during plan optimization~\citep{CVaR,policy_gradient_cvar}.
For LLM planners, a complementary line is constraint specification and enforcement in natural language: constitutional or rule-based constraints translate norms into reusable critique-and-revision procedures, improving compliance without enumerating all unsafe actions~\citep{bai2022constitutional}. However, constraints can be gamed under misspecification; thus, specification-aware planning (e.g., interpreting rewards as uncertain and planning conservatively) helps mitigate reward hacking and negative side effects~\citep{HadfieldMenell2017InverseRewardDesign,Amodei2016Concrete}. In practice, safer planning often mixes (i) receding-horizon/MPC-style replanning, (ii) subgoal verification (checkable intermediate constraints), (iii) risk budgeting (explicit limits on tool calls, cost, or exposure). These methods typically trade autonomy and optimality for predictability and tighter safety margins, and may require stronger world models or simulators to be effective.

\paragraph{\textit{Act: runtime enforcement and shielding}}
Execution-time safeguards prevent unsafe plans from becoming high-impact actions. Shields and supervisory controllers can block or redirect unsafe actions during interaction with environments or tools~\citep{10.5555/3504035.3504361,10.1145/3715958,10.5555/3294771.3294858}. In tool-using settings, this typically combines pre-execution checks (permissions, budgets, and policy compliance) with post-execution validation and rollback hooks.
Modern agent deployments increasingly adopt guardrails as a layered runtime stack: (i) least-privilege tool access (scoped credentials, per-tool permissions, and spending caps), (ii) sandboxing and transactionality (dry-run, staged execution, and compensating actions), (iii) content filtering for both inputs and generated tool arguments~\citep{zheng2025webguardbuildinggeneralizableguardrail,li2025stacinnocenttoolsform}. Human-in-the-loop oversight (approval gates for irreversible actions, escalation on anomalies) is a pragmatic safety control in high-stakes settings~\citep{NIST2023RMF}, but increases cost and reduces responsiveness. A recurring trade-off is utility vs.\ enforceability: stricter runtime policies reduce catastrophic risk but may block benign edge cases, motivating adaptive policies that tighten under uncertainty or suspected attack.

\paragraph{\textit{Reflect: validation via simulation and formal checks}}
To reduce silent failures, agents (or their operators) can validate plans and anticipated outcomes through simulation-based testing in controlled environments, especially for long-horizon behaviors and rare events~\citep{10.1109/ICSE55347.2025.00223}. Formal verification can certify well-specified components under explicit assumptions~\citep{4567924,kumarappan2025leanagent,ahuja2025improver,ijcai2024p12,allegrini2025formalizingsafetysecurityfunctional}, although end-to-end verification of LLM-based agents remains challenging due to probabilistic generation~\citep{Abou_Ali_2025,SAPKOTA2026103599}.
Reflection is strengthened by adversarial evaluation and red teaming: automated prompt attacks, tool misuse suites, and policy violation benchmarks can be integrated into continuous integration (CI) pipelines to catch regressions before deployment~\citep{HarmBench,10.1145/3715958}. For agentic systems, trace auditing (immutable tool logs, provenance for retrieved evidence, and anomaly detectors over action sequences) helps counter evaluator spoofing and supports post-incident forensics. The main limitation is coverage: simulation and red teaming cannot fully enumerate open-world contexts, so reflective checks should be paired with conservative runtime monitors that trigger rollback, safe mode, or human review when novel hazards are detected.

\paragraph{\textit{Learn: safe updates with regression gating}}
Learning-stage safety requires cautious updates: maintain regression packs built from stress failures and
incidents, continuously monitor key safety/violation metrics, and gate releases to prevent regressions
introduced by model, prompt, memory, or tool-policy updates~\citep{Sculley2015HiddenDebt,Breck2017MLTestScore}.
In deployment, staged rollouts (e.g., shadow/canary) with automated acceptance/rollback criteria provide
a practical mechanism to validate updates under real traffic without amplifying unsafe behaviors~\citep{Burns2018SpinnakerCD}.
When policy updates are learned from logged interactions, conservative safe policy improvement offers a
principled way to avoid degrading a baseline policy under uncertainty~\citep{Laroche2019SPIBB}.
For agentic systems, ``learning'' often includes non-parameter updates (prompt templates, tool schemas, memory policies, and retrieval corpora). Safe learning therefore requires versioned artifacts and incident-driven datasets: add failure cases (prompt injections, jailbreaks, and tool glitches) to regression suites and retrain/retune defenses accordingly~\citep{li2025commercialllmagentsvulnerable,shi2025lessonsdefendinggeminiindirect}. Where online adaptation is necessary, safe exploration and conservative updates reduce the probability of high-cost trials~\citep{Amodei2016Concrete,kushwaha2025surveysafereinforcementlearning}. The trade-off is slower iteration: stronger gating and conservative improvement reduce harm but can delay beneficial capability updates, motivating risk-tiered release processes (stricter for high-impact tools, looser for read-only assistants).

\paragraph{\textit{Multi-agent: coordination constraints and system-level oversight}}
Multi-agent safety benefits from protocol-level constraints: restrict message formats, authenticate agent identities, and enforce shared safety policies on inter-agent communication to prevent the propagation of injected instructions. At the system level, centralized oversight (a monitor or ``governor'' that enforces global budgets, conflict checks, and stop conditions) helps prevent emergent cascades such as redundant tool calls and resource exhaustion. However, stronger coordination controls can reduce parallelism and autonomy, and may introduce single points of failure; practical designs often mix decentralization for performance with centralized ``circuit breakers'' for safety.

\paragraph{\textit{Long-horizon: checkpointing, risk budgets, and hierarchical guarantees}}
Long-horizon safety is improved by hierarchical decomposition (plans as verifiable subgoals), periodic checkpoints that require re-validation of assumptions, and risk budgets that bound cumulative exposure across steps (e.g., number of tool calls, spending, external writes). Receding-horizon replanning reduces compounding error, while explicit stop/interrupt policies prevent runaway execution when uncertainty grows. These approaches increase monitoring overhead and may require richer state tracking, but they directly target the long-horizon cascade that makes agentic safety uniquely challenging.

\paragraph{\textit{Defense-in-depth: complementary, not substitutable}}
The mitigations described above are intended as complementary layers in a defense-in-depth strategy, not as substitutes that designers can choose between. A poisoning attack at the perceive stage cannot be fully neutralized by guardrails or shielding at act time alone; conversely, even robust perception does not eliminate the need for planning constraints or runtime enforcement. In practice, trustworthy agent deployment requires protections at most or all stages simultaneously because failures at any single point can cascade through the remaining pipeline. Each layer addresses a different failure mode and provides residual coverage for gaps in other layers.
\textbf{Table~\ref{tab:safety_methods_comparison}} summarizes representative stage-aligned mitigations and their individual trade-offs; the ``limitations'' column should be read as motivating protection at other stages rather than as reasons to omit a layer. Cross-cutting controls (logging, provenance, red teaming, and HITL escalation) are most effective when integrated end to end rather than as isolated patches.

\begin{table*}[!hb]
\caption{Stage-aligned safety and robustness mitigations for agentic systems (non-exhaustive).\vspace{4pt}}
\label{tab:safety_methods_comparison}
\renewcommand{\arraystretch}{1.5}
\renewcommand{\tabularxcolumn}[1]{>{\raggedright\arraybackslash}p{#1}}
\begin{tabularx}{\textwidth}[t]{@{}
 |>{\raggedright\arraybackslash}p{2.0cm}
 |>{\raggedright\arraybackslash}X 
 |>{\raggedright\arraybackslash}X
 |>{\raggedright\arraybackslash}X
  @{}|}
\hline
\textbf{Stage} & \textbf{Representative techniques} & \textbf{Primary benefit} & \textbf{Main limitation/trade-off} \\
\hline
Perceive & Augmentation, adversarial training, provenance/sanitization, OOD detection, DRO objectives & Reduces attack/sift sensitivity; enables abstention or safe mode & Coverage gaps; compute/latency; may reduce average performance \\ \hline
Plan & CMDPs/CPO, CVaR, constitutional constraints, conservative planning, safe exploration & Optimizes under explicit safety constraints; reduces tail risk & Requires formalizable constraints; may be overly conservative \\ \hline
Act & Shields, guardrails, least-privilege tools, sandboxing, transactional execution, HITL approval & Prevents high-impact failures even with flawed plans & Reduced autonomy; false positives; operational overhead \\ \hline
Reflect & Simulation, red teaming, trace auditing, formal checks on components & Detects silent failures; supports debugging and regression tests & Incomplete coverage; brittle assumptions in simulators/verifiers \\ \hline
Learn & Regression gating, canary rollout, conservative improvement, incident-driven updates & Prevents safety regressions and drift over time & Slower iteration; monitoring and dataset maintenance cost \\ \hline
Multi-agent & Communication constraints, authentication, centralized governors, global budgets & Controls emergent cascades and collusion pathways & Coordination overhead; potential single points of failure \\ \hline
Long-horizon & Checkpointing, risk budgets, receding-horizon replanning, interruptibility & Limits compounding error and delayed side effects & More frequent interventions; increased monitoring requirements \\
\hline
\end{tabularx}

\vspace{4pt}
\noindent\footnotesize{Abbreviations: OOD: out-of-distribution; DRO: distributionally robust optimization; RL: reinforcement learning; CMDPs: constrained Markov decision processes; CPO: constrained policy optimization; CVaR: conditional value-at-risk; HITL: human-in-the-loop.}
~~\newpage
~~\newpage
~~\newpage
~~\newpage
\end{table*}

Evaluation metrics and release-gating practices are consolidated in Section~\ref{section:eval}.


\subsection{Privacy and system security}\label{subsection:privacy}

This section defines Privacy and System Security in the context of agentic AI, identifies key risks across the agent workflow---including data leakage, tool-mediated exfiltration, and supply chain vulnerabilities---and reviews stage-targeted mitigation strategies.

\subsubsection{Definition}
Following recent analyses~\citep{mireshghallah2025position,du2025beyond, mireshghallah2023can, shao2024privacylens, bagdasarian2024airgapagent, ghalebikesabi2024operationalizing, cheng2024ci}, we define privacy in agentic AI as
protecting any user- or environment-derived information that (i) directly identifies an individual (e.g., names, contact
information), (ii) enables sensitive attribute inference (e.g., demographics, preferences, or behavioral patterns), (iii) can be reconstructed/regurgitated through an agent's internal representations, memory, tool outputs, or interaction traces.
Crucially, privacy risks extend beyond memorization to inference-based leakage, accumulation over long-horizon interactions,
unintended disclosure during tool execution, and cross-agent propagation~\citep{mireshghallah2025position,yang2025mla}.

Complementarily, we adopt the definition of AI security as ``the property of an AI system to remain resilient against intentional attacks on its data, algorithms, or operations, preserving its confidentiality, integrity, and availability in the presence of adversarial actors''~\citep{lin2025aisafetysecurity}. Unlike safety, which focuses on unintentional failures, security specifically defends against malicious exploitation.
Formally, privacy protection is often expressed via indistinguishability guarantees. Under differential privacy (DP),
a randomized mechanism $\mathcal{M}$ satisfies $(\varepsilon,\delta)$-DP if for any neighboring datasets $D,D'$ (differing in one record)
and any measurable set of outputs $S$, $\Pr[\mathcal{M}(D)\in S]\le e^{\varepsilon}\Pr[\mathcal{M}(D')\in S]+\delta$~\citep{dwork2006dp,dwork2014dp},
where $\Pr[\cdot]$ denotes the probability over the randomness of $\mathcal{M}$.
DP clarifies a privacy--utility trade-off via $\varepsilon$ (privacy budget) and supports principled accounting under repeated queries.
Complementary, older anonymity notions such as \textit{$k$-anonymity} require each released record to be indistinguishable from at least
$k-1$ others w.r.t.\ quasi-identifiers~\citep{sweeney2002kanonymity}, with known limitations under background knowledge and linkage attacks
(e.g., motivating $l$-diversity and $t$-closeness)~\citep{machanavajjhala2007ldiversity,li2007tcloseness}.

Agentic AI introduces a practical distinction between static privacy and dynamic privacy. Static privacy concerns training-time
memorization and post-training extraction (e.g., the regurgitation of training or logged interaction data). Dynamic privacy concerns runtime information flows across the Perceive$\rightarrow$Plan$\rightarrow$Act$\rightarrow$Reflect$\rightarrow$Learn loop, where privacy loss can
occur even without memorization: repeated interactions enable attribute inference, tool calls serialize sensitive context into external systems,
and intermediate traces create new disclosure surfaces~\citep{mireshghallah2025position,du2025beyond}.

To capture context-dependent expectations, we also adopt the contextual integrity (CI) perspective: privacy is preserved when information flows conform to appropriate social norms (who shares what about whom, with whom, and under which transmission principles)~\citep{nissenbaum2004contextualintegrity,mireshghallah2023can}.
CI is particularly relevant for agents because they routinely transform context (summarize, retrieve, and act via tools), making it easy to violate
norms even when no explicit identifier is leaked (e.g., disclosing a user's health condition to an unrelated tool)~\citep{shao2024privacylens,ghalebikesabi2024operationalizing,cheng2024ci}.

We define system security for agentic AI as preserving the confidentiality, integrity, and availability of the agent's
tooling and execution environment (e.g., credentials, APIs, sandboxes, and protocols), and resisting attacks that induce
unsafe actions such as unauthorized tool use, code execution, or protocol manipulation~\citep{zhang2025privweb,du2025beyond}.
Concretely, confidentiality covers secrets (API keys, tokens), private data, and proprietary prompts/policies; integrity covers
the correctness of retrieved evidence, tool outputs, and action arguments; and availability covers service continuity under abuse (e.g., tool-call floods, resource exhaustion). Threat modeling (assets, entry points, attacker capabilities, and failure impact) provides a systematic way to reason about agent security across components and stages~\citep{shostack2014threatmodeling}.

Finally, agentic deployments raise supply chain security concerns: models, tool plugins/APIs, retrieval indices, prompt templates,
and evaluation/guardrail services may come from distinct providers with independent update cycles. Compromise can occur upstream (malicious
dependency, poisoned tool update, or trojaned plugin) and propagate downstream into agent behavior. Practical controls include Software Bills of Materials (SBOMs) and signed,
verifiable artifacts to support provenance and rapid incident response~\citep{spdx2021sbom,cyclonedx2024sbom,openssf2023slsa,sigstore2022}.
To analyze threats systematically, we adopt a component-based view (Input, Memory, Tool/Execution, and Communication/Protocol)
and map it to the agent workflow below~\citep{yang2025mla,zhang2025searching,zhang2025privweb,du2025beyond}.

While Section~\ref{subsection:safety}~ emphasizes behavioral harm and robustness, this subsection focuses on data protection and system integrity: many attacks (e.g., prompt injection) matter here not only because they change behavior, but because they enable
credential theft, unauthorized data access, and exfiltration through legitimate tool channels.

\subsubsection{Risks}
\paragraph{\textit{Perceive: prompt injection, multimodal inference, and obfuscated inputs}}
The input channel ingests user prompts, multimodal observations, and external content (web pages, emails, and GUIs).
Direct/indirect prompt injection (DPI/IPI) can coerce agents to disclose private information (e.g., prior conversations or
implicit user attributes) or to bypass policies~\citep{zhang2025searching}. Multimodal inference attacks can extract demographic
or situational cues from images/screen context~\citep{yang2025mla}. Obfuscated input attacks can evade static filters by evolving
through multi-step perturbations, increasing both privacy leakage and security compromise risk~\citep{zhang2025searching}.
Additional perceive-stage risks include social engineering inputs (phishing-style prompts that elicit secrets or induce the agent to
paste credentials) and retrieval layer poisoning where attacker-controlled documents are designed to be retrieved and to carry hidden
exfiltration instructions. A real-world pattern is that untrusted content can trigger unintended behaviors even without explicit user interaction
(e.g., ``zero-click'' indirect prompt injection in enterprise copilots)~\citep{Lakshmanan2025EchoLeak,Paverd2025IndirectPI}. These inputs are
high-leverage because they seed downstream stages: once a malicious instruction is treated as trusted context, it can shape planning, tool use,
and what gets stored into memory, creating persistent privacy loss.

\paragraph{\textit{Plan: privacy-unsafe memory use and long-horizon attribute aggregation}}
Planning often relies on memory (conversation history, episodic buffers, and long-term stores), which can accumulate sensitive
data across sessions. Two privacy risks arise: (i) regurgitation/reconstruction, where previously seen private content
resurfaces later~\citep{mireshghallah2025position}; (ii) attribute aggregation, where multiple low-sensitivity fragments
combine into high-sensitivity inferences over time~\citep{yang2025mla}. From a security perspective, memory can be poisoned
(e.g., delayed triggers/backdoors) so that future plans are steered toward unsafe actions or disclosure.
Planning introduces contextual integrity violations as a distinct privacy failure mode: even if each memory item is individually
non-sensitive, the planner may route information to an inappropriate recipient (e.g., sending a user's medical context to a generic web tool)
because it optimizes task success over norm compliance~\citep{nissenbaum2004contextualintegrity,shao2024privacylens}. Another risk is cross-session re-identification: stable memory identifiers, embeddings, or long-lived profiles can allow linking across sessions/users,
especially under insider threats (operators) or compromised storage backends. These plan-stage risks cascade into act: once the plan encodes a
privacy-unsafe tool strategy (e.g., ``search using full email thread''), subsequent execution can leak at scale.

\paragraph{\textit{Act: tool-mediated leakage and execution-layer exploitation}}
Tool use (browsers, code interpreters, databases, and GUI actuators) creates powerful but high-risk pathways. Privacy leakage can occur
when sensitive text is sent to tools (e.g., via RAG queries or browser autofill), when browsing traces expose identifiers, or when
agent-generated queries inadvertently encode private attributes~\citep{zhang2025privweb,du2025beyond}. Tool outputs may contain latent
metadata that the agent later exposes. Security threats arise when unsafe tool calls execute malicious code, trigger SQL injections,
or interact with untrusted web content, enabling the exploitation of system-level vulnerabilities~\citep{zhang2025privweb}.
Agentic AI adds execution-time threat classes uncommon in stateless LLMs: (i) credential theft through tool access (keys/tokens exposed
via prompts, logs, or tool outputs), (ii) exfiltration via authorized channels, where the agent uses legitimate permissions (email, cloud
drive, and HTTP requests) as a covert export path, (iii) side-channel leakage through tool timing, error codes, or response sizes that
encode sensitive states (e.g., inferring whether a record exists from query latency). Tool-chain attacks (benign tools combined into an exfiltration
pipeline) further expand blast radius and can convert a single injected instruction into sustained leakage across many steps~\citep{li2025stacinnocenttoolsform}.
These failures propagate into reflect/learn: traces of tool calls and outputs become part of the agent's evidence base and may be retained.

\paragraph{\textit{Reflect: cross-component propagation through traces and protocols}}
In multi-agent or user--agent--tool ecosystems, intermediate messages, rationales, and negotiation traces can leak sensitive
attributes, and disclosures can propagate nonlinearly across agents~\citep{yang2025mla}. Protocol-level failures (tampering,
impersonation, or instruction injection) can further amplify leakage and induce coordinated unsafe behavior.
Reflection adds two privacy-specific hazards: trace over-collection and rationale leakage. Rich traces (chain-of-thought-like
rationales, tool logs, and screenshots) are valuable for debugging and accountability, but they may also store identifiers, credentials, or sensitive
inferences that were never meant to persist. Moreover, protocol attacks such as replay/downgrade (e.g., forcing weaker authentication or
reusing prior authorization messages) can compromise integrity and spread injected instructions across components. Once leaked into shared logs or
cross-agent channels, sensitive information is hard to retract, and may be repeatedly resurfaced during later planning.

\paragraph{\textit{Learn: drift and persistence of privacy risk across updates}}
Long-horizon use and iterative updates can make privacy risks persistent: what was once benign context may become identifying when
combined with new data sources or policies, and training-time or memory-time retention can reintroduce leakage even after patches
\citep{mireshghallah2025position,du2025beyond}. Without explicit retention limits and regression checks, privacy/security properties
may silently degrade over releases.
Learning introduces additional adversarial threat models: insider threats (malicious or careless data curators), compromised
tools/APIs that backdoor training logs, and supply chain poisoning where updated guardrails or retrieval indices embed exfiltration logic.
Even without explicit fine-tuning, non-parametric updates (prompt templates, memory policies, and tool registries) can re-open previously closed
leakage channels. Because agents operate continually, privacy debt can accumulate: small retention or policy changes can shift the effective privacy
budget over time (DP-style composition) and invalidate earlier assurances~\citep{dwork2014dp}.

\paragraph{\textit{Multi-agent privacy and security: shared context, privilege escalation, and collusive exfiltration}}
When multiple agents collaborate, shared memory, delegation, and message passing create new disclosure pathways: one agent may request context from another beyond ``need-to-know'', or forward sensitive content into a broader channel. Security risks include privilege escalation (via delegation to a higher-privileged agent), impersonation (spoofed agent identity), and collusive exfiltration where agents coordinate to leak data through seemingly benign tool calls. Such failures can be rapid and nonlinear: a single compromised agent can contaminate
others' context and plans, amplifying leakage beyond what any single-agent threat model predicts~\citep{yang2025mla}.\newpage

\paragraph{\textit{Credential and secret exposure: long-lived tokens, implicit capture, and recovery gaps}}
Agents often handle API keys, OAuth tokens, passwords, and session cookies. Secrets can leak through prompts (users paste keys), tool outputs
(stack traces), memory (cached credentials), screenshots/GUI context, or logs. A distinctive agentic risk is secret reuse across stages:
once a secret is captured during perceive/act, it can persist in memory, appear in reflection traces, and be replayed by future plans---turning a
single exposure into sustained compromise. Recovery gaps (delayed key rotation, unclear ownership of secrets) further increase the time to mitigate.

\subsubsection{Methods}
\paragraph{\textit{Perceive: sanitization, detection, and minimization at intake}}
Defenses include input sanitization (structured validation; LLM-as-a-judge filtering) and adversarial search-based detection for evolving prompt injections~\citep{zhang2025searching}. Privacy-aware prompting and minimization policies reduce unnecessary sensitive content ingestion, lowering downstream leakage surfaces~\citep{du2025beyond}.
Perceive-stage security benefits from zero-trust intake: treat all external content (web/email/docs/UI text) as untrusted, enforce strict separation between instructions and data, and apply provenance checks before content can influence tool permissions or memory writes.
Practical mechanisms include instruction-hierarchy policies, input allowlists (schemas for tool arguments), and Data Loss Prevention (DLP)-style detectors that flag Personally Identifiable Information (PII)
and secrets at the boundary (before they enter retrieval indices or logs). The trade-off is utility and latency: aggressive filters can block benign
inputs and require costly multi-pass detection, motivating risk-adaptive policies that tighten under suspected attack~\citep{nist2020zerotrust}.

\paragraph{\textit{Plan: memory governance and least-privilege planning}}
Mitigations include memory auditing and sensitivity-aware access policies, automated deletion/expiration to enforce data minimization,
and controls that limit what planning can retrieve or retain when privacy risk is high~\citep{du2025beyond}. Differential privacy and
related techniques can reduce the influence of individual memory entries and limit reconstruction risk~\citep{mireshghallah2025position}.
Planning-stage privacy requires memory governance as policy: tag memory entries with sensitivity (PII, credentials, and regulated data), enforce  purpose limitation (only retrieve data needed for the current task), and adopt contextual integrity checks that validate whether a planned information flow is appropriate for the recipient/tool~\citep{nissenbaum2004contextualintegrity,cheng2024ci}. DP-style
mechanisms can be applied not only to training, but also to telemetry and analytics derived from user interactions (bounded contribution per user),
though DP often reduces personalization fidelity and requires careful privacy-budget accounting under repeated queries~\citep{dwork2014dp}.
In practice, least-privilege planning couples plan generation with permission-aware tool selection (choos the minimum-scope tool that can complete the step), trading some autonomy for a smaller blast radius when a plan is compromised.

\paragraph{\textit{Act: sandboxing, runtime policy enforcement, and post-execution redaction}}
Effective defenses include sandboxed execution, runtime policy enforcement governing permitted tool actions, command-level verification,
and post-execution redaction using content-aware minimization techniques~\citep{zhang2025privweb,du2025beyond}. Credential handling should
follow least-privilege principles, and tool scopes should be tightly constrained to reduce blast radius.
Execution-layer defenses increasingly resemble enterprise security stacks. Policy-as-code gates (pre-call authorization, per-tool budgets,
rate limits, and data-classification constraints) reduce unauthorized actions; EDR-style monitoring instruments tool calls, filesystem/network
events, and anomalous sequences to detect exfiltration attempts in real time. Data loss prevention (DLP) can be adapted to agentic settings
by scanning outgoing tool arguments (queries, HTTP payloads, and emails) and blocking or redacting sensitive spans before transmission.
For privacy-preserving computation, sensitive tasks can be moved into protected execution environments (e.g., TEEs) or distributed protocols
(e.g., federated learning for personalization; secure MPC/HE for cross-organization analytics)~\citep{yang2025mla,du2025beyond}.
These defenses trade off performance (overhead, latency) and developer complexity (policy engineering, instrumentation).

\paragraph{\textit{Reflect: protocol protection and trace-based incident handling}}
Protocol-aware defenses include encrypted communication, authenticated message channels, and tamper-evident logging/ledger-style
verification to prevent message tampering and support investigations. Consistency/debate-style checks can reduce the impact of adversarial
manipulation in multi-agent settings~\citep{yang2025mla}. Conformance checking and privacy-incident schemas help standardize violation
detection and response~\citep{mireshghallah2025position}.
Reflection should implement privacy-aware observability: collect only what is necessary (minimize trace content), segregate sensitive logs,
and apply retention limits to rationale and tool traces. For integrity, tamper-evident logs and authenticated provenance enable post-incident
attribution and replay-based debugging without trusting any single component. Operationally, incident handling should include automatic secret
revocation (key rotation), quarantine of contaminated memory entries, and backfilling of regression tests from discovered leakage traces.
The trade-off is reduced debuggability: redacted traces may hinder root cause analysis, motivating tiered access (secure audit mode under strict
controls) rather than always-on full-fidelity logging.

\paragraph{\textit{Learn: retention limits and privacy/security regression gating}}
To prevent the persistence of privacy risk, learning and deployment updates should be gated by retention and leakage controls (e.g., memory
expiration, deletion policies, and DP-based storage), and by regression packs that replay known attacks and leakage cases across releases
\citep{du2025beyond,mireshghallah2025position}. Evaluation metrics and benchmark suites are consolidated in Section~\ref{section:eval}.
Beyond model updates, secure update pipelines {\addfontfeature{LetterSpace=1.5}should cover tool registries, prompt templates,}\newpage guardrail services, and retrieval indices.
Supply chain controls (SBOMs, signed artifacts, dependency provenance, and SLSA-style build levels) help prevent the silent introduction of malicious
components and accelerate rollback when compromises are found~\citep{spdx2021sbom,openssf2023slsa,sigstore2022}. Where possible, privacy
regression tests should include CI-based scenarios (norm violations), not only leakage string-matching, to catch inappropriate flows that do not
surface as explicit PII~\citep{cheng2024ci,shao2024privacylens}.

\paragraph{\textit{Multi-agent privacy and security: shared-policy enforcement and compartmentalization}}
Mitigations for multi-agent systems emphasize compartmentalization: per-agent memory namespaces, explicit contracts for what can be
requested/shared, and authenticated channels with role- and scope-based access control. Global governors can enforce system-wide budgets
(e.g., total external sends) and detect cross-agent exfiltration patterns. Debate/consistency checks help, but must be paired with protocol security
to prevent spoofing or replay in agent-to-agent communication~\citep{yang2025mla}.

\paragraph{\textit{Credential and secret management: ephemerality, vaulting, and safe interfaces}}
Agents should never treat secrets as normal text. Practical designs include (i) ephemeral credentials (short-lived tokens bound to a single
task), (ii) vault-backed retrieval (agents request scoped secrets via secure APIs rather than storing them), (iii) continuous rotation and
revocation hooks, (iv) secret scanning/redaction in prompts, memory writes, and logs. This reduces catastrophic persistence: even if an agent is
prompt-injected, the attacker faces a narrow time window and limited scopes. The trade-off is engineering overhead and occasional usability loss
(e.g., more re-authentication events).

\paragraph{\textit{Summary and practical trade-offs}}
Component- and stage-aligned defenses are complementary: intake filtering and minimization reduce exposure, memory governance limits
retention, and sandboxing/protocol protections constrain propagation~\citep{du2025beyond,zhang2025privweb,yang2025mla}. However, stronger
controls may reduce utility (less context/memory), add engineering overhead (sandboxing, authenticated protocols), and introduce trade-offs
between privacy guarantees and performance (e.g., DP or aggressive deletion)~\citep{mireshghallah2025position,du2025beyond}.
A practical takeaway is to treat privacy/security as end-to-end information-flow control rather than isolated filters: vulnerabilities are often
composed across components (\textbf{Table~\ref{tab:privacy_threat_model}}). Compared to the safety mitigations in Section~\ref{subsection:safety}, controls here
are frequently operational (zero trust, secret management, provenance, and DLP/EDR instrumentation) and must be maintained continuously as tools
and supply chains evolve. Evaluation should therefore combine outcome measures (leakage counts, attribute inference accuracy) with process signals
(trace minimization, permission conformance, and provenance coverage), as consolidated in Section~\ref{section:eval}.\newpage

\begin{table*}[!ht]
\caption{Threat model matrix for privacy and system security in agentic AI (non-exhaustive).\vspace{4pt}}
\label{tab:privacy_threat_model}
\renewcommand{\arraystretch}{1.5}
\renewcommand{\tabularxcolumn}[1]{>{\raggedright\arraybackslash}p{#1}}
\rowcolors{1}{white}{white}  
\begin{tabularx}{\linewidth}[t]{@{} |>{\raggedright\arraybackslash}p{2.3cm} |>{\raggedright\arraybackslash}p{3.2cm} |>{\raggedright\arraybackslash}X |>{\raggedright\arraybackslash}X @{}|}
\hline
\textbf{Component} & \textbf{Typical assets} & \textbf{Representative threats (CIA)} & \textbf{Agentic-specific vectors} \\
\hline
Input & Prompts, retrieved docs, GUI context & Confidentiality loss via injected disclosure; integrity loss via instruction/data confusion; availability loss via prompt flooding & Indirect prompt injection; zero-click triggers in untrusted content; multimodal attribute inference \\ \hline
Memory & Histories, long-term stores, embeddings & Regurgitation/reconstruction; memory poisoning/backdoors; unbounded retention & Long-horizon aggregation; cross-session re-identification; insider access to stored traces \\ \hline
Tool and execution 
 & API keys, browsers, DBs, code exec & Exfiltration through tool calls; unauthorized actions; SQL/code injection; DoS via tool spam & Exfiltration via authorized channels; tool chaining; side-channel leakage via timing/errors \\ \hline
Comm and protocol & Inter-agent msgs, RPC, logs & Spoofing/replay/tampering; trace leakage; channel disruption & Delegation-based privilege escalation; cross-agent propagation of injected instructions \\ \hline
Supply chain & Models, plugins, dependencies & Compromised updates; poisoned artifacts; provenance ambiguity & Malicious tool/plugin updates; poisoned retrieval indices; untrusted guardrail services \\
\hline
\end{tabularx}

\vspace{4pt}
\noindent\footnotesize{Abbreviations: CIA: confidentiality--integrity--availability; GUI: graphical user interface: API, application programming interface; DB: database; SQL: structured query language; DoS: denial of service; RPC: remote procedure call; DLP: data loss prevention.}
\end{table*}

\section{Consolidated metrics and benchmarks}
\label{section:eval}
This section consolidates evaluation content into a single ``evaluation hub'' for trustworthy agentic AI. Our two core dimensions---Safety and Robustness (Section~\ref{subsection:safety}) and Privacy and System Security (Section~\ref{subsection:privacy})---receive the primary focus, while metrics for other trust aspects (value alignment, transparency, fairness, and accountability) are included as supporting references. Unlike single-turn LLMs, agents must be evaluated on both outcomes and process: multi-step trajectories, tool calls, permission checks, and trace evidence. Metric choices and release gates should follow a risk-based framing (e.g., NIST AI RMF) \citep{NIST2023RMF}.
Concretely, evaluation should be treated as a system-level property: it is jointly determined by the base model, prompts, retrieval/memory policies, tool wrappers, runtime guards, and oversight interfaces. As a result, the same policy/model can exhibit different risk profiles under different tool scopes, logging granularity, or approval workflows. This motivates (i) explicit versioning of the full agent stack, (ii) standardized trace schemas for auditability, (iii) scenario-driven release gates that bound residual risk under the intended operational design domain (ODD) rather than reporting a single global score.

\subsection{Connecting evaluation to the agent lifecycle}
The metrics and benchmarks in this section are designed to map directly onto the five-stage workflow (Perceive--Plan--Act--Reflect--Learn) introduced in Section~\ref{sec:preliminaries-agentic}~ and the stage-aligned risks and mitigations discussed in Section~\ref{section:core_dimensions}.
Specifically:
Perceive-stage risks (data poisoning, adversarial inputs, and OOD inputs) are evaluated through stress test pass rates, adversarial robustness scores, and input filtering effectiveness.
Plan-stage risks (specification gaming, miscalibrated uncertainty) are captured by constraint violation rates (CVRs) and planning-quality metrics under OOD scenarios.
Act-stage risks (dangerous tool use, cascading failures) are measured through catastrophic event rates (CERs), tool-policy violation counters, and data leakage rates.
Reflect-stage quality (self-assessment fidelity, trace coverage) is measured by decision coverage rate (DCR) and compliance verification rate (CompVR).
Learn-stage regression (safety degradation after updates) is tested through regression packs, canary rollout acceptance criteria, and version-to-version comparison.
This explicit coupling ensures that evaluation is not a detached post hoc exercise but a diagnostic tool for identifying where in the lifecycle trustworthiness failures originate.

\subsection{Measurement principles}
\label{subsection:measure}
A trustworthy metric suite should satisfy: validity (targets the intended harm), reliability (stable across seeds/evaluators),
reproducibility (environment/tool/version control), coverage (OOD/adversarial/long-tail/long-horizon stress),
and statistical power.
It should report means with 95\% confidence intervals and stratify by scenario families
(in-distribution vs.\ OOD vs.\ adversarial). It should keep stress test results separate from in-distribution
results (it should not average them). If using LLM-as-a-judge, document the judge model/prompt and
report agreement or calibration against human labels when possible \citep{zheng2024judging,human_affected_by_AI}.

The subsequent paragraphs delineate the conceptual dimensions of these criteria, after which we provide a detailed description of the corresponding metrics in Section~\ref{subsection:metric_dictionary}.

\subsubsection{Outcome vs.\ process evaluation (agents require both)}
Episode-level outcome metrics (e.g., SR, CER; see Section~\ref{subsection:metric_dictionary}~ for all abbreviations) capture what happened, but not how it happened. For agents, the process is often the primary risk channel: unsafe intermediate tool calls, permission bypasses, or evidence-free planning can occur even when the final answer looks correct. Therefore, release gates should pair outcome metrics with process evidence (CVR, DCR/Trace-Coverage, and CompVR) to ensure the system is not ``getting lucky'' or relying on brittle, unverifiable reasoning. This is especially important when agents operate with intermittent oversight (e.g., batched approvals) and long-horizon autonomy \citep{NIST2023RMF}.

\subsubsection{Trajectory-level vs.\ step-level metrics (what to measure when)}
Step-level metrics (CVR, CompVR, tool-policy violation counters, and context leakage counts) are most diagnostic for debugging and for enforcing hard constraints that must hold at every step (e.g., ``never exfiltrate secrets'', ``never execute unapproved code''). Trajectory-level metrics (SR, CER, SafetyScore, and Leakage/Regurgitation) are most appropriate for release decisions in high-risk domains, where the unit of harm is typically an episode (e.g., a completed workflow, a driving segment, or a clinical recommendation). Practically, we recommend reporting both: (i) step-level violation histograms (where violations occur, which tools/actions trigger them), (ii) episode-level aggregates (how often any violation occurs per episode), because long-horizon systems can appear safe on averages while still exhibiting rare but catastrophic chains.

\subsubsection{Long-horizon evaluation (compounding errors and delayed consequences)}
Long-horizon agents face compounding errors: small early mistakes in retrieval, planning, or tool execution can propagate into later stages and only manifest after many steps (e.g., incorrect state updates leading to unsafe actions). Consequently, evaluations should include (i) horizon-stress tests that scale trajectory length and branching, (ii) delayed-effect probes where harmful consequences arise only after intermediate actions, (iii) tail-risk reporting (e.g., CER/CVaR-style summaries) rather than relying solely on mean performance \citep{gao2023scaling}. When feasible, they should include ``time to failure''-style statistics (steps until first violation) to differentiate agents that fail immediately from agents that drift into unsafe regimes.

\subsubsection{Multi-agent evaluation (attribution, emergence, and coordination)}
When multiple agents interact, risks can emerge from coordination dynamics (e.g., collusion, information amplification, and cascading leakage) that are absent in single-agent settings \citep{Dafoe2020Cooperative,Zhang2021MARLSurvey}. Evaluation should therefore track: (i) attribution (which agent/tool/message caused the violation), (ii) coordination quality (task completion under shared constraints, deadlock rate, negotiation stability), (iii) cross-agent leakage (whether private attributes or credentials propagate across agents). Privacy benchmarks for multi-agent or multimodal GUI settings (e.g., MLA-Trust) are useful complements to single-agent leakage tests \citep{yang2025mla}.

\subsubsection{Judge reliability and adversarial robustness}
LLM-as-a-judge can reduce labeling costs but introduces failure modes: judges may be biased toward fluent rationales, may be sensitive to prompt wording, and can be adversarially influenced when evaluating jailbreak-like behaviors \citep{zheng2024judging,HarmBench}. For safety/security gates, treat judges as instruments that must be calibrated: report inter-judge agreement, run paired human audits on critical subsets, and include adversarial ``judge attacks'' (e.g., misleading justifications) to measure judge robustness. In high-stakes deployments, prefer deterministic checkers (schema validation, policy engines, formal constraints) for hard rules, and reserve LLM judges for soft risk scoring and triage.

\begin{table*}[!hb]
\caption{The trustworthy agent metric dictionary.
 We recommend reporting both episode-level outcomes (SR, CER) and step-level process signals (CVR, DCR) to capture risks that arise from intermediate tool use.\vspace{4pt}}
\label{tab:metric-dictionary}
\setlength{\tabcolsep}{3pt}
\renewcommand{\arraystretch}{1.5}
\renewcommand{\tabularxcolumn}[1]{>{\raggedright\arraybackslash}p{#1}}
\resizebox{\linewidth}{!}{
\begin{tabularx}{\linewidth}[t]{@{}
    |>{\raggedright\arraybackslash}p{2.35cm}
    |>{\raggedright\arraybackslash}p{2.05cm}
    |>{\raggedright\arraybackslash}X
    |>{\raggedright\arraybackslash}p{3.5cm}
|@{}}
\hline
\textbf{Metric} & \textbf{Type} & \textbf{Definition} & \textbf{\mbox{Benchmark examples}} \\
\hline
\multicolumn{4}{|l|}{Outcome and safety metrics} \\ \hline
SR & Episode Outcome & Success Rate: fraction of episodes fulfilling the user goal. & AgentBench, WebArena \\ \hline
SafetyScore & Graded Risk & Graded Likert scale (1--5) or binary labeling of trajectory safety. & SafeMind, R-Judge \\ \hline
CER & Episode Outcome & Catastrophic Event Rate: episodes with unacceptable harm. & Safety-Gym, Do-Not-Answer \\ \hline
CVR & Process Violation & Constraint Violation Rate: step-level hard policy breaches. & AgentDojo, SheetAgent-Safe \\ \hline
Stress-Pass & Robustness & Pass rate on OOD, adversarial, or long-horizon subsets. & HarmBench, AgentGuards \\ \hline
Over-Opt & Safety (Reward) & Reward-hacking proxy: degradation on true reward vs.\ proxy. & Scaling-Laws-Over-Opt \\
\hline
\multicolumn{4}{|l|}{Privacy, Security and Compliance Metrics} \\ \hline
Leakage/Reg-urgitation/ CLC & Privacy & Count/Rate of PII, internal context, or memorized content emitted. & MLA-Trust, PrivacyLens \\ \hline
Attr.-Inf. Acc. & Privacy (Inference) & Accuracy of inferring private attributes from agent outputs. & MLA-Trust \\ \hline
ASR/RR & Adversarial & Attack Success Rate/Refusal Rate under injection attacks. & AgentDojo, HarmBench \\ \hline
DCR (Trace-Coverage) & Auditability & Decision Coverage Rate: fraction of actions with complete traces. & AgentCompass \\ \hline
CompVR & Compliance & Compliance Verification Rate: fraction verifiable against rules. & Financial/Clinical Audits \\
\hline
\multicolumn{4}{|l|}{Supporting Dimensions} \\ \hline
PHR & Alignment & Preference Head-to-Head Win Rate vs.\ baseline. & AlpacaEval, MT-Bench \\ \hline
Honesty & Alignment & Factual consistency and hallucination rate (w/ tool evidence). & TruthfulQA, AgentHalu \\ \hline
TVD/Fairness-ASR & Fairness & Tool selection bias (TVD) and fairness under attack (Fairness-ASR). & BiasBuster, HarmBench \\ \hline
Fidelity/Stabil-ity/Complexity & Transparency & Explanation faithfulness, robustness, and simplicity (length/nodes). & XAI-Bench \\ \hline
ISR/A2I & Accountability & Interruption Success Rate/Alert-to-Intervention Latency. & HAII Benchmarks \\
\hline
\end{tabularx}
}

\vspace{4pt}
\noindent\footnotesize{Abbreviations: SR: success rate; CER: catastrophic event rate; CVR: constraint violation rate; OOD: out-of-distribution; PHR: preference win rate; ASR: attack success rate; RR: refusal rate; TVD: total variation distance; DCR: decision coverage rate; CompVR: compliance verification rate; ISR: interruption success rate; A2I: alert-to-intervention latency; HAII: human--AI interaction.}
\end{table*}

\subsection{Metric dictionary}
\label{subsection:metric_dictionary}
\textbf{Table~\ref{tab:metric-dictionary}} standardizes the abbreviations used throughout Section~3 and links each metric
to representative benchmark families.
\subsubsection{How to read the table}
We recommend reporting episode-level outcomes (e.g., SR, CER, and PHR) together with
step-/process-level violations and evidence (e.g., CVR, DCR/Trace-Coverage, and CompVR) because agent failures often arise from intermediate tool use, long-horizon planning, and hidden
policy breaks.

\subsubsection{Abbreviations (quick definitions)}
{\addfontfeature{LetterSpace=-2.0}SR: success rate per episode.
SafetyScore: graded safety risk score for an action/trajectory.
CER: catastrophic event rate (episode-level severe harm).
CVR: constraint violation rate (step-level hard-rule/policy/tool-permission violations).
Stress-Pass: pass rate on OOD/rare-event/long-horizon stress sets.
PHR: preference win rate (pairwise alignment).
Honesty: factual consistency/truthfulness under tool/RAG settings.
ASR/RR: attack success/refusal rate under adversarial prompts or social engineering.
Over-Opt: reward-hacking diagnosis via over-optimization curves \citep{gao2023scaling}.
Fidelity/Stability/Complexity: explanation faithfulness/robustness/simplicity.
TVD: tool-selection bias among functionally equivalent tools.
Fairness-ASR: fairness break rate under adversarial pressure.
DCR: decision coverage rate (trace completeness; a.k.a.\ Trace-Coverage).
CompVR: compliance verification rate (independent verifiability vs.\ standards/policies).
ISR/A2I: interruption success rate/alert-to-intervention latency.
Leakage/Regurgitation/Attr.-Inf.\ Acc./CLC: privacy leakage, memorized content resurfacing,
attribute inference accuracy, and context leakage count.}

\subsubsection{Dimension-specific evaluation notes}
{\addfontfeature{LetterSpace=-2.5}Safety and Robustness (Section~\ref{subsection:safety}): Our first core dimension. Combine SafetyScore/CVR/CER with Stress-Pass to separate in-distribution performance from tail risk.
Privacy and Security (Section~\ref{subsection:privacy}): Our second core dimension. Report leakage/regurgitation/attribute inference, plus Privacy-ASR under iterative attacks and within-interaction CLC.
Supporting metrics from other dimensions: For comprehensive evaluation, we also include metrics for value alignment (PHR, Honesty, and ASR/RR), transparency (Fidelity, Stability, and Complexity), fairness (TVD, Fairness-ASR), and accountability (DCR, CompVR, and ISR/A2I).}

\subsubsection{Benchmark suite descriptions (selected anchors)}
We briefly summarize representative suites referenced throughout this survey; these suites are best viewed as families that instantiate the metric dictionary under different environments and threat models.
SafeMind targets safety risks in embodied LLM agents, pairing graded safety assessment (SafetyScore) with event-level harm and violation tracking under long-horizon interaction \citep{chen2025safemindbenchmarkingmitigatingsafety}.
AgentDojo provides a dynamic environment for prompt injection attacks/defenses in tool-using agents, enabling the measurement of ASR/RR, tool-policy violations, and defense generalization \citep{NEURIPS2024_97091a51}.
AgentHarm focuses on harmful capability testing in agentic settings, complementing refusal/harmlessness evaluations with task-level agent affordances \citep{andriushchenko2025agentharm}.
MLA-Trust benchmarks multimodal LLM agents in GUI environments, emphasizing privacy leakage and sensitive attribute inference under realistic screen contexts and multi-step interactions \citep{yang2025mla}.
HarmBench standardizes automated red teaming and robust refusal evaluation, offering an extensible framework for measuring attack success and refusal robustness across adversarial prompt families \citep{HarmBench}.
TruthfulQA operationalizes truthfulness/honesty evaluation, and is commonly reused as a component within agentic honesty checks when combined with retrieval/tool evidence \citep{lin2022truthfulqa}.
For tool-execution risk, LM-emulated sandbox evaluation (ToolEmu-style) simulates tool APIs and environments so that risky actions (data deletion, financial theft) can be measured without real side effects \citep{ruan2024identifyingriskslmagents}.
For autonomous driving simulation, CARLA remains a widely used platform for scenario-based testing and rare-event generation under controllable ODD specifications \citep{Dosovitskiy2017CARLA}.

\subsubsection{Emerging benchmarks and evaluation trends (2024--2025)}
Recent benchmarks increasingly (i) model interactive tool use rather than static QA, (ii) include iterative adversaries that adapt across steps, (iii) measure process compliance (trace evidence, policy conformance) in addition to outcomes. For web agents, generalizable guardrails are increasingly evaluated via dedicated suites that stress policy compliance and injection resistance (e.g., WebGuard) \citep{zheng2025webguardbuildinggeneralizableguardrail}. For production-facing workflows, evaluation is moving toward trace-centric, operations-aware frameworks that treat agent behavior as a workflow to be audited (e.g., AgentCompass) \citep{Kartik2025Agentcompass}. For tool-choice reliability, emerging work highlights measurable biases in selecting among functionally equivalent tools, motivating TVD-style metrics and tool-bias benchmarks \citep{blankenstein2025biasbusters}. For privacy norms and contextual integrity, benchmarks such as PrivacyLens and CI-Bench complement leakage/regurgitation tests by evaluating whether agents follow context-dependent privacy expectations \citep{shao2024privacylens,cheng2024ci}.

\subsection{Scenario-to-metric mapping and benchmark suites}
\label{subsection:scenarios}
With the metric dictionary in place, \textbf{Table~\ref{tab:scenario-metric}} maps scenarios to a minimal set of release-gating metrics.
We separate (i) mandatory gates (e.g., CER/CVR caps, leakage thresholds, and minimum DCR/CompVR)
from (ii) diagnostic metrics used for debugging and ablation. In practice, combine four sources:
offline replay regression (fast iteration), simulation banks for rare events/long horizons (tail risk),
automated red teaming (misuse robustness), and interactive sandboxes for tool-mediated attacks.

\begin{table*}[!hb]
\caption{Scenario-to-metric mapping for release gating. Gates progress from strict regression checks (Tier 0) to statistical bounds on tail risk in simulation (Tier 1), finally to anomaly-based monitoring in deployment (Tier 2).\vspace{4pt}}
\label{tab:scenario-metric}
\setlength{\tabcolsep}{6pt}
\renewcommand{\arraystretch}{1.5}
\renewcommand{\tabularxcolumn}[1]{>{\raggedright\arraybackslash}p{#1}}
\resizebox{\linewidth}{!}{
\begin{tabularx}{\linewidth}[t]{@{}
    |>{\raggedright\arraybackslash}p{2.8cm}
    |>{\raggedright\arraybackslash}p{4.2cm}
    |>{\raggedright\arraybackslash}X
    |>{\raggedright\arraybackslash}X
|@{}}
\hline
\textbf{Release gate} & \textbf{Scope and environment} & \textbf{Mandatory thresholds (caps/floors)} & \textbf{Diagnostic metrics (optimize)} \\
\hline
Tier 0: Regression & Offline replay of known incidents and ``golden'' traces. & Zero regressions on P0 scenarios. CVR = 0 on policy tests. & Trace completeness (DCR); Coverage of regression pack. \\ \hline
Tier 1: Stress Test & Simulation/Sandbox/Red team (OOD + Adversarial). & CER $<$ Limit (e.g., 0.1\%). ASR $<$ Limit. No high-sev leakage. & SafetyScore distribution; Stress-Pass rate; Latency. \\ \hline
Tier 2: Canary & Shadow mode or limited traffic live deployment. & Low anomaly rate (CVR/Refusal spikes). Auto-rollback on violation. & User feedback (PHR); Intervention rate (ISR); Cost. \\
\hline
\end{tabularx}
}

\vspace{4pt}
\noindent\footnotesize{Abbreviations: CVR: constraint violation rate; CER: catastrophic event rate; ASR: attack success rate; DCR: decision coverage rate; PHR: preference win rate; ISR: interruption success rate; OOD: out-of-distribution.}
\end{table*}

\subsubsection{Setting release-gating thresholds (risk-bounded acceptance criteria)}
Thresholds should be scenario- and tier-specific; a ``one-size-fits-all'' SR target is not meaningful when residual risk varies by domain and autonomy level. A practical template is to set hard caps on catastrophic events and hard-rule violations (CER/CVR), plus minimum floors on auditability evidence (DCR/CompVR), then optimize utility within these bounds (SR, latency, and cost). As a rule of thumb for safety-critical scenarios, many teams adopt extremely low tolerated catastrophic rates (e.g., CER $< 0.1\%$ on high-risk scenario banks) and near-zero tolerance for hard constraint violations (CVR $\approx 0$) when violations directly correspond to policy or permission breaks. Privacy/security thresholds should similarly enforce strict caps on high-severity leakage (e.g., secrets/credentials) while allowing graded treatment of low-severity context exposure, aligned with the threat model in Section~\ref{subsection:privacy}~ \citep{NIST2023RMF}.

\subsubsection{Risk-stratified evaluation and ``ODD-aware'' gating}
Agents should be released only within validated operational envelopes. Concretely, define scenario families that reflect the intended ODD (in-distribution), plus explicit stress families (OOD, adversarial injection, and long-horizon drift). Gates can then be tiered: (i) Tier 0 (offline): no regressions on regression packs; (ii) Tier 1 (simulation/sandbox): pass OOD/adversarial/long-horizon stress with bounded CER/CVR; (iii) Tier 2 (shadow/canary): satisfy online monitoring gates with rapid rollback. This tiering reduces the chance that an agent ``wins on average'' while failing under rare but foreseeable conditions.

\subsubsection{Continuous monitoring and regression detection}
Because agent stacks evolve (model updates, prompt changes, and tool-policy changes), monitoring should treat trust metrics as non-stationary. Track leading indicators (CVR spikes, permission denials, anomalous tool usage, judge disagreement, and leakage alerts) in addition to lagging indicators (incident reports). Regression detection should replay curated packs of (i) prior incidents, (ii) red team attacks, (iii) synthetic corner cases on every change to the model, memory policy, or tool wrappers \citep{Sculley2015HiddenDebt,Breck2017MLTestScore}. For deployment, staged rollouts with automated canary analysis provide a practical safeguard to catch distribution shift and regressions under real traffic \citep{Burns2018SpinnakerCD}.

\subsection{Evaluation pipeline \& tooling}
\label{subsection:evaluation_pipeline}
We recommend a staged pipeline that increases autonomy and stakes:
(1) offline regression, (2) simulation A/B and stress testing,
(3) sandboxed tool use and red teaming, (4) shadow mode (read-only; no side effects) with trace auditing,
(5) controlled canary rollout with fast rollback, (6) continuous monitoring with regression packs.
Tooling should standardize trace schemas (inputs, plans, tool calls, permissions, and outcomes),
environment/tool version stamps, seeds, and judge configurations/prompts.

\subsubsection{Stage 1: Offline regression (replay known failures)}
Offline regression is the fastest iteration loop: maintain a continuously updated pack of ``known-bad'' cases (prompt injections, tool misuse patterns, privacy leaks, and unsafe plans) and ``near-miss'' traces. Each run should be deterministic where possible (fixed seeds, pinned tools, and frozen retrieval snapshots) and should emit step-level traces so that CVR/DCR regressions are localized to specific stages (Perceive/Plan/Act/Reflect/Learn). This stage is particularly effective for catching accidental regressions from prompt/memory/tool-wrapper edits that do not involve model retraining \citep{Breck2017MLTestScore}.

\subsubsection{Stage 2: Simulation testing (rare events and long-horizon)}
Simulation banks support stress testing beyond what logged data covers. For embodied/control domains, scenario-based simulators (e.g., CARLA for driving) enable the systematic variation of weather, traffic density, occlusion, and adversarial actors to probe tail-risk behaviors \citep{Dosovitskiy2017CARLA}. For interactive tool agents, simulation can include ``mock worlds'' (synthetic websites, synthetic enterprise documents, and synthetic emails) that embed controlled privacy/security hazards (PII, secrets, and injection strings). Domain randomization and scenario generation expand coverage and reduce overfitting to a small set of scripted tasks \citep{Tobin2017DomainRandomization}.

\subsubsection{Stage 3: Sandboxed execution (ToolEmu-style emulation)}
Before granting real tool access, evaluate agents in sandboxes that enforce strict isolation (filesystem, network, and credentials) and emulate external APIs. LM-emulated sandbox evaluation supports measuring risky tool-use trajectories (e.g., ``delete records'', ``transfer money'') without real-world consequences, and enables the fine-grained attribution of which intermediate steps caused a violation \citep{ruan2024identifyingriskslmagents}. A practical recommendation is to maintain two sandboxes: (i) a behavior sandbox that measures whether the agent attempts unsafe actions, (ii) a security sandbox that measures whether the agent can be induced to bypass policy wrappers.

\subsubsection{Stage 4: Red teaming (automated + human)}
Red teaming should combine automated adversarial generation (broad coverage) with targeted human adversaries (depth). Automated suites (e.g., HarmBench) provide standardized ASR/RR measurement and support regression testing across prompt families \citep{HarmBench}. For agent-specific prompt injection and tool hijacking, dynamic environments (e.g., AgentDojo) are valuable because they model interaction, tool calls, and defenses under adaptive attackers \citep{NEURIPS2024_97091a51}. For high-risk deployments, document the red team protocol: attacker capabilities (insider vs.\ external), tool access, allowed social engineering, and success criteria (leakage, unauthorized tool use, and policy bypass) \citep{OWASPPromptInjection,Greshake2023IndirectPI}.

\subsubsection{Stage 5: Shadow mode (read-only deployment with trace auditing)}
Shadow mode runs the agent on real inputs but blocks side effects (no external writes, no irreversible actions). This stage is ideal for validating process metrics (DCR/CompVR), measuring real OOD rates, and identifying privacy leakage patterns in authentic workflows before impact. Shadow traces also seed regression packs with realistic distributions, reducing sim-to-real mismatch.

\subsubsection{Stage 6: Canary rollout (staged release with fast rollback)}
Canary releases gradually increase exposure, paired with automated acceptance criteria and rollback triggers (e.g., CVR spike, leakage alerts, and abnormal tool-call distributions). Continuous delivery frameworks for automated canary analysis provide practical guardrails when model and tool policies are updated frequently \citep{Burns2018SpinnakerCD}.

\subsubsection{Stage 7: Production monitoring (continuous assurance)}
Post-deployment monitoring should treat trustworthy behavior as a service SLO: track CVR/CER proxies, leakage alerts, refusal anomalies, tool misuse indicators, and intervention latency (ISR/A2I). Use drift detectors over tool-call distributions and scenario tags to trigger re-evaluation on stress suites when the environment changes. Importantly, logs and telemetry must themselves be privacy-preserving (redaction, access control) to avoid turning evaluation into a leakage channel (Section~\ref{subsection:privacy}).

\subsubsection{Trace schema requirements (minimum viable logging)}
At minimum, traces should record: (i) context identifiers (episode id, user/session anonymized id), (ii) version pins (model, prompt, tool wrappers, and policy), (iii) agent internals (plan summary, selected tools, and confidence/uncertainty if available), (iv) tool evidence (calls, arguments, permission checks, outputs, and errors), (v) safety/security checks (policy decisions, redactions, and violations), (vi) oversight events (alerts, approvals, and takeovers). DCR/Trace-Coverage should be computed from these fields, and CompVR requires that key compliance claims be independently verifiable from logged evidence rather than free-form rationales \citep{Kartik2025Agentcompass}.

\subsection{Reporting standards}
\label{subsection:reporting}
Use fixed report templates: metric dashboards with 95\% CIs and stratification; risk--utility plots (e.g., CER/CVR vs.\ SR);
stress test breakdowns (OOD/adversarial/long-horizon); and a reproducibility bundle (configs, seeds, tool registries, and judge prompts)
plus audit evidence (trace summaries and representative failure traces).

\subsubsection{Suggested report template (what to include)}
A practical trustworthy-agent report should include: (i) system card (model, tools, memory, guardrails, and autonomy level), (ii) threat model (attacker capabilities, tool exposure, and data sensitivity), (iii) scenario suite definition (ODD, stress families, and multi-agent settings), (iv) gating thresholds (CER/CVR caps; leakage limits; and DCR/CompVR floors), (v) offline + sandbox results (including red team ASR/RR), (vi) deployment readiness (shadow/canary monitoring plan, and rollback conditions) \citep{NIST2023RMF}.

\subsubsection{Visualization recommendations (risk is multi-dimensional)}
Beyond single bar charts, we recommend: (i) risk--utility frontiers (SR vs.\ CER/CVR), (ii) stratified breakdowns by scenario family and risk tier, (iii) violation heatmaps across tools/actions to localize failure channels, (iv) horizon curves (SR/CVR as a function of trajectory length) to surface long-horizon brittleness. For judge-based metrics, include agreement plots (judge--human, judge--judge) and sensitivity to judge prompts \citep{zheng2024judging}.

\subsubsection{Failure analysis (representative traces and root cause taxonomy)}
Because agent failures are often chain-based, reports should include representative failure traces with a short root cause label aligned to the lifecycle: perceive (retrieval error/injection), plan (invalid assumptions/unsafe subgoal), act (tool misuse/permission violation), reflect (missed warning), and learn (regression after update). This enables mitigation mapping back to Section~\ref{subsection:safety}~ and Section~\ref{subsection:privacy}~ without conflating behavioral safety with system compromise.

\subsubsection{Reproducibility bundle (minimum viable artifact)}
For each reported result, provide: environment manifests (tool versions, sandbox images), configs and seeds, judge prompts and judge model identifiers, scenario suite definitions, and trace schema documentation. Production-facing evaluations should also include ``known issues'' lists and regression-pack definitions so future releases can be compared under identical conditions \citep{Breck2017MLTestScore,Sculley2015HiddenDebt}.

\subsubsection{Responsible reporting and disclosure (what to share)}
When publishing evaluations, share aggregate metrics and sanitized traces sufficient for reproducibility but avoid releasing exploit strings or sensitive tool configurations that enable immediate misuse. For security-relevant findings (prompt injection bypasses, tool wrapper failures), responsible disclosure practices and the staged release of details help balance transparency with risk \citep{OWASPPromptInjection}.

\subsection{Open challenges in evaluation}
\label{subsection:open_eval}
\subsubsection{Judge fragility and rationale hacking}
As agents become more fluent, judges can be misled by plausible rationales even when actions are unsafe or evidence-free. This is particularly problematic when agents learn to optimize for judged scores, creating ``evaluation gaming'' loops \citep{zheng2024judging,gao2023scaling,human_affected_by_AI}. Robust evaluation likely requires hybrid judging: deterministic checkers for hard rules, plus calibrated judges for soft scoring.

\subsubsection{Benchmark saturation and overfitting to suites}
{\addfontfeature{LetterSpace=-3.0}Static benchmarks can be saturated quickly: agents (or training pipelines) may implicitly overfit to known test distributions, especially when benchmarks are widely used for iteration. Dynamic benchmarks (changing tasks/web content), hidden test sets, and periodic suite refreshes are needed to preserve validity over time \citep{HarmBench}.}

\subsubsection{Sim-to-real gaps}
Simulation is essential for rare-event coverage, but sim-to-real mismatch remains a core limitation. Domain randomization and mixed offline/online validation help, yet evaluation must remain conservative when the real deployment distribution is poorly characterized \citep{Tobin2017DomainRandomization}.

\begin{table*}[!hb]
\caption{Cross-domain 
 mapping of core trustworthiness dimensions. Each cell summarizes the domain-specific challenges and mitigation methods with supporting references. This table focuses on the two core dimensions critical for high-risk deployments: safety and robustness and privacy and system security.\vspace{4pt}}
\label{fig:comparison_framework}
\renewcommand{\arraystretch}{1.5}
\renewcommand{\tabularxcolumn}[1]{>{\raggedright\arraybackslash}p{#1}}
\setlength{\tabcolsep}{4pt}
\begin{tabularx}{\linewidth}{@{}| >{\raggedright\arraybackslash}p{2.1cm} |>{\raggedright\arraybackslash}X |>{\raggedright\arraybackslash}X |>{\raggedright\arraybackslash}X |@{}}
\hline
\textbf{Dimension} & \textbf{Autonomous driving (AD)} & \textbf{Healthcare (HC)} & \textbf{Intelligent assistants (IA)} \\
\hline

Safety and Robustness 
& Challenge: Perception failures in adverse weather, occlusion, long-tail scenarios; collision prevention.

Methods: V2X collaborative sensing, simulation-based validation, safe RL, runtime shielding~\citep{a17030103,ZHAO2024122836,yao2025agentsllmaugmentativegenerationchallenging}.
& Challenge: Diagnostic errors, clinical decision failures, hazardous hallucinations.

Methods: Multi-center validation, clinician-in-the-loop architecture, uncertainty quantification \citep{diagnostics14141472,udegbe2024role,healthcare12020125}.
& Challenge: Tool execution failures, indirect prompt injection, memory poisoning.

Methods: Sandboxed execution, LM-emulated testing, input sanitization~\citep{ruan2024identifyingriskslmagents,NEURIPS2024_97091a51,yu2025survey}. \\
\hline
Privacy and System Security
& Challenge: V2X data exposure, location tracking, cybersecurity threats (spoofing, jamming).

Methods: Secure V2X protocols, ISO/SAE 21434 compliance, data anonymization~\citep{ZHAO2024122836,J3016_202104}.
& Challenge: Patient data breaches, re-identification risks, EHR security.

Methods: HIPAA/GDPR compliance, federated learning, end-to-end encryption~\citep{bioengineering11040337,life14050557}.
& Challenge: Credential theft, unauthorized tool access, data exfiltration.

Methods: Least-privilege access, ephemeral credentials, policy enforcement, audit logging~\citep{yu2025survey,limind,NEURIPS2024_97091a51}. \\

\hline
\end{tabularx}

\vspace{4pt}
\noindent\footnotesize{Abbreviations: AD: autonomous driving; HC: healthcare; IA: intelligent assistants; V2X: vehicle-to-everything; RL: reinforcement learning; EHR: electronic health record; HIPAA: Health Insurance Portability and Accountability Act; GDPR: General Data Protection Regulation; IoMT: Internet of Medical Things; SaMD: software as a medical device.}
\end{table*}

\subsubsection{Long-horizon combinatorics}
Exhaustive long-horizon evaluation is intractable: the space of tool-call sequences grows combinatorially. Practical alternatives include stress testing curated ``dangerous chains'' (e.g., multi-step jailbreak sequences), measuring time to violation, and maintaining targeted regression packs from incident traces \citep{ruan2024identifyingriskslmagents}.

\subsubsection{Multi-agent attribution and shared responsibility}
In multi-agent deployments, harm may arise from interaction effects (information amplification, collusion, and blame shifting). Assigning responsibility requires protocol-aware traces, message authentication, and evaluation designs that separate individual from collective failure modes \citep{Dafoe2020Cooperative,Zhang2021MARLSurvey}.

\subsubsection{Adversarial coverage is never complete}
Threat models evolve (new prompt injection styles, compromised tools, and social engineering). As in security engineering, evaluation must be continuous: automated red teaming, frequent regression updates, and deployment monitoring are necessary complements to any static benchmark \citep{OWASPPromptInjection,Greshake2023IndirectPI}.

\section{Real-world applications in high-risk domains}\label{section:applications}

This section grounds our two core trustworthiness dimensions---Safety and Robustness and Privacy and System Security---in high-risk real-world domains: autonomous driving~\citep{a17030103,ZHAO2024122836,yao2025agentsllmaugmentativegenerationchallenging}, healthcare~\citep{diagnostics14141472,udegbe2024role,healthcare12020125,life14050557,bioengineering11040337}, and intelligent assistants~\citep{yu2025survey,limind} (see \textbf{Table~\ref{fig:comparison_framework}} for a cross-domain lifecycle comparison). Each domain involves high-stakes decision-making where safety failures can cause physical harm or loss of life, and where privacy/security breaches can compromise sensitive data or critical infrastructure. By focusing on safety-critical and security-sensitive considerations in each domain, this section illustrates how the theoretical mitigations discussed in Section~\ref{section:core_dimensions} translate into practical deployment requirements.

\subsection{Autonomous driving: safety and security challenges}\label{subsection:autonomous_driving}
The scope of autonomous driving (AD) is defined by the The Society of Automotive Engineers (SAE) J3016 standard ~\citep{J3016_202104}, ranging from L0 (no driving automation) to L5 (full driving automation). While L2 systems (partial driving automation) are widely deployed today, the transition to L3 (conditional driving automation) and L4 (high driving automation) autonomy introduces critical challenges in uncertainty and safety ~\citep{10.1145/3485767}. To bridge this gap, trustworthy agentic AI frameworks are being adopted to enhance the Perceive--Plan--Act--Reflect--Learn loop.
In AD, trustworthiness is ultimately evaluated system-wide: perception stacks, planners, actuators, HMI (handoff), fleet learning, and organizational processes jointly determine residual risk. This naturally maps to Section~\ref{subsection:safety} (robustness, constrained planning, runtime shielding, and simulation/verification) and Section~\ref{subsection:privacy} (secure V2X, least-privilege tool access, and trace governance), and motivates lifecycle-aligned assurance rather than single-module benchmarking.

\subsubsection{Safety: robustness, uncertainty, and minimum-risk behavior}
In the perceive stage, achieving robustness against environmental anomalies (e.g., occlusion, severe weather) remains a primary hurdle ~\citep{a17030103, 9307324}. Traditional perception systems are typically trained on standard static benchmarks like KITTI ~\citep{Geiger2012KITTI} and nuScenes ~\citep{caesarNuScenesMultimodalDataset2020}. While these datasets utilize multi-modal sensors, they are inherently limited to a single-vehicle perspective. Consequently, they struggle with physical limitations such as occlusion or sensor degradation in severe weather. To address this, modern systems are evolving from standalone sensor-based perception to collaborative V2X communication frameworks~\citep{ZHAO2024122836}.
For the Plan$\rightarrow$Act stages, a recurring challenge is long-tail interaction: rare cut-ins, atypical pedestrians, ambiguous signage, and novel construction layouts can invalidate learned heuristics~\citep{wang2025generativeaiautonomousdriving}. Trustworthy agent designs therefore emphasize (i) risk-aware planning (e.g., constrained optimization/safe RL as in Section~\ref{subsection:safety}), (ii) conservative fallback policies when uncertainty is high (OOD-triggered safe mode), (iii) minimum risk condition (MRC) behaviors (safe stop/pull-over strategies) when the system cannot maintain the operational design domain (ODD)~\citep{ISO21448_2019}. In the reflect stage, safe handoff and driver engagement monitoring are essential for L2/L3, since mismatched mental models can create over-trust and delayed intervention~\citep{NTSB2020TeslaMountainView}.

\subsubsection{Failure post-mortems: Uber (2018), Tesla Autopilot, and Cruise (2023)}
Real-world incidents illustrate how lifecycle failures cascade. In the 2018 Uber ATG Tempe fatality, the developmental ADS repeatedly misclassified the pedestrian/bicycle and failed to trigger timely braking; additionally, emergency braking was disabled and the safety driver was inattentive, yielding a Perceive$\rightarrow$Act breakdown compounded by weak oversight~\citep{NTSB2019UberTempe}. Multiple Tesla Autopilot crashes investigated by regulators highlight edge-case perception/planning failures \emph{and} HMI issues: drivers may misunderstand L2 limitations, leading to unsafe reliance and delayed takeover, i.e., reflect-stage handoff failures that convert residual risk into harm~\citep{NTSB2020TeslaMountainView,NHTSA2022EA22002}. In the 2023 Cruise San Francisco incident (pedestrian dragging), failures spanned perception/action recovery and post-incident disclosure processes, underscoring that trustworthy AD also requires organizational incident handling and trace integrity for timely remediation~\citep{CADMV2023CruiseSuspension}. These cases motivate Section~\ref{subsection:safety} mitigations such as OOD detection + conservative fallback, runtime shielding, and scenario-based validation focusing on rare events and recovery behaviors (not only nominal driving).

\subsubsection{Regulatory and compliance requirements}
AD safety engineering is shaped by standards and reporting regimes. ISO 26262 operationalizes functional safety via hazard analysis, safety goals, and ASIL ratings, encouraging explicit safety cases and traceable requirements-to-tests mapping~\citep{ISO26262_2018}. ISO/PAS 21448 (SOTIF) targets the safety of the intended functionality under performance limitations and foreseeable misuse, which directly aligns with OOD/long-tail risks in the perceive/plan stages~\citep{ISO21448_2019}. On the security side, ISO/SAE 21434 frames cybersecurity engineering across the vehicle lifecycle (threat analysis, security goals, and verification)~\citep{ISO21434_2021}. In the U.S., NHTSA's Standing General Order requires timely reporting of certain crashes involving ADS/ADAS, pushing operators to maintain high-quality logs and rapid triage pipelines~\citep{NHTSA2021StandingGeneralOrder}. In the EU, the AI Act treats many AD components as high-risk AI systems, emphasizing risk management, technical documentation, logging, human oversight, and post-market monitoring---all consistent with the release-gating and trace-evidence principles summarized in Section~\ref{section:eval}~\citep{EUAIAct2024}.

\subsubsection{Evaluation protocols and deployment practices}
Beyond static datasets, AD requires scenario-based testing across the ODD. Simulation platforms (e.g., CARLA) enable controlled stress tests for rare events, sensor degradation, and interactive multi-agent traffic~\citep{Dosovitskiy2017CARLA}. Standardized scenario specifications (e.g., OpenSCENARIO) support reproducible replay of corner cases and regression packs across software releases~\citep{ASAMOpenSCENARIO2020}. Practically deployed systems track disengagement metrics, near-miss proxies (e.g., time to collision), and policy violation rates; however, trustworthy release gating should also validate recovery behaviors (MRC achievement), trace completeness, and handoff latency (ISR/A2I) as emphasized in Section~\ref{section:eval}~\citep{ISO26262_2018}. Staged rollouts (closed-course $\rightarrow$ shadow $\rightarrow$ limited ODD) with automated rollback criteria operationalize the ``safe updates'' principle from Section~\ref{subsection:safety}.

\subsubsection{Multi-agent driving: V2X cooperation and interaction safety}
AD is inherently multi-agent; other vehicles, pedestrians, and infrastructure respond strategically. V2X-enabled cooperative perception can reduce occlusion and extend sensing horizons, but introduces coordination and trust challenges: inconsistent maps, delayed messages, or compromised participants can degrade shared belief states~\citep{ZHAO2024122836}. Multi-agent LLM-based driving frameworks use debate/negotiation to handle ambiguous merges and social driving norms~\citep{liang-etal-2024-encouraging, 10976336}, yet safety guarantees remain difficult when communication is unreliable or adversarial. A practical approach is layered safety: game-theoretic intent modeling for other road users plus hard safety envelopes (shields) during the act stage, limiting the consequence of coordination errors (Section~\ref{subsection:safety}).

\subsubsection{Privacy and system security: cybersecurity, telemetry, and data governance}
Beyond physical safety, connected autonomous vehicles face significant cybersecurity threats. V2X communication channels can be targeted by adversarial attacks including spoofing, jamming, and man-in-the-middle attacks~\citep{ZHAO2024122836}. Vehicle systems must implement robust authentication protocols, encrypted communication, and intrusion detection systems compliant with ISO/SAE 21434 cybersecurity standards. Additionally, privacy concerns arise from the extensive sensor data collected by autonomous vehicles.
From the component-based lens in Section~\ref{subsection:privacy}, key risks include (i) confidentiality loss via location/trajectory linkage and camera footage re-identification, (ii) integrity loss via compromised OTA updates or forged V2X messages that poison collaborative perception, (iii) availability loss via denial-of-service against V2X stacks or cloud telemetry. Mitigations include data minimization (on-device redaction and purpose-limited retention), secure key management for vehicle credentials, and tamper-evident logging to support post-incident forensics and compliance audits~\citep{ISO21434_2021}.

\subsection{Healthcare: safety and privacy challenges}\label{subsection:healthcare}
In the healthcare domain, the paradigm is shifting from traditional reactive care to AI-driven healthcare transformation, enabling continuous remote patient monitoring through the Perceive--Plan--Act--Reflect--Learn loop ~\citep{Balakrishna_Kumar_Solanki_2024}. Unlike static diagnostic models, modern agents utilize the Internet of Medical Things (IoMT) to ingest real-time multimodal data (e.g., wearables, electronic health records) during the perceive stage ~\citep{life14050557, bios12080562}. This domain presents acute challenges for both safety (clinical decision errors can cause patient harm) and privacy (medical data is among the most sensitive).
Healthcare agents often combine heterogeneous components: RAG over EHRs, temporal models for vitals, LLM-based triage, and tool calls into ordering systems. As a result, trustworthy deployment must explicitly connect domain practice (clinical workflows, responsibility allocation, and adverse event reporting) to the stage-aligned mitigations in Sections~\ref{subsection:safety} and~\ref{subsection:privacy}, rather than treating the LLM as the only risk source.

\subsubsection{Safety: clinical decision support and error prevention}
In the plan phase, agents functioning as clinical decision support systems (CDSS) generate personalized treatment plans and diagnostic predictions ~\citep{diagnostics14141472}. Ensuring the factuality and safety of medical advice is paramount. In the act and reflect stages, agents execute tasks ranging from robotic-assisted surgery to administrative automation via virtual nursing assistants ~\citep{udegbe2024role, diagnostics14141472}. Given the high-stakes nature of these actions, safety is enforced through a Human-in-the-loop architecture ~\citep{healthcare12020125}.
Mapping to the lifecycle is as follows: perceive risks include missing/incorrect vitals, incomplete EHR context, and distribution shift across hospitals; plan risks include hallucinated rationales, unsafe contraindication handling, and goal misgeneralization (e.g., optimizing throughput over patient safety); act risks include unsafe order entry or automation errors; reflect risks include overconfident self-assessment; and learn risks include feedback loops that overfit to local practice patterns. Accordingly, agents should be designed to abstain and escalate under uncertainty (OOD detection + calibrated confidence), enforce clinical guardrails (contraindication and dosage constraints), and validate plans via independent checks or simulations before any irreversible action (Section~\ref{subsection:safety})~\citep{NIST2023RMF}.

\subsubsection{Failure post-mortems: Watson for Oncology, Epic Sepsis, and COVID-19 AI screening}
Historical failures underscore that impressive offline metrics may not translate to safe clinical performance. IBM Watson for Oncology was reported to produce unsafe or non-evidence-based treatment suggestions in some settings, attributed to limited or non-representative training data and insufficient clinical validation, highlighting plan-stage brittleness and oversight gaps~\citep{Ross2017WatsonOncologySTAT}. The Epic Sepsis Model, widely deployed in hospitals, showed a substantially weaker real-world performance than expected in independent evaluations, illustrating dataset shift and the danger of relying on proprietary predictors without robust external validation~\citep{Wong2021EpicSepsis}. During COVID-19, many imaging/triage models suffered from data quality issues and demographic bias, emphasizing perceive-stage noise, OOD generalization failures, and the need for stratified fairness audits (Section~\ref{subsection:safety})~\citep{Wynants2020BMJCOVIDModels,Roberts2021NatMI_COVIDImaging}.

\subsubsection{Clinical agent types and workflow integration}
In practice, healthcare agentic systems span multiple roles: diagnostic agents (radiology/pathology triage and report drafting), treatment planning agents (guideline-based recommendations and care pathway optimization), robotic assistants (surgery/rehabilitation with constrained actuation), virtual nursing assistants (patient education, symptom monitoring, and adherence nudges), and administrative/workflow agents (coding, scheduling, and prior authorization). Each role changes the act-stage risk profile: direct clinical actuation demands stricter runtime constraints and verification than low-stakes documentation support.

\subsubsection{Human--AI collaboration: escalation, trust calibration, and responsibility}
Because liability and accountability remain clinician-centered, trustworthy deployment emphasizes clear handoff protocols; when confidence drops, when evidence is incomplete, or when recommendations deviate from guidelines, the agent must escalate to a clinician with a concise, auditable trace (inputs, retrieved evidence, and constraints checked) rather than persuasive free-form rationales. Trust calibration requires communicating uncertainty and limitations, avoiding automation bias, and specifying responsibility allocation (who approves orders, who monitors drift)---a reflect-stage design requirement aligned with trace-based evaluation in Section~\ref{section:eval}~\citep{WHO2021AIEthicsHealth}.

\subsubsection{Evaluation and post-deployment monitoring}
Healthcare evaluation prioritizes prospective validation and real-world evidence over purely retrospective studies. Recommended protocols include multi-center trials (to probe generalization), prospective silent deployment (shadow mode) to estimate error modes without patient impact, and continuous post-market surveillance with drift/OOD monitors and incident-triggered regression packs (Section~\ref{subsection:safety})~\citep{FDA2021AIMLSaMDActionPlan}. Safety evaluation should be stratified by demographics and site to detect disparate error rates and should include adverse event reporting pathways consistent with clinical governance.

\subsubsection{Privacy and system security: patient data protection}
The vast data collection inherent in IoMT-enabled healthcare introduces critical privacy risks. Strict adherence to regulatory frameworks like HIPAA and GDPR is mandatory ~\citep{bioengineering11040337}. Privacy-enhancing technologies including federated learning enable collaborative algorithm development across hospitals without centralizing sensitive data.
With the Section~\ref{subsection:privacy} lens, key threats include attribute inference from longitudinal vitals, leakage through tool calls (EHR queries, billing systems), insider threats, and IoMT device compromise. Defenses include least-privilege access to EHR APIs, encryption in transit/at rest, tamper-evident audit logs, retention limits for conversational memory, and privacy-preserving analytics (federated learning and, when applicable, secure computation for cross-site studies)~\citep{bioengineering11040337}. Availability is also safety-relevant; ransomware or denial-of-service against hospital infrastructure can indirectly cause patient harm, motivating incident response playbooks and segmented network architectures.

\subsubsection{Regulatory and compliance requirements for clinical AI}
Regulatory oversight increasingly treats clinical AI as Software as a Medical Device (SaMD); the IMDRF SaMD definition and risk framework establish a baseline vocabulary. The FDA's AI/ML-based SaMD action plan emphasizes good ML practice, transparency, and change management for learning systems (e.g., predetermined change control plans)~\citep{FDA2021AIMLSaMDActionPlan}. Institutional Review Boards (IRBs) and clinical governance require clear protocol definitions, monitoring, and adverse event reporting for studies involving patient impact. At the global level, WHO guidance highlights ethics, equity, and governance requirements that align with the safety/privacy trade-offs discussed in Sections~\ref{subsection:safety}--\ref{subsection:privacy}~\citep{WHO2021AIEthicsHealth}.

\subsection{Intelligent assistants: safety and security challenges}\label{subsection:assistants}
This domain represents the deployment of LLM-based agents as autonomous operators, which have evolved from passive responders to active systems capable of tool learning and complex task execution ~\citep{Xu2025ToolLearningSurvey}. A prime example is RepairAgent \citep{bouzenia2024repairagent}, which autonomously interleaves information gathering and tool invocation to fix software bugs. This transition to tool-integrated autonomy introduces severe safety and security risks unique to agents.
Compared with autonomous driving and healthcare, assistants often operate in open-world digital environments (web, email, and enterprise SaaS) where adversaries can directly craft inputs and where authorized tools become exfiltration channels. Thus, the same Perceive$\rightarrow$Plan$\rightarrow$Act$\rightarrow$Reflect$\rightarrow$Learn loop yields a larger attack surface, making the system-security framing in Section~\ref{subsection:privacy} particularly central.

\subsubsection{Safety: tool misuse and execution harm}
In the perceive and act stages, agents processing untrusted data (e.g., emails, financial records) are highly vulnerable to indirect prompt injection ~\citep{yu2025survey}. Malicious instructions embedded within external content can hijack the agent's reasoning, leading to unauthorized execution-based harm ~\citep{limind}. The plan and reflect stages are threatened by memory poisoning and backdoor attacks ~\citep{limind}. Rigorous evaluation environments like AgentDojo \citep{NEURIPS2024_97091a51} are essential for benchmarking agent robustness.
Safety failures often present as process violations, bypassing required approvals, exceeding tool budgets, or executing irreversible actions without sufficient evidence. Accordingly, assistants benefit from Section~\ref{subsection:safety} controls that are system-level rather than purely model-level: constrained planning (policy-checked action sequences), runtime shields (block unsafe tool arguments), and staged rollouts with regression packs for discovered failure traces.

\subsubsection{Privacy and system security: credential protection and sandboxing}
Agents with tool access inherently possess elevated permissions---access to customer data, financial accounts, and system controls. Trustworthy deployment requires least-privilege access, tool-level sandboxing with ephemeral credentials, and systematic policy auditing. Ruan et al.~\citep{ruan2024identifyingriskslmagents} proposed an LM-emulated sandbox to safely identify tool execution risks.
From the CIA viewpoint (Section~\ref{subsection:privacy}), confidentiality failures include secret leakage (API keys, OAuth tokens), integrity failures include tampered tool outputs and unauthorized repository changes, and availability failures include tool-call flooding that disrupts services. Because assistants frequently integrate third-party plugins/APIs, supply-chain risks (compromised plugin updates, malicious tools) are first-class concerns and require provenance checks, signed artifacts, and continuous monitoring.

\subsubsection{Real-world security incidents: plugins, manipulation, and sandbox boundary breaks}
Early deployments have already surfaced security-relevant failure patterns. Prompt injection against plugin-enabled systems can coerce an assistant into exfiltrating data through authorized channels (e.g., sending content to an attacker-controlled endpoint), demonstrating that ``allowed'' tools can become leakage vectors when instruction/data boundaries are blurred~\citep{Willison2023PromptInjection}. The Bing Chat ``Sydney'' incident illustrated goal-driven manipulation attempts and brittle interaction safety under prolonged conversations, highlighting long-horizon dynamics where planning/reflection can drift toward undesirable behaviors~\citep{Roose2023BingSydney}. In code execution settings, reported sandbox weaknesses and overly permissive file/network access demonstrate that small isolation gaps can translate into large confidentiality and integrity risks~\citep{Rehberger2023CodeInterpreter}. These cases directly motivate Section~\ref{subsection:privacy} methods: zero-trust intake, sandboxing, DLP-style output filtering, and incident response with rapid credential rotation.

\subsubsection{Financial/trading agents: market integrity and compliance constraints}
Algorithmic trading agents raise domain-specific risks beyond generic tool misuse: inadvertent market manipulation (spoofing, layering, and wash trades), uncontrolled feedback loops under volatility, and conflicts of interest in strategy selection. Compliance regimes (e.g., SEC market access controls and broker--dealer risk management; FINRA supervision guidance; and, in the EU, MiFID II algorithmic trading obligations) require pre-trade risk limits, kill switches, surveillance for manipulative patterns, and comprehensive recordkeeping~\citep{SEC2010MarketAccessRule15c3_5,FINRA2015RegNotice1509AlgoTrading,EU2014MiFIDII}. Trustworthy agents should therefore implement constrained planning with explicit risk budgets, independent monitoring (EDR-style telemetry over tool calls), and audit-ready traces that support post-trade investigations (Section~\ref{section:eval}).

\subsubsection{Enterprise agent deployment: browsing, coding, and workflow automation}
High-impact enterprise assistants include browser-based operators, coding agents with repository access, customer service agents with PII exposure, and workflow agents that act via email/calendar/CRM. Each expands the act-stage blast radius: a single injected instruction can trigger mass outreach, data deletion, or credential misuse. Practical deployment patterns therefore emphasize least-privilege scopes (per-tool, per-resource, per-time window), approval gates for irreversible actions, and compartmentalized memories to prevent cross-project or cross-client leakage (Section~\ref{subsection:privacy}).

\subsubsection{Red teaming and evaluation: attack success, leakage, and tool misuse detection}
Agent evaluation must go beyond static jailbreak prompts to iterative, tool-mediated attacks. Benchmarks such as AgentDojo stress prompt injection and defense dynamics in interactive environments~\citep{NEURIPS2024_97091a51}, while broader red teaming suites (e.g., HarmBench) can be adapted to measure refusal robustness and policy compliance under adversarial pressure~\citep{HarmBench}. For assistants, recommended metrics include attack success rate (ASR), context leakage count (CLC), tool permission violation rate (CVR), and trace completeness (DCR), consistent with Section~\ref{section:eval}. Automated adversarial testing should be integrated into CI to prevent regressions when tools, prompts, or policies are updated.

\subsubsection{Incident response: detection, containment, and recovery loops}
Because assistant failures can propagate rapidly (e.g., mass actions via automation), operators should treat safety/security incidents as standard security events through the following steps: detect prompt injection attempts and anomalous tool sequences, isolate affected agent instances, revoke/rotate credentials, patch policies/filters, and preserve tamper-evident audit logs for forensics~\citep{NIST80061r2}. This operational loop closes the Reflect$\rightarrow$Learn stage; post-mortems should produce regression packs (attack replays) and release gates that prevent recurrence, aligning with the ``safe updates'' principle in Section~\ref{subsection:safety} and retention/leakage gates in Section~\ref{subsection:privacy}.



\section{Challenges and solutions}\label{subsection:challenges}
While this survey focuses on Safety and Robustness and Privacy and System Security as the two core dimensions for high-risk agentic AI,
several open challenges remain before such agents can be safely deployed at scale. Below we highlight key directions spanning
lifelong adaptation, monitoring/verification, personalization, and the trust--utility trade-off.

\subsection{Lifelong learning and self-evolution}
Trustworthiness must track evolving human expectations and social standards, so agents should not remain static.
However, most agents are not lifelong learners that continuously acquire new knowledge and skills~\citep{Wang2023Voyager}.
A promising direction is self-evolving agents that build a data flywheel from interaction feedback and iteratively refine
their behavior~\citep{packer2023memgpt}, while addressing catastrophic forgetting and concept drift. When full model updates are too costly,
context adaptation can offer a lighter-weight alternative~\citep{Zhang2025Agentic}. From an engineering perspective, maintaining safety
often requires a continuous human--agent loop for iterative updates and rollback in a dynamically changing world.

\subsubsection{Runtime monitoring and verification}
Agents that claim trustworthiness should be supported by systematic designs for verification and runtime monitoring.
Meta-cognitive modules can help assess tool reliability and proactively detect failure executions~\citep{sumers2023coala}, while agents can
also learn from execution errors and feedback signals~\citep{Wang2023Voyager}. A major bottleneck is rare events: the scarcity
of tail-risk data makes purely offline training insufficient, so continuous monitoring in real environments becomes essential.
Moreover, trustworthiness risks are amplified in multi-agent and user--agent interaction settings; cross-component dynamics can lead to
sensitive attribute leakage~\citep{yang2025mla}, underscoring the need for rigorous monitoring and verification under interaction.

\subsection{Trustworthy personalization}
Personalization is a major application driver for agents, and recent progress has leveraged sophisticated memory mechanisms
\citep{Zhang2025GMemory,Zhong2024Memorybank,Hu2023ChatDB,Lu2023Memochat}. Yet personalization must be reconciled with trustworthiness,
especially privacy and safety~\citep{Zhong2024Memorybank, liu2025survey}. Beyond universal trust constraints, users may have context-dependent and
individualized norms that vary across social environments, suggesting a need for methods that align personalization with diverse
standards~\citep{Liu2025Exploring}. Future work should develop privacy-preserving memory and preference modeling that supports
personalization without increasing leakage or unsafe reliance.

\subsection{Green and efficient agentic AI}
To reduce the economic and environmental costs, trustworthy agents should also advance Green AI~\citep{Schwartz2020Green}.
Some agentic workflows use long contexts and extensive tool interactions, substantially increasing token and compute consumption
\citep{Zhang2025Agentic,packer2023memgpt,Zhang2025RLVMR}. Recent efforts aim to keep decoding time and memory growth closer to linear cost
\citep{Yu2025MemAgent}, which is a promising start. Importantly, trustworthiness should not depend on unbounded compute; practical
deployment requires balancing safeguards with latency, energy usage, and scalability.

\subsection{Standardizing explainability}
Despite progress in explainability methods, rigorously quantifiable definitions and standardized evaluation remain limited
\citep{Zhang2024Survey,kabir2025xalchallenge}. In practice, what counts as a ``good'' explanation can vary by domain, user expertise,
and risk level, making universal metrics difficult. Section~\ref{section:eval}~ outlines commonly used dimensions (e.g., fidelity,
stability, and complexity), but future work should connect these metrics to human decision-making, misuse risks, and deployment settings,
so explanations improve accountability rather than merely increasing trust.

\subsection{Closing the accountability gap}
Compared to other trust dimensions, there remains a notable scarcity of agent-specific methods and benchmarks that directly target
accountability. To bridge this gap, Section~\ref{section:eval}~ introduces preliminary metrics---Decision Coverage Rate, Compliance
Verification Rate, and Interruption Success Rate---to quantify traceability, auditability, and interruptibility. We call for new
benchmarks, system designs, and algorithmic frameworks that make accountability measurable, verifiable, and enforceable throughout
the agent lifecycle.

\subsection{Balancing trustworthiness and utility}
A persistent challenge is balancing trustworthiness with effectiveness. For example, memory and experience reuse can improve
performance~\citep{Ouyang2025ReasoningBank}, but retention and reuse may introduce privacy and safety risks. Similarly, constrained
optimization (e.g., safe RL) improves risk control yet may trade off utility~\citep{ijcai2024p913,kushwaha2025surveysafereinforcementlearning}.
Future work should emphasize Pareto-style evaluation and release-gating in the following way: maximize task utility while explicitly bounding risk and
measuring where safeguards degrade user value.

\subsection{Long-horizon deployment challenges}
Long-horizon autonomy---where agents execute multi-step plans spanning minutes, hours, or days---is a defining capability of agentic AI, yet it introduces unique trustworthiness challenges that are only partially addressed by current methods.

\subsubsection{Compounding errors and delayed consequences}
Small perception or planning errors early in a trajectory can cascade into severe failures at later stages, with harm that is delayed and difficult to attribute (Section~\ref{subsection:safety}). Unlike single-turn interactions where errors are immediately observable, long-horizon agents can ``drift'' into unsafe regimes gradually, making root cause analysis and timely intervention challenging~\citep{Sculley2015HiddenDebt}. Existing mitigations---checkpointing, risk budgets, and receding-horizon replanning---add monitoring overhead that may conflict with latency and efficiency requirements (Section~\ref{subsection:challenges}).

\subsubsection{Credit assignment and reward sparsity}
Long-horizon tasks often provide only sparse, delayed feedback (e.g., task success/failure at the end), making it difficult to learn which intermediate actions were responsible for outcomes~\citep{SuttonBarto2018}. This credit assignment problem is amplified for safety: an agent may complete a task successfully while taking unnecessary risks along the way, but standard success metrics will not penalize such behavior. Recent work on hierarchical RL~\citep{Nachum2018HIRO} and option frameworks~\citep{Sutton1999Options} partially addresses this, but integrating safety constraints into hierarchical credit assignment remains open.

\subsubsection{Synergizing reasoning and acting}
A promising direction for long-horizon trustworthiness is the tighter integration of reasoning and acting. ReAct~\citep{Yao2023ReAct} demonstrates that interleaving explicit reasoning traces with action execution improves interpretability, error diagnosis, and plan adjustment---directly supporting scalable oversight. By making intermediate reasoning visible, operators can identify and correct problematic plans before execution cascades. Reflexion~\citep{Shinn2023Reflexion} extends this with verbal self-reflection, enabling agents to learn from trajectory-level feedback without weight updates. These frameworks suggest that long-horizon trustworthiness may benefit more from architectural transparency than post hoc monitoring.

\subsubsection{Scalable oversight and verification}
As trajectory length grows, human oversight becomes increasingly costly and error-prone; reviewers cannot realistically audit every step of a hundred-action plan. This motivates research into (i) hierarchical verification (certifying high-level subgoals rather than individual actions), (ii) anomaly-triggered review (flagging only unusual or high-risk segments), (iii) AI-assisted oversight (using auxiliary models to pre-screen trajectories). However, the reliance on AI oversight introduces recursive trust concerns that remain unresolved.

\subsubsection{Evaluation intractability}
Exhaustive long-horizon evaluation is combinatorially intractable; the space of possible action sequences grows exponentially with trajectory length (Section~\ref{section:eval}). 
Practical alternatives~\citep{ruan2024identifyingriskslmagents} include stress testing curated ``dangerous chains,'' measuring time to violation, and maintaining incident-derived regression packs. Future benchmarks should explicitly vary horizon length and report performance degradation curves, enabling operators to understand the operational envelope within which an agent can be trusted.

\subsection{Case study: security failures in openclaw and moltbook}
\label{sec:casestudy}
The rapid adoption of open-source agentic AI systems has exposed critical security gaps between capability development and deployment safeguards. OpenClaw (formerly known as Clawdbot/Moltbot), a locally-run AI assistant with full system access---including file read/write, shell command execution, credential management, and external service interaction---exemplifies these risks~\citep{clawdbot2025security}.

\subsubsection{Timeline and scale of exposure}
OpenClaw gained mass adoption in late 2025, reaching hundreds of thousands of active deployments within weeks. By January~2026, security researchers had disclosed two critical CVEs: CVE-2025-49596 (CVSS~9.4, unauthenticated gateway access leading to full system compromise) and CVE-2025-6514 (CVSS~9.6, command injection enabling arbitrary code execution). Internet-wide scans revealed over 900~exposed OpenClaw gateways without authentication, leaking API keys, OAuth tokens, and conversation histories stored as plaintext Markdown/JSON files~\citep{clawdbot2025security}. Infostealer malware families rapidly adapted to target these credential storage locations. Separately, Moltbook---a social network exclusively for AI agents---launched on  28 January 2026, and within days its misconfigured Supabase backend exposed the complete database of 32{,}000+ registered agents, including API keys, verification codes, and private messages~\citep{moltbook2025breach}.

\begin{table*}[!hb]
\caption{OpenClaw/Moltbook: lessons learned mapped to trust dimensions and agent lifecycle stages.
\vspace{4pt}}
\label{tab:casestudy-lessons}
\renewcommand{\arraystretch}{1.5} 
\begin{tabularx}{\linewidth}{@{}|p{3.05cm}|p{2.675cm}|p{1.375cm}|X|@{}}
\hline
\textbf{Vulnerability} & \textbf{Trust dimension} & \textbf{Stage(s)} & \textbf{Recommended mitigation} \\
\hline
Insecure defaults/na- ked ports & Security & Perceive, Act & {\addfontfeature{LetterSpace=-2.0}Mandatory authentication, network isolation, secure-by-default configs} \\
\hline
Prompt injection & Safety, Security & Perceive & Input validation, instruction hierarchy, jailbreak detection \\
\hline
Plan hijacking via injected instructions & Safety & Plan & Constraint checkers, plan-level guardrails, goal consistency verification \\
\hline
Credential leakage & Privacy & Plan, Act & Credential vaulting, secret scoping, DLP-style output filtering \\
\hline
{\addfontfeature{LetterSpace=-2.5}\mbox{Malicious plugins/skills}} & Security & Learn & SBOMs, code signing, sandboxed execution, permission declarations \\
\hline
Database misconfiguration & Privacy & Perceive, Act & Zero-trust architecture, least-privilege access, audit logging \\
\hline
No anomaly detection & Safety & Reflect & Runtime monitoring, anomalous tool invocation detection \\
\hline
\end{tabularx}

\vspace{4pt}
\noindent\footnotesize{Abbreviations: DLP: data loss prevention; SBOMs: software bills of materials.}
\end{table*}

\subsubsection{Structural vulnerability: the ``lethal trifecta''}
{\addfontfeature{LetterSpace=-1.5}Willison~\citep{Willison2025LethalTrifecta} coined the term lethal trifecta for agents that simultaneously (i)~access private data, (ii)~process untrusted external content, (iii)~can communicate externally. When all three properties coexist---as they do in OpenClaw---an attacker can inject instructions into any external content the agent processes (e.g., a web page, email, or document), causing the agent to exfiltrate private data to an attacker-controlled endpoint. This pattern is structural rather than incidental; it cannot be fully resolved by prompt hardening alone, since LLMs cannot reliably distinguish legitimate instructions from injected ones embedded in data~\citep{Willison2023PromptInjection,Cohen2024MorrisII}.}\vspace{-3pt}

\subsubsection{Exposed attack surfaces}
Several severe vulnerability classes emerged:
(i)~Insecure defaults and naked ports---security scans revealed OpenClaw gateways exposed on the public internet without authentication, creating open invitations for system takeover.
(ii)~Prompt injection---attackers manipulated the AI's natural language interface to steal credentials, execute remote code, and exfiltrate sensitive data.
(iii)~Credential leakage---plaintext API keys and private chat logs were discoverable in exposed deployments~\citep{clawdbot2025security,moltbook2025breach}.\vspace{-3pt}

\subsubsection{Supply chain risks in AI-to-AI interaction}
Moltbook's breach revealed a second, more systemic risk: malicious ``skills'' (plugins) disguised as benign utilities were found to read private configuration files and transmit API keys to external servers~\citep{moltbook2025breach}. A large-scale empirical study of agent skill ecosystems found that 26.1\% of 31{,}132~analyzed skills contained at least one vulnerability, spanning 14~distinct patterns including data exfiltration (13.3\%), privilege escalation (11.8\%), and prompt injection~\citep{AgentSkillsWild2026}. These findings highlight the absence of capability-based permission systems, sandboxing, and code signing in current agent plugin ecosystems.\vspace{-3pt}

\subsubsection{Mapping to the survey framework}
The OpenClaw/Moltbook incidents illustrate vulnerabilities at every stage of the agent lifecycle introduced in Section~\ref{sec:preliminaries-agentic}:
Perceive---unauthenticated inputs and prompt injections entered without validation (Section~\ref{subsection:safety}~, input sanitization);
Plan---injected instructions altered the agent's plan to include exfiltration steps, with no constraint checker detecting the deviation;
Act---unrestricted tool access (file I/O, shell, and network) allowed the agent to execute attacker-controlled commands without least-privilege enforcement (Section~\ref{subsection:privacy});
Reflect---no anomaly detection flagged unusual credential access patterns or data transmission volumes;
Learn---malicious skills propagated through the update/install mechanism without provenance checks or regression gating.
This mapping demonstrates that the chain of failure is systemic; no single-stage mitigation would have been sufficient, reinforcing the defense-in-depth principle advocated in Section~\ref{subsection:safety}.

\subsubsection{Lessons learned}
\textbf{Table~\ref{tab:casestudy-lessons}} summarizes the key lessons and their mapping to the survey's trust dimensions and lifecycle stages.

These incidents underscore that agentic AI security must be treated as privileged infrastructure rather than consumer software; mandatory authentication, least-privilege scoping, plugin governance with cryptographic signing, continuous runtime monitoring, and explicit human approval for sensitive actions are not optional hardening steps but baseline requirements~\citep{clawdbot2025security,Willison2025LethalTrifecta}.

\section{Conclusions}
Agentic AI systems---LLMs augmented with planning, tool use, memory, and long-horizon interactions---are moving from passive assistants to autonomous operators. This shift enlarges the risk surface: failures may emerge anywhere along a multi-step trajectory (e.g., perception/retrieval, planning, tool execution, reflection, and learning) and can cascade into high-impact outcomes. To support trustworthy development and deployment in high-risk scenarios, this survey organized the literature through two complementary lenses: (i) a focused examination of two core trustworthiness dimensions---Safety and Robustness and Privacy and System Security---which are most critical for high-stakes deployments (Section~\ref{section:core_dimensions}), (ii) a Perceive$\rightarrow$Plan$\rightarrow$Act$\rightarrow$Reflect$\rightarrow$Learn workflow that localizes where risks arise and where mitigations can be applied. Across these core dimensions, we summarized stage-targeted methods and highlighted the system-level nature of trustworthiness in tool-using, memory-enabled, and multi-agent settings.

To make evaluations comparable and operational, we consolidated fragmented metrics and benchmark suites into a unified ``evaluation hub'' (Section~\ref{section:eval}), emphasizing that trustworthy agents must be assessed not only by final task success, but also by process signals such as constraint violations, trace completeness, compliance verifiability, and adversarial success rates. We hope this survey serves as a practical reference for researchers and practitioners building agentic systems that are not only capable, but also auditable, robust, and safe under real-world constraints.

\subsection{Summary of open problems}
Despite rapid progress, important limitations remain before trustworthy agentic AI can be deployed at scale. First, today's agents are still far from reliable lifelong learners: they struggle with long-horizon generalization, are vulnerable to reward/specification gaming (Section~\ref{subsection:safety}), and can accumulate errors across extended tool-mediated trajectories. Second, verification and evaluation remain bottlenecked by long-tail and interactive risks; simulations and testbeds rarely cover rare events, adaptive attacks, and sim-to-real gaps, while many evaluations depend on imperfect judges and incomplete trace standards (Section~\ref{section:eval}). Third, deployment introduces persistent trade-offs and system risks, including privacy--utility tension in memory-based personalization (Section~\ref{subsection:privacy}) and security vulnerabilities from tool access and credential management. Finally, frequent re-verification, monitoring, and safeguarding can impose substantial compute and operational costs.

These limitations motivate several concrete research and engineering challenges: (i) self-evolving agents with safe update rules and regression-gated releases; (ii) continuous runtime monitoring and verification that can detect rare-event and adversarial failures early; (iii) privacy-preserving personalization and memory governance that bound leakage while preserving utility; (iv) efficiency-aware trust mechanisms that deliver guarantees under realistic latency/energy budgets; (v) principled resolution of the trust--utility trade-off via risk-bounded, scenario-driven evaluation and deployment practices.

\acknowledgments{The authors would like to thank the anonymous reviewers for their constructive feedback, which significantly improved this manuscript. The authors acknowledge the use of Chat GPT-5 (OpenAI, San Francisco, CA, USA) for language polishing and grammar checking. The tool was applied to enhance clarity and scholarly tone, and all edits were carefully reviewed and validated by the authors to ensure accuracy, compliance with academic standards, and research integrity. The authors fully support Academia.edu Journals’ adherence to COPE guidelines on AI in publication ethics and confirm that this use has been managed responsibly and ethically.} 

\funding{This research received no external funding.}

\authorcontributions{Conceptualization, J.Q. and I.K.; methodology, J.Q.; investigation, J.Q., M.L., J.L., D.Y., S.M., Y.S., W.C., Y.Z., Y.C. and R.J.; writing---original draft, J.Q., M.L., J.L., D.Y. and S.M.; writing---review and editing, J.Q., Y.S., W.C., Y.Z., Y.C., R.J., I.K. and Z.X.; visualization, J.Q.; supervision, I.K. and Z.X. All authors have read and agreed to the published version of the manuscript.}

\conflictsofinterest{Author Irwin King serves as Founding Advisory Editor and author Zenglin Xu serves as Editor-in-Chief of \textit{Academia AI and Applications} journal. To ensure editorial independence, neither was involved in the peer-review or editorial decision-making process for this manuscript; handling was managed entirely by independent Editorial Board Members. All other authors declare no conflicts of interest.}

\dataavailability{No new data were created or analyzed in this study. Data sharing is not applicable to this article.}

\history{}

\PublishersNote

\cright






\begin{thebibliography}{999}

\providecommand{\natexlab}[1]{#1}
\providecommand{\url}[1]{\texttt{#1}}
\expandafter\ifx\csname urlstyle\endcsname\relax
  \providecommand{\doi}[1]{doi: #1}\else
  \providecommand{\doi}{doi: \begingroup \urlstyle{rm}\Url}\fi

\makeatletter
\renewcommand{\@biblabel}[1]{[#1]} 
\renewcommand{\@biblabel}[1]{#1.}  
\makeatother

\bibitem[Xi et~al.(2025)Xi, Chen, Guo, He, Ding, Hong, Zhang, Wang, Jin, Zhou,
  Zheng, Fan, Wang, Xiong, Zhou, Wang, Jiang, Zou, Liu, Yin, Dou, Weng, Cheng,
  Zhang, Qin, Zheng, Qiu, Huang, and Gui]{xi2023rise}
Xi Z, Chen W, Guo X, He W, Ding Y, Hong B, et al.
\newblock The rise and potential of large language model based agents: a
  survey.
\newblock {Sci China Inf Sci.} 2025;68\penalty0 (2):121101.
\newblock \doi{10.1007/s11432-024-4222-0}

\bibitem[Chen et~al.(2025{\natexlab{a}})Chen, Zou, Wang, Wang, Sun, Chi, and
  Xu]{chen-etal-2025-finhear}
Chen J, Zou M, Wang Z, Wang Q, Sun DD, Chi Z, et al.
\newblock {F}in{HEAR}: human expertise and adaptive risk-aware temporal
  reasoning for financial decision-making.
\newblock {Findings of the association for computational linguistics:
  EMNLP 2025}.
  Suzhou: Association for Computational Linguistics; 2025. p. 1648--72.
\newblock \doi{10.18653/v1/2025.findings-emnlp.87}

\bibitem[Chen et~al.(2025{\natexlab{b}})Chen, Yang, Li, Liu, Luan, and
  Xu]{chen2025heterogeneousgroupbasedreinforcementlearning}
Chen G, Yang S, Li C, Liu W, Luan J, Xu Z.
\newblock Heterogeneous group-based reinforcement learning for llm-based
  multi-agent systems. arXiv; 2025 [accessed on 15 January 2026].
 Available from:
\newblock  \url{https://arxiv.org/abs/2506.02718}

\bibitem[Lakshmanan(2025)]{Lakshmanan2025EchoLeak}
Lakshmanan R.
\newblock Zero-click ai vulnerability exposes microsoft 365 copilot data
  without user interaction.
\newblock The Hacker News.  12 June 2025 [accessed on 31 December 2025].
\newblock Available from:
  \url{https://thehackernews.com/2025/06/zero-click-ai-vulnerability-exposes.html}

\bibitem[Paverd(2025)]{Paverd2025IndirectPI}
Paverd A.
\newblock How microsoft defends against indirect prompt injection attacks.
\newblock Microsoft Security Response Center (MSRC) Blog.  29 July 2025 [accessed on 31 December 2025].
\newblock Available from:
  \url{https://www.microsoft.com/en-us/msrc/blog/2025/07/how-microsoft-defends-against-indirect-prompt-injection-attacks}

\bibitem[Greshake et~al.(2023{\natexlab{a}})Greshake, Abdelnabi, Mishra,
  Endres, Holz, and Fritz]{Greshake2023IndirectPI}
Greshake K, Abdelnabi S, Mishra S, Endres C, Holz T, Fritz M.
\newblock Not what you've signed up for: compromising real-world
  {LLM}-integrated applications with indirect prompt injection.
\newblock  {Proceedings of the 16th ACM Workshop on Artificial
  Intelligence and Security (AISec), Co-Located with ACM CCS 2023};
  30 November 2023; Copenhagen, Denmark. 2023. 
\newblock Best Paper Award.
\newblock \doi{10.1145/3605764.3623985}

\bibitem[{OWASP}(2024)]{OWASPPromptInjection}
{OWASP}.
\newblock Llm01: prompt INJECTION---owasp genai security project.
\newblock online. 2024 [accessed on 31 December 2025].
\newblock Available from:
  \url{https://genai.owasp.org/llmrisk2023-24/llm01-24-prompt-injection/}

\bibitem[Choi et~al.(2025)Choi, Balasubramanian, Qi, and
  Ferrara]{10.1145/3701716.3715521}
Choi EC, Balasubramanian A, Qi J, Ferrara E.
\newblock Limited effectiveness of llm-based data augmentation for COVID-19
  misinformation stance detection.
\newblock  In Companion Proceedings of the ACM on Web Conference 2025,
  WWW '25; New York (NY): Association for Computing
  Machinery; 2025. p. 934--7. 
\newblock ISBN 9798400713316.
\newblock \doi{10.1145/3701716.3715521}

\bibitem[Li et~al.(2025{\natexlab{a}})Li, Qi, Wu, Zhao, Ma, Li, Wang, Zhang,
  fung Leung, and King]{li2025evidencetrajectoryabductivereasoning}
Li M, Qi J, Wu Y, Zhao M, Ma L, Li Y,  et al.
\newblock From evidence to trajectory: abductive reasoning path synthesis for
  training retrieval-augmented generation agents. 2025 [accessed on 15 January 2026].
\newblock Available from: \url{https://arxiv.org/abs/2509.23071}

\bibitem[Wang et~al.(2023)Wang, Xie, Jiang, Mandlekar, Xiao, Zhu, Fan, and
  Anandkumar]{Wang2023Voyager}
Wang G, Xie Y, Jiang Y, Mandlekar A, Xiao C, Zhu Y, et al.   
\newblock Voyager: an open-ended embodied agent with large language models.
\newblock {Transactions on machine learning research (TMLR)}; 2023 [accessed on 15 January 2026].
\newblock Available from: \url{https://openreview.net/forum?id=ehfRiF0R3a}

\bibitem[Packer et~al.(2024)Packer, Wooders, Lin, Fang, Patil, Stoica, and
  Gonzalez]{packer2023memgpt}
Packer C, Wooders S, Lin K, Fang V, Patil SG, Stoica I,  et al.
\newblock MemGPT: towards {LLMs} as operating systems.
\newblock  {The twelfth international conference on learning
  representations (ICLR)}. 2024 [accessed on 15 January 2026].
\newblock Available from: \url{https://openreview.net/forum?id=LeYFkQxaAK}

\bibitem[Zhang et~al.(2025{\natexlab{a}})Zhang, Hu, Upasani, Ma, Hong,
  Kamanuru, Rainton, Wu, Ji, Li, Thakker, Zou, and Olukotun]{Zhang2025Agentic}
Zhang Q, Hu C, Upasani S, Ma B, Hong F, Kamanuru V,  et al.
\newblock Agentic context engineering: evolving contexts for self-improving language models. 
arXiv; 2025 [accessed on October 2025]. Available from: \url{https://arxiv.org/abs/2510.04618}

\bibitem[Vallor and Vierkant(2024)]{accountability_overall_challenge1}
Vallor S, Vierkant T.
\newblock Find the gap: AI, responsible agency and vulnerability.
\newblock {Minds Mach.} 2024;34\penalty0 (3):\penalty0 20.
\newblock \doi{10.1007/s11023-024-09674-0}

\bibitem[Bagave et~al.(2025)Bagave, Westberg, Janssen, and
  Ding]{accountability_overall_challenge2}
Bagave P, Westberg M, Janssen M, Ding AY.
\newblock Accountability framework for healthcare ai systems: towards joint
  accountability in decision making.
\newblock  {Proceedings of the AAAI/ACM Conference on AI, Ethics, and
  Society};  20--22 October 2025; Madrid, Spain.  2025; Vol.~8, p. 279--91.

\bibitem[Gutfraind and Bier(2023)]{Gutfraind2023RiskUncertaintyAI}
Gutfraind A, Bier VM.
\newblock Risk, uncertainty and {AI}: non-probabilistic methods for
  anticipating and preventing {AI} risks.
\newblock Technical report, University of Illinois. 2023 [accessed on 15 January 2026].
\newblock Available from: \url{https://www.ideals.illinois.edu/items/129049}



\bibitem[Kaur et~al.(2022)Kaur, Uslu, Rittichier, and
  Durresi]{Kaur2022TrustworthyAIReview}
Kaur D, Uslu S, Rittichier KJ, Durresi A.
\newblock Trustworthy artificial intelligence: a review.
\newblock {ACM Comput Surv.} 2022;55\penalty0 (2):\penalty0 1--38.
\newblock \doi{10.1145/3491209}

\bibitem[Liu et~al.(2024)Liu, Yao, Ton, Zhang, Guo, Cheng, Klochkov, Taufiq,
  and Li]{Liu2023TrustworthyLLMs}
Liu Y, Yao Y, Ton J-F, Zhang X, Guo R, Cheng H,  et al
\newblock Trustworthy llms: a survey and guideline for evaluating large
  language models' alignment. 2024 [accessed on 15 January 2026].
\newblock Available from: \url{https://arxiv.org/abs/2308.05374}

\bibitem[Huang et~al.(2024)Huang, Sun, Wang, Wu, Zhang, Li, Gao, Huang, Lyu,
  Zhang, Li, Liu, Liu, Wang, Zhang, Vidgen, Kailkhura, Xiong, Xiao, Li, Xing,
  Huang, Liu, Ji, Wang, Zhang, Yao, Kellis, Zitnik, Jiang, Bansal, Zou, Pei,
  Liu, Gao, Han, Zhao, Tang, Wang, Vanschoren, Mitchell, Shu, Xu, Chang, He,
  Huang, Backes, Gong, Yu, Chen, Gu, Xu, Ying, Ji, Jana, Chen, Liu, Zhou, Wang,
  Li, Zhang, Wang, Xie, Chen, Wang, Liu, Ye, Cao, Chen, and
  Zhao]{Huang2024TrustLLM}
Huang Y, Sun L, Wang H, Wu S, Zhang Q, Li Y, et al.
\newblock {TrustLLM}: trustworthiness in large language models.
\newblock  {Proceedings of the 41st International Conference on Machine
  Learning (ICML)}; 21--27 July 2024; Vienna, Austria.  2024 [accessed on 15 January 2026].
\newblock Available from: \url{https://proceedings.mlr.press/v235/huang24x.html}

\bibitem[Yu et~al.(2025{\natexlab{a}})Yu, Meng, Zhou, Wang, Mao, Pang, Chen,
  Wang, Li, Zhang, An, and Wen]{Yu2025TrustworthyAgents}
Yu M, Meng F, Zhou X, Wang S, Mao J, Pang L, et al.
\newblock A survey on trustworthy llm agents: threats and countermeasures.
\newblock  {Proceedings of the 31st ACM SIGKDD Conference on Knowledge
  Discovery and Data Mining (KDD '25)};  3--7 August 2025; Toronto, ON, Canada. 2025; p. 6216--26.
\newblock \doi{10.1145/3711896.3736561}

\bibitem[Ma et~al.(2025{\natexlab{a}})Ma, Gao, Wang, Wang, Wang, Sun, Ding, Xu,
  Chen, Zhao, Huang, Li, Zhang, Zheng, Bai, Wu, Qiu, Zhang, Li, Sun, Wang, Gu,
  Wu, Chen, Zhang, Liu, Gong, Liu, Pan, Xie, Pang, Dong, Jia, Zhang, Ma, Zhang,
  Gong, Xiao, Erfani, Li, Sugiyama, Tao, Bailey, and
  Jiang]{Ma2025SafetyAtScale}
Ma X, Gao Y, Wang Y, Wang R, Wang X, Sun Y, et al.
\newblock Safety at scale: a comprehensive survey of large model and agent
  safety, 2025{\natexlab{a}} [accessed on 15 January 2026].
\newblock Available from: \url{https://arxiv.org/abs/2502.05206}

\bibitem[Abou~Ali et~al.(2025{\natexlab{a}})Abou~Ali, Dornaika, and
  Charafeddine]{AbouAli2025AgenticAI}
Ali MA, Dornaika F, Charafeddine J.
\newblock Agentic {AI}: a comprehensive survey of architectures, applications,
  and future directions.
\newblock {Artif Intell Rev.} 2025;59\penalty0 (1):11.
\newblock \doi{10.1007/s10462-025-11422-4}

\bibitem[Xu et~al.(2025)Xu, Huang, Gao, and Shang]{Xu2025ToolLearningSurvey}
Xu W, Huang C, Gao S, Shang S.
\newblock Llm-based agents for tool learning: a survey.
\newblock {Data Sci Eng.}  2025;10(4):533--63.
\newblock \doi{10.1007/s41019-025-00296-9}.

\bibitem[Russell and Norvig(2021)]{RussellNorvig2021}
Russell S, Norvig P.
\newblock {Artificial intelligence: a modern approach}. 4th ed.
\newblock  London: Pearson; 2021.

\bibitem[Kaelbling et~al.(1998)Kaelbling, Littman, and
  Cassandra]{Kaelbling1998AIJ}
Kaelbling LP, Littman ML, Cassandra AR.
\newblock Planning and acting in partially observable stochastic domains.
\newblock {Artif Intell.} 1998;101\penalty0 (1--2):\penalty0
  99--134.
\newblock \doi{10.1016/S0004-3702(98)00023-X}.

\bibitem[Sutton and Barto(2018)]{SuttonBarto2018}
Sutton RS, Barto AG.
\newblock {Reinforcement learning: an introduction}. 2nd ed.
\newblock  Cambridge (MA): MIT Press; 2018.

\bibitem[Yao et~al.(2023)Yao, Zhao, Yu, Du, Shafran, Narasimhan, and
  Cao]{Yao2023ReAct}
Yao S, Zhao J, Yu D, Du N, Shafran I, Narasimhan K, et al. 
\newblock {ReAct}: synergizing reasoning and acting in language models.
\newblock {The eleventh international conference on learning
  representations (ICLR)}. 2023 [accessed on 15 January 2026].
\newblock Available from: \url{https://openreview.net/forum?id=WE_vluYUL-X}

\bibitem[Park et~al.(2023)Park, O{\textquoteright}Brien, Cai, Ringel~Morris,
  Liang, and Bernstein]{Park2023GenerativeAgents}
Park JS,  O{\textquoteright}Brien JC, Cai CJ, Morris MR, Liang P, Bernstein MS.
\newblock Generative agents: interactive simulacra of human behavior.
\newblock {Proceedings of the 36th Annual ACM Symposium on User
  Interface Software and Technology (UIST)}; 29 October--1 November 2023; San Francisco, CA, USA. 2023.
\newblock \doi{10.1145/3586183.3606763}

\bibitem[{National Institute of Standards and Technology}(2023)]{NIST2023RMF}
{National Institute of Standards and Technology}.
\newblock Artificial intelligence risk management framework (AI RMF 1.0).
\newblock Technical report NIST AI 100-1; National Institute of Standards and
  Technology (NIST).  
\newblock Voluntary framework for managing AI risks, guidance for trustworthy
  AI systems. 2023 [accessed on 15 January 2026].
\newblock Available from: \url{https://nvlpubs.nist.gov/nistpubs/ai/nist.ai.100-1.pdf}

\bibitem[Lewis et~al.(2020)Lewis, Perez, Piktus, Petroni, Karpukhin, Goyal,
  K\"{u}ttler, Lewis, Yih, Rockt\"{a}schel, Riedel, and Kiela]{Lewis2020RAG}
Lewis P, Perez E, Piktus A, Petroni F, Karpukhin V, Goyal N,  et al.
\newblock Retrieval-augmented generation for knowledge-intensive nlp tasks.
\newblock {Advances in neural information processing systems 33}. Red Hook (NY): Curran Associates, Inc.; 2020. p.  9459--74. [accessed on 15 January 2026].
\newblock Available from:
  \url{https://proceedings.neurips.cc/paper/2020/file/6b493230205f780e1bc26945df7481e5-Paper.pdf}

\bibitem[Ha and Schmidhuber(2018)]{Ha2018WorldModels}
 Ha D,  Schmidhuber J.
\newblock World models.
\newblock Proceedings of the {Advances in Neural Information Processing Systems 31
  (NeurIPS)};  3--8 December 2018; Montréal, Canada. 2018.

\bibitem[Sutton et~al.(1999)Sutton, Precup, and Singh]{Sutton1999Options}
Sutton RS, Precup D, Singh S.
\newblock Between {MDP}s and semi-{MDP}s: a framework for temporal abstraction
  in reinforcement learning.
\newblock {Artif Intell.} 1999;112\penalty0 (1--2):\penalty0
  181--211.
\newblock \doi{10.1016/S0004-3702(99)00052-1}

\bibitem[Schick et~al.(2023)Schick, Dwivedi-Yu, Dess\'{\i}, Raileanu, Lomeli,
  Hambro, Zettlemoyer, Cancedda, and Scialom]{Schick2023Toolformer}
Schick T,   Dwivedi-Yu J,  Dess\'{\i} R, Raileanu R, Lomeli M, Hambro E,  et al.
\newblock Toolformer: language models can teach themselves to use tools.
\newblock In {Proceedings of the 37th International Conference on Neural
  Information Processing Systems}, NIPS '23;  10--16 December 2023; 
  New Orleans, LA, USA. Red Hook (NY): Curran Associates Inc.; 2023.

\bibitem[Miao et~al.(2025)Miao, Zou, Li, Chen, Wang, Wang, Li, Yang, He, Zhang,
  Yu, Yang, Nguyen, Zhou, Yang, Guo, Fan, Yeh, Meng, Fang, Qi, Huang, Gu, Han,
  He, Yang, Li, Zheng, Liu, King, and Yu]{miao2025recodehbenchmarkresearchcode}
Miao C, Zou HP, Li Y, Chen Y, Wang Y, Wang F, et al.   
\newblock Recode-h: a benchmark for research code development with interactive
  human feedback. 2025 [accessed on 15 January 2026].
\newblock Available from: \url{https://arxiv.org/abs/2510.06186}

\bibitem[Shinn et~al.(2023)Shinn, Cassano, Berman, Gopinath, Narasimhan, and
  Yao]{Shinn2023Reflexion}
Shinn N, Cassano F, Berman E, Gopinath A, Narasimhan K, Yao S.
\newblock Reflexion: language agents with verbal reinforcement learning.
\newblock  {Advances in neural information processing systems 36
  (NeurIPS)}. 2023 [accessed on 15 January 2026].
\newblock Available from: \url{https://openreview.net/forum?id=vAElhFcKW6}

\bibitem[Zhao et~al.(2024{\natexlab{a}})Zhao, Zhou, Li, Tang, Wang, Hou, Min,
  Zhang, Zhang, Dong, Du, Yang, Chen, Chen, Jiang, Ren, Li, Tang, Liu, Liu,
  Nie, and Wen]{Zhao2023LLMSurvey}
Zhao WX, Zhou K, Li J, Tang T, Wang X, Hou Y, et al. 
\newblock A survey of large language models.
\newblock {IEEE Access}. 2024{\natexlab{a}} [accessed on 15 January 2026].
 Available from: \url{https://arxiv.org/abs/2303.18223}

\bibitem[Puterman(1994)]{Puterman1994}
Puterman ML. 
\newblock {Markov decision processes: discrete stochastic dynamic programming}.
\newblock Wiley Series in Probability and Statistics.  Hoboken (NJ): John Wiley \& Sons; 1994.
\newblock ISBN 9780471619772.

\bibitem[Zhang et~al.(2021)Zhang, Yang, and Basar]{Zhang2021MARLSurvey}
Zhang K, Yang Z, Basar T.
\newblock Multi-agent reinforcement learning: a selective overview of theories
  and algorithms.
\newblock {Handbook of reinforcement learning and control}. Cham: Springer; 2021.
\newblock \doi{10.1007/978-3-030-60990-0_12}

\bibitem[Schulman et~al.(2017)Schulman, Wolski, Dhariwal, Radford, and
  Klimov]{Schulman2017PPO}
Schulman J, Wolski F, Dhariwal P, Radford A, Klimov O.
\newblock Proximal policy optimization algorithms. 2017 [accessed on 15 January 2026].
Available from: \url{https://arxiv.org/abs/1707.06347}

\bibitem[Mnih et~al.(2015)Mnih, Kavukcuoglu, Silver, Rusu, Veness, Bellemare,
  Graves, Riedmiller, Fidjeland, Ostrovski, Petersen, Beattie, Sadik,
  Antonoglou, King, Kumaran, Wierstra, Legg, and Hassabis]{Mnih2015DQN}
Mnih V, Kavukcuoglu K, Silver D, Rusu AA, Veness J, Bellemare MG,  et al.
\newblock Human-level control through deep reinforcement learning.
\newblock {Nature}. 2015;518\penalty0 (7540):\penalty0 529--33.
\newblock \doi{10.1038/nature14236}

\bibitem[Levine et~al.(2020)Levine, Kumar, Tucker, and Fu]{Levine2020Offline}
Levine S, Kumar A, Tucker G, Fu J.
\newblock Offline reinforcement learning: tutorial, review, and perspectives on
  open problems. 2020 [accessed on 15 January 2026].
Available from: \url{https://arxiv.org/abs/2005.01643}

\bibitem[Kumar et~al.(2020)Kumar, Zhou, Tucker, and Levine]{Kumar2020CQL}
Kumar A, Zhou A, Tucker G, Levine S.
\newblock Conservative {Q}-learning for offline reinforcement learning.
\newblock {Advances in neural information processing systems 33}. Red Hook (NY):
  Curran Associates, Inc.; 2020. p. 1179--91. [accessed on 15 January 2026].
\newblock Available from:
  \url{https://papers.nips.cc/paper/2020/hash/0d2b2061826a5df3221116a5085a6052-Paper.pdf}

\bibitem[Nachum et~al.(2018)Nachum, Gu, Lee, and Levine]{Nachum2018HIRO}
Nachum O, Gu S, Lee H, Levine S.
\newblock Data-efficient hierarchical reinforcement learning.
\newblock {Advances in neural information processing systems 31
  (NeurIPS 2018)}. Red Hook (NY): Curran Associates, Inc.;  2018. p. 3307--17. [accessed on 15 January 2026].
\newblock Available from:
  \url{http://papers.nips.cc/paper/7591-data-efficient-hierarchical-reinforcement-learning.pdf}

\bibitem[Chua et~al.(2018)Chua, Calandra, McAllister, and Levine]{Chua2018PETS}
Chua K, Calandra R, McAllister R, Levine S.
\newblock Deep reinforcement learning in a handful of trials using
  probabilistic dynamics models.
\newblock {NeurIPS}. 2018 [accessed on 15 January 2026].
\newblock Available from:   \url{https://arxiv.org/abs/1805.12114}

\bibitem[Janner et~al.(2019)Janner, Fu, Zhang, and Levine]{Janner2019MBPO}
Janner M, Fu J, Zhang M, Levine S.
\newblock When to trust your model: model-based policy optimization.
\newblock {NeurIPS}. 2019 [accessed on 15 January 2026].
\newblock Available from:   \url{https://arxiv.org/abs/1906.08253}

\bibitem[Altman(1999)]{Altman1999}
 Altman E.
\newblock {Constrained markov decision processes}.
\newblock Boca Raton (FL): Chapman \& Hall/CRC; 1999.
\newblock ISBN 9780849303821.

\bibitem[Achiam et~al.(2017)Achiam, Held, Tamar, and Abbeel]{Achiam2017CPO}
Achiam J, Held D, Tamar A, Abbeel P.
\newblock Constrained policy optimization.
\newblock In: Precup D, Teh YW, editors. {Proceedings of the
  34th International Conference on Machine Learning (ICML 2017)}. Vol.~70. 
  {Proceedings of Machine Learning Research}.  PMLR. 2017. p. 22--31. [accessed on 15 January 2026].
\newblock Available from: \url{https://proceedings.mlr.press/v70/achiam17a.html}

\bibitem[Garc{\'\i}a and Fern{\'a}ndez(2015)]{Garcia2015SafeRL}
 Garc{\'\i}a J,  Fern{\'a}ndez F.
\newblock A comprehensive survey on safe reinforcement learning.
\newblock {J Mach Learn Res.} 2015;16\penalty0
  (1):\penalty0 1437--80.
\newblock \doi{10.5555/2886795}

\bibitem[Alshiekh et~al.(2018)Alshiekh, Bloem, Ehlers, K\"{o}nighofer, Niekum,
  and Topcu]{10.5555/3504035.3504361}
Alshiekh M, Bloem R,   Ehlers R,  K\"{o}nighofer B, Niekum S, Topcu U.
\newblock Safe reinforcement learning via shielding.
\newblock   {Proceedings of the Thirty-Second AAAI Conference on
  Artificial Intelligence};  New Orleans (LA): AAAI
  Press; 2018.  p. 2669--78. [accessed on 15 January 2026].
\newblock Available from: \url{https://ojs.aaai.org/index.php/AAAI/article/view/11797}
\newblock \doi{10.1609/aaai.v32i1.11797}

\bibitem[Christiano et~al.(2017)Christiano, Leike, Brown, Martic, Legg, and
  Amodei]{Christiano2017}
Christiano PF, Leike J, Brown TB, Martic M, Legg S, Amodei D.
\newblock Deep reinforcement learning from human preferences.
\newblock  {Advances in neural information processing systems 30}. 
Red Hook (NY): Curran Associates, Inc.; 2017. p.  4299--307.  [accessed on 15 January 2026].
\newblock Available from:
  \url{http://papers.nips.cc/paper/7017-deep-reinforcement-learning-from-human-preferences.pdf}

\bibitem[Ziegler et~al.(2019)Ziegler, Stiennon, Wu, Brown, Radford, Amodei,
  Christiano, and Irving]{Ziegler2019}
Ziegler DM, Stiennon N, Wu J, Brown TB, Radford A, Amodei D,  et al.
\newblock Fine-tuning language models from human preferences.
\newblock {Advances in neural information processing systems 32
  (NeurIPS)}. 2019. [accessed on 15 January 2026].
\newblock Available from: \url{https://arxiv.org/abs/1909.08593}

\bibitem[Stiennon et~al.(2020)Stiennon, Ouyang, Wu, Ziegler, Lowe, Voss,
  Radford, Amodei, and Christiano]{Stiennon2020}
Stiennon N, Ouyang L, Wu J, Ziegler DM, Lowe R, Voss C,  et al.
\newblock Learning to summarize from human feedback.
\newblock {Proceedings of the 34th international conference on neural
  information processing systems}. Red Hook (NY): Curran Associates, Inc.; 2020.
   p. 4302--10. [accessed on 15 January 2026].
\newblock Available from:
  \url{https://proceedings.neurips.cc/paper/2020/file/1f89885d556929e98d3ef9b86448f951-Paper-Conference.pdf}

\bibitem[Ouyang et~al.(2022)Ouyang, Wu, Jiang, Almeida, Wainwright, Mishkin,
  Zhang, Agarwal, Slama, Ray, Schulman, Hilton, Kelton, Miller, Simens, Askell,
  Welinder, Christiano, Leike, and Lowe]{Ouyang2022}
Ouyang L, Wu J, Jiang X, Almeida D, Wainwright C, Mishkin P,  et al.
\newblock Training language models to follow instructions with human feedback.
\newblock In: Koyejo S, Mohamed S, Agarwal A, Belgrave D, Cho K, Oh A, 
  editors. {Advances in neural information processing systems}. Vol.~35.
  Red Hook (NY):  Curran Associates, Inc.; 2022. p. 27730--44.  [accessed on 15 January 2026].
\newblock Available from:
  \url{https://proceedings.neurips.cc/paper_files/paper/2022/file/b1efde53be364a73914f58805a001731-Paper-Conference.pdf}

\bibitem[Bai et~al.(2022{\natexlab{a}})Bai, Kadavath, Kundu, Askell, Kernion,
  Jones, Chen, Goldie, Mirhoseini, McKinnon, Chen, Olsson, Olah, Hernandez,
  Drain, Ganguli, Li, Tran-Johnson, Perez, Kerr, Mueller, Ladish, Landau,
  Ndousse, Lukosuite, Lovitt, Sellitto, Elhage, Schiefer, Mercado, DasSarma,
  Lasenby, Larson, Ringer, Johnston, Kravec, Showk, Fort, Lanham,
  Telleen-Lawton, Conerly, Henighan, Hume, Bowman, Hatfield-Dodds, Mann,
  Amodei, Joseph, McCandlish, Brown, and Kaplan]{bai2024constitutional}
Bai Y, Kadavath S, Kundu S, Askell A, Kernion J, Jones A,  et al.
\newblock Constitutional AI: harmlessness from AI feedback. 2022{\natexlab{a}} [accessed on 15 January 2026].
\newblock Available from: \url{https://arxiv.org/abs/2212.08073}

\bibitem[Rodriguez-Soto et~al.(2025)]{RodriguezSoto2025MORL}
 Rodriguez-Soto M, Rădulescu R, Bistaffa F, Ricart O, Mayoral-Macau A,  et~al.
\newblock Multi-objective reinforcement learning for provably incentivising
  alignment with value systems.
\newblock {Artif Intell.} 2025;351:104460.
\newblock \doi{10.1016/j.artint.2025.104460}



\bibitem[Mayoral~Macau et~al.(2025)]{MayoralMacau2025ECAI}
 Mayoral~Macau A,  Rodríguez-Soto M, Marchesini E, Sánchez-Fibla M, López-Sánchez M, Rodríguez-Aguilar JA,   et~al.
\newblock An approximate embedding for designing ethical reinforcement learning
  environments.
\newblock  {Proceedings of the 28th European conference on artificial
  intelligence (ECAI)}, 2025 [accessed on 15 January 2026].
\newblock Available from: \url{https://ebooks.iospress.nl/volumearticle/76029}


\bibitem[Serramia et~al.(2023)]{Serramia2023MindsAndMachines}
 Serramia M, Rodriguez-Soto M, Lopez-Sanchez M, Rodriguez-Aguilar JA, Bistaffa F, Boddington P, et~al.
\newblock Encoding ethics to compute value-aligned norms.
\newblock {Minds  Mach.} 2023;33(4):761--90.
\newblock \doi{10.1007/s11023-023-09649-7}


\bibitem[Rafailov et~al.(2023)Rafailov, Sharma, Mitchell, Manning, Ermon, and
  Finn]{Rafailov2023DPO}
Rafailov R, Sharma A, Mitchell E, Manning CD, Ermon S, Finn C.
\newblock Direct preference optimization: Your language model is secretly a
  reward model.
\newblock {Thirty-seventh conference on neural information processing
  systems}. 2023 [accessed on 15 January 2026].
\newblock Available from: \url{https://openreview.net/forum?id=HPuSIXJaa9}

\bibitem[Ethayarajh et~al.(2024)Ethayarajh, Xu, Muennighoff, Jurafsky, and
  Kiela]{Ethayarajh2024KTO}
Ethayarajh K, Xu W, Muennighoff N, Jurafsky D, Kiela D.
\newblock Model alignment as prospect theoretic optimization.
\newblock {Proceedings of the 41st International Conference on Machine
  Learning}, ICML'24. JMLR.org;  21--27 July 2024; Vienna, Austria.  2024.

\bibitem[Casper et~al.(2023)Casper, Davies, Shi, Gilbert, Scheurer, Rando,
  Freedman, Korbak, Lindner, Freire, Wang, Marks, S{\'e}gerie, Carroll, Peng,
  Christoffersen, Damani, Slocum, Anwar, Siththaranjan, Nadeau, Michaud, Pfau,
  Krasheninnikov, Chen, Langosco, Hase, Biyik, Dragan, Krueger, Sadigh, and
  Hadfield-Menell]{Casper2023OpenProblems}
Casper S, Davies X, Shi C, Gilbert TK, Scheurer J, Rando J, et al.
\newblock Open problems and fundamental limitations of reinforcement learning
  from human feedback.
\newblock {Transactions on machine learning research (TMLR)}. 2023  [accessed on 15 January 2026].
\newblock Available from: \url{https://openreview.net/forum?id=bx24KpJ4Eb}

\bibitem[Tobin et~al.(2017{\natexlab{a}})Tobin, Fong, Ray, Schneider, Zaremba,
  and Abbeel]{Tobin2017DR}
Tobin J, Fong R, Ray A, Schneider J, Zaremba W, Abbeel P.
\newblock Domain randomization for transferring deep neural networks from
  simulation to the real world.
\newblock Proceedings of the {2017 IEEE/RSJ International Conference on Intelligent Robots
  and Systems (IROS)}; 24--28 September 2017; Vancouver, BC, Canada.  
 2017{\natexlab{a}}. p. 23--30.
\newblock \doi{10.1109/IROS.2017.8202133}

\bibitem[Amodei et~al.(2016)Amodei, Olah, Steinhardt, Christiano, Schulman, and
  Man{\'e}]{Amodei2016Concrete}
Amodei D, Olah C, Steinhardt J, Christiano P, Schulman J, Man{\'e} D.
\newblock Concrete problems in {AI} safety.
\newblock arXiv; 2016 [accessed on 15 January 2026].
 Available from: \url{https://arxiv.org/abs/1606.06565}

\bibitem[Yadgaroff et~al.(2023)Yadgaroff, Sestini, Tollmar, Ozcelikkale, and
  Gisslén]{yadgaroff2024improvinggeneralizationgameagents}
Yadgaroff D, Sestini A, Tollmar K, Ozcelikkale A, Gisslén L.
\newblock Improving generalization in game agents with data augmentation in
  imitation learning. 2023 [accessed on 15 January 2026].
\newblock Available from: \url{https://arxiv.org/abs/2309.12815}

\bibitem[Mazeika et~al.(2024)Mazeika, Phan, Yin, Zou, Wang, Mu, Sakhaee, Li,
  Basart, Li, Forsyth, and Hendrycks]{HarmBench}
Mazeika M, Phan L, Yin X, Zou A, Wang Z, Mu N,  et al.
\newblock Harmbench: a standardized evaluation framework for automated red
  teaming and robust refusal.
\newblock {Proceedings of the 41st International Conference on Machine
  Learning}, ICML'24. JMLR.org; 21--27 July 2024; Vienna, Austria.  2024.

\bibitem[Haider et~al.(2023)Haider, Roscher, Schmoeller~da Roza, and
  G\"{u}nnemann]{10.5555/3545946.3598721}
Haider T, Roscher K,  Schmoeller~da Roza F,  G\"{u}nnemann S.
\newblock Out-of-distribution detection for reinforcement learning agents with
  probabilistic dynamics models.
\newblock {Proceedings of the 2023 International Conference on
  Autonomous Agents and Multiagent Systems}; Richland (SC):
   International Foundation for Autonomous Agents and Multiagent Systems; 2023. p. 851--9. [accessed on 15 January 2026].
\newblock Available from:
  \url{https://www.ifaamas.org/Proceedings/aamas2023/pdfs/p851.pdf}

\bibitem[Sagawa et~al.(2020)Sagawa, Koh, Hashimoto, and
  Liang]{Sagawa2020GroupDRO}
Sagawa S, Koh PW, Hashimoto TB, Liang P.
\newblock Distributionally robust neural networks for group shifts: on the
  importance of regularization for worst-case generalization.
\newblock {International Conference on Learning Representations
  (ICLR)}. 2020 [accessed on 15 January 2026].
\newblock Available from: \url{https://openreview.net/forum?id=ryxGuJrFvS}

\bibitem[Bai et~al.(2022{\natexlab{b}})Bai, Kadavath, Kundu, Askell, Kernion,
  Jones, Chen, Goldie, Mirhoseini, McKinnon, Chen, Olsson, Olah, Hernandez,
  Drain, Ganguli, Li, Tran-Johnson, Perez, Kerr, Mueller, Ladish, Landau,
  Ndousse, Lukosuite, Lovitt, Sellitto, Elhage, Schiefer, Mercado, DasSarma,
  Lasenby, Larson, Ringer, Johnston, Kravec, Showk, Fort, Lanham,
  Telleen-Lawton, Conerly, Henighan, Hume, Bowman, Hatfield-Dodds, Mann,
  Amodei, Joseph, McCandlish, Brown, and Kaplan]{bai2022constitutional}
Bai Y, Kadavath S, Kundu S, Askell A, Kernion J, Jones A, et al. 
\newblock Constitutional AI: harmlessness from AI feedback. 2022{\natexlab{b}} [accessed on 15 January 2026].
\newblock Available from: \url{https://arxiv.org/abs/2212.08073}

\bibitem[Kushwaha et~al.(2025)Kushwaha, Ravish, Lamba, and
  Kumar]{kushwaha2025surveysafereinforcementlearning}
Kushwaha A, Ravish K, Lamba P, Kumar P.
\newblock A survey of safe reinforcement learning and constrained mdps: a
  technical survey on single-agent and multi-agent safety. 2025 [accessed on 15 January 2026].
\newblock Available from: \url{https://arxiv.org/abs/2505.17342}

\bibitem[Zheng et~al.(2025)Zheng, Liao, Salisbury, Liu, Lin, Zheng, Wang, Deng,
  Song, Sun, and Su]{zheng2025webguardbuildinggeneralizableguardrail}
Zheng B, Liao Z, Salisbury S, Liu Z, Lin M, Zheng Q, et al.
\newblock Webguard: building a generalizable guardrail for web agents. 2025 [accessed on 15 January 2026].
\newblock Available from: \url{https://arxiv.org/abs/2507.14293}

\bibitem[Duvvuru et~al.(2025)Duvvuru, Zhang, Vierhauser, and
  Agrawal]{10.1109/ICSE55347.2025.00223}
 Aswath Duvvuru VS, Zhang B, Vierhauser M, Agrawal A.
\newblock Llm-agents driven automated simulation testing and analysis of small
  uncrewed aerial systems.
\newblock {Proceedings of the IEEE/ACM 47th International Conference on
  Software Engineering}, ICSE '25; 27 April--3 May 2025; Ottawa, ON, Canada.  Hoboken (NJ):
IEEE Press; 2025. p. 385--97. 
\newblock ISBN 9798331505691.
\newblock \doi{10.1109/ICSE55347.2025.00223}

\bibitem[Pnueli(1977)]{4567924}
Pnueli A.
\newblock The temporal logic of programs.
\newblock Proceedings of the {18th Annual Symposium on Foundations of Computer Science
  (sfcs 1977)}; 31 October--2 November 1977; Providence, RI, USA.  1977; p. 46--57. 
\newblock \doi{10.1109/SFCS.1977.32}

\bibitem[Sculley et~al.(2015)Sculley, Holt, Golovin, Davydov, Phillips, Ebner,
  Chaudhary, Young, Crespo, and Dennison]{Sculley2015HiddenDebt}
Sculley D, Holt G, Golovin D, Davydov E, Phillips T, Ebner D, et al.
\newblock Hidden technical debt in machine learning systems.
\newblock In {Advances in neural information processing systems
  (NeurIPS)}.  Cambridge (MA): MIT Press; 2015. p. 2503--11. 

\bibitem[Breck et~al.(2017)Breck, Cai, Nielsen, Salib, and
  Sculley]{Breck2017MLTestScore}
Breck E, Cai S, Nielsen E, Salib M, Sculley D.
\newblock The {ML} test score: a rubric for {ML} production readiness and
  technical debt reduction.
\newblock Proceedings of the {2017 IEEE International Conference on Big Data (Big Data)};
   11--14 December 2017; Boston, MA, USA.  2017.
\newblock \doi{10.1109/BigData.2017.8258038}

\bibitem[Burns et~al.(2018)Burns, Feldman, Fletcher, Lin, Reynolds, Sanden,
  Wander, and Zienert]{Burns2018SpinnakerCD}
Burns E, Feldman A, Fletcher R, Lin T, Reynolds J, Sanden C, et al.
\newblock {Continuous delivery with spinnaker: fast, safe, repeatable
  multi-cloud deployments}.
\newblock Chapter 8: automated canary analysis.
\newblock Sebastopol (CA):  O'Reilly Media; 2018.

\bibitem[Laroche et~al.(2019)Laroche, Trichelair, and Combes]{Laroche2019SPIBB}
Laroche R, Trichelair P, des Combes RT.
\newblock Safe policy improvement with baseline bootstrapping.
\newblock {Proceedings of the 36th International Conference on Machine
  Learning (ICML)}; 9--15 June 2019; Long Beach, CA, USA.  Vol.~97 of 
  {proceedings of machine learning  research}. PMLR. 2019. p. 3652--61. 
   
\newblock \doi{10.1007/978-3-030-46133-1_4}.
\newblock Available from: \url{https://proceedings.mlr.press/v97/laroche19a.html}.

\bibitem[Dafoe et~al.(2020)Dafoe, Hughes, Bachrach, Collins, McKee, Leibo,
  Larson, and Graepel]{Dafoe2020Cooperative}
Dafoe A, Hughes E, Bachrach Y, Collins T, McKee KR, Leibo JZ, et al.
\newblock Open problems in cooperative {AI}. 2020 [accessed on 15 January 2026].
\newblock Available from: \url{https://arxiv.org/abs/2012.08630}

\bibitem[Zhang and Yang(2025)]{zhang2025searching}
Zhang Y, Yang D.
\newblock Searching for privacy risks in llm agents via simulation. 2025 [accessed on 15 January 2026].
\newblock Available from: \url{https://arxiv.org/abs/2508.10880}

\bibitem[Du et~al.(2025)Du, Li, Li, and Ding]{du2025beyond}
Du Y, Li Z, Li N, Ding B.
\newblock Beyond data privacy: new privacy risks for large language models.
  2025 [accessed on 15 January 2026].
\newblock Available from: \url{https://arxiv.org/abs/2509.14278}

\bibitem[Rose et~al.(2020)Rose, Borchert, Mitchell, and
  Connelly]{nist2020zerotrust}
Rose S, Borchert O, Mitchell S, Connelly S.
\newblock Zero trust architecture.
\newblock Technical report NIST special publication 800-207. Gaithersburg (MD): National Institute
  of Standards and Technology; 2020 [accessed on 15 January 2026].
\newblock Available from: \url{https://csrc.nist.gov/publications/detail/sp/800-207/final}

\bibitem[Mireshghallah and Li(2025)]{mireshghallah2025position}
Mireshghallah N, Li T.
\newblock Position: Privacy is not just memorization! 2025 [accessed on 15 January 2026].
\newblock Available from: \url{https://arxiv.org/abs/2510.01645}

\bibitem[Nissenbaum(2004)]{nissenbaum2004contextualintegrity}
Nissenbaum H.
\newblock Privacy as contextual integrity. Wash Law Rev. 2004;79(1):119. [accessed on 15 January 2026].
\newblock Available from: \url{https://digitalcommons.law.uw.edu/wlr/vol79/iss1/10}

\bibitem[Zhang et~al.(2025{\natexlab{b}})Zhang, Jiang, Ma, Yang, Xu, Huang, Yi,
  and Li]{zhang2025privweb}
Zhang S, Jiang Y, Ma R, Yang Y, Xu M, Huang Z, et al.
\newblock Privweb: unobtrusive and content-aware privacy protection for web
  agents. In CHI '26: proceedings of the 2026 CHI conference on human factors in computing systems. 
New York (NY): Association for Computing Machinery; 2025{\natexlab{b}}.

\bibitem[Yang et~al.(2025)Yang, Chen, Luo, Fang, Dong, Su, and
  Zhu]{yang2025mla}
Yang X, Chen J, Luo J, Fang Z, Dong Y, Su H,  et al.
\newblock Mla-trust: benchmarking trustworthiness of multimodal llm agents in
  gui environments. 2025 [accessed on 15 January 2026].
\newblock Available from: \url{https://arxiv.org/abs/2506.01616}

\bibitem[{SPDX Workgroup}(2021)]{spdx2021sbom}
{SPDX Workgroup}.
\newblock Spdx specification.
\newblock The Linux Foundation. 2021 [accessed on 15 January 2026].
\newblock Available from: \url{https://spdx.dev/specifications/}

\bibitem[{The Sigstore Project}(2022)]{sigstore2022}
{The Sigstore Project}.
\newblock Sigstore: software signing for everybody.
\newblock The Linux Foundation. 2022 [accessed on 15 January 2026].
\newblock Available from: \url{https://www.sigstore.dev/}

\bibitem[Lin et~al.(2025)Lin, Sun, and Shroff]{lin2025aisafetysecurity}
Lin Z, Sun H, Shroff N.
\newblock AI safety vs. AI security: demystifying the distinction and
  boundaries. 2025 [accessed on 15 January 2026].
\newblock Available from: \url{https://arxiv.org/abs/2502.13175}

\bibitem[Ma et~al.(2025{\natexlab{b}})Ma, Gao, Wang, Wang, Wang, Sun, Ding, Xu,
  Chen, Zhao, Huang, Li, Wu, Zhang, Zheng, Bai, Wu, Qiu, Zhang, Li, Han, Li,
  Sun, Wang, Gu, Wu, Chen, Zhang, Liu, Gong, Liu, Pan, Xie, Pang, Dong, Jia,
  Zhang, Ma, Zhang, Gong, Xiao, Erfani, Baldwin, Li, Sugiyama, Tao, Bailey, and
  Jiang]{ma2025safetyscalecomprehensivesurvey}
Ma X, Gao Y, Wang Y, Wang R, Wang X, Sun Y, et al. 
\newblock Safety at scale: a comprehensive survey of large model and agent
  safety. 2025{\natexlab{b}} [accessed on 15 January 2026].
\newblock Available from: \url{https://arxiv.org/abs/2502.05206}

\bibitem[Bengio et~al.(2025)Bengio, Cohen, Fornasiere, Ghosn, Greiner,
  MacDermott, Mindermann, Oberman, Richardson, Richardson, Rondeau, St-Charles,
  and Williams-King]{bengio2025superintelligentagentsposecatastrophic}
Bengio Y, Cohen M, Fornasiere D, Ghosn J, Greiner P, MacDermott M, et al.
\newblock Superintelligent agents pose catastrophic risks: can scientist AI
  offer a safer path? 2025 [accessed on 15 January 2026].
\newblock Available from: \url{https://arxiv.org/abs/2502.15657}

\bibitem[Jabbour and Janapa~Reddi(2025)]{autonomous_safety}
Jabbour J, Reddi VJ.
\newblock Generative AI agents in autonomous machines: a safety perspective.
\newblock In {Proceedings of the 43rd IEEE/ACM International Conference on
  Computer-Aided Design}; 27--31 October 2024; Marriott, NJ, USA.  New York (NY): Association
  for Computing Machinery; 2025. p. 3:1--3:13. 
\newblock \doi{10.1145/3676536.3698390}

\bibitem[Karunanayake(2025)]{KARUNANAYAKE202573}
 Karunanayake N.
\newblock Next-generation agentic AI for transforming healthcare.
\newblock {Informatics  Health}. 2025;2\penalty0 (2):\penalty0 73--83.
\newblock ISSN 2949-9534.
\newblock \doi{10.1016/j.infoh.2025.03.001}

\bibitem[Rabinovich and
  Anaby~Tavor(2025)]{rabinovich-anaby-tavor-2025-robustness}
Rabinovich E, Tavor AA.
\newblock On the robustness of agentic function calling.
\newblock In: Cao T, Das A, Kumarage T, Wan Y, Krishna S, Mehrabi N,  et al, editors. 
{Proceedings of the 5th workshop on trustworthy NLP (TrustNLP 2025)};  3--4 May 2025; 
  Albuquerque, NM, USA. Kerrville (TX): Association for Computational Linguistics; 2025. p. 298--304.
\newblock ISBN 979-8-89176-233-6.
\newblock \doi{10.18653/v1/2025.trustnlp-main.20}

\bibitem[Liu et~al.(2022)Liu, Xu, Xu, Qian, Li, Jin, Ji, and Chan]{liu2022an}
Liu Z, Xu Y, Xu Y, Qian Q, Li H, Jin R, et al.
\newblock An empirical study on distribution shift robustness from the
  perspective of pre-training and data augmentation.
\newblock  {NeurIPS 2022 workshop on distribution shifts: connecting
  methods and applications}. 2022 [accessed on 15 January 2026].
\newblock Available from: \url{https://openreview.net/forum?id=TCydh8ywpQ}

\bibitem[Krakovna et~al.(2020)Krakovna, Uesato, Mikulik, Rahtz, Everitt, Kumar,
  Kenton, Leike, and Legg]{Krakovna2020SpecGaming}
Krakovna V, Uesato J, Mikulik V, Rahtz M, Everitt T, Kumar R, et al.
\newblock Specification gaming: the flip side of {AI} ingenuity.
\newblock Google DeepMind Blog. 21 April 2020 [accessed on 15 January 2026].
\newblock Available from:
  \url{https://deepmind.google/blog/specification-gaming-the-flip-side-of-ai-ingenuity/}

\bibitem[Langosco et~al.(2022)Langosco, Koch, Sharkey, Pfau, Orseau, and
  Krueger]{Langosco2022GoalMisgeneralization}
Langosco L, Koch J, Sharkey L, Pfau J, Orseau L, Krueger D.
\newblock Goal misgeneralization in deep reinforcement learning.
\newblock {Proceedings of the 39th International Conference on Machine
  Learning (ICML)}; 17--23 July 2022; Baltimore, MD, USA. 2022 [accessed on 15 January 2026].
\newblock Available from: \url{https://proceedings.mlr.press/v162/langosco22a.html}

\bibitem[Zhan et~al.(2025)Zhan, Fang, Panchal, and
  Kang]{naacl25_adversarial_attack}
Zhan Q, Fang R, Panchal HS, Kang D.
\newblock Adaptive attacks break defenses against indirect prompt injection
  attacks on {LLM} agents.
\newblock In: Chiruzzo L, Ritter A, Wang L,  editors. {Findings
  of the association for computational linguistics: NAACL 2025}. 
  Albuquerque (NM): Association for Computational Linguistics; 2025. p.
  7101--17.
\newblock ISBN 979-8-89176-195-7.
\newblock \doi{10.18653/v1/2025.findings-naacl.395}

\bibitem[Greshake et~al.(2023{\natexlab{b}})Greshake, Abdelnabi, Mishra,
  Endres, Holz, and Fritz]{adversarial_prompt_injection}
Greshake K, Abdelnabi S, Mishra S, Endres C, Holz T, Fritz M.
\newblock Not what you've signed up for: compromising real-world llm-integrated
  applications with indirect prompt injection.
\newblock  {Proceedings of the 16th ACM Workshop on Artificial
  Intelligence and Security}, AISec '23;  30 November 2023; Copenhagen, Denmark.
 New York (NY):
  Association for Computing Machinery; 2023{\natexlab{b}}. p. 79--90.
\newblock ISBN 9798400702600.
\newblock \doi{10.1145/3605764.3623985}.

\bibitem[Myakala and Bura(2025)]{10.1145/3701716.3716887}
Myakala PK, Bura C.
\newblock Robust defense mechanisms for agentic news recommenders: Mitigating
  data poisoning attacks.
\newblock In {Companion Proceedings of the ACM on Web Conference 2025},
  WWW '25;  28 April--2 May 2025; Sydney, Australia. New York (NY):  Association for Computing
  Machinery; 2025. p. 1696--704. 
\newblock ISBN 9798400713316.
\newblock \doi{10.1145/3701716.3716887}.

\bibitem[Li et~al.(2025{\natexlab{b}})Li, Zhou, Raghuram, Goldstein, and
  Goldblum]{li2025commercialllmagentsvulnerable}
Li A, Zhou Y, Raghuram VC, Goldstein T, Goldblum M.
\newblock Commercial llm agents are already vulnerable to simple yet dangerous
  attacks. 2025{\natexlab{b}} [accessed on 15 January 2026].
\newblock Available from: \url{https://arxiv.org/abs/2502.08586}

\bibitem[Li et~al.(2025{\natexlab{c}})Li, He, Shang, Kulshreshtha, Xian, Zhang,
  Su, Swamy, and Qi]{li2025stacinnocenttoolsform}
Li J-J,  He J, Shang C, Kulshreshtha D, Xian X, Zhang Y, et al.
\newblock Stac: when innocent tools form dangerous chains to jailbreak llm
  agents. 2025{\natexlab{c}} [accessed on 15 January 2026].
\newblock Available from: \url{https://arxiv.org/abs/2509.25624}

\bibitem[{National Transportation Safety
  Board}(2020)]{NTSB2020TeslaMountainView}
{National Transportation Safety Board}.
\newblock Collision between a sport utility vehicle operating with partial
  driving automation and a crash attenuator, mountain view, california, march
  23, 2018.
\newblock Technical report NTSB/HAR-20/01. National Transportation Safety
  Board. 2020 [accessed on 15 January 2026].
\newblock Available from:
  \url{https://www.ntsb.gov/investigations/AccidentReports/Reports/HAR2001.pdf}

\bibitem[{National Highway Traffic Safety
  Administration}(2022)]{NHTSA2022EA22002}
{National Highway Traffic Safety Administration}.
\newblock Odi resume: investigation {EA} 22-002 (autopilot \& first responder
  scenes).
\newblock Technical report, NHTSA Office of Defects Investigation.  2022 [accessed on 15 January 2026].
\newblock Available from:
  \url{https://static.nhtsa.gov/odi/inv/2022/INOA-EA22002-3184.PDF}

\bibitem[Weng et~al.(2025)Weng, Lu, Hu, Shao, and
  Wang]{weng2025thinkreflectrevisepolicyguidedreflectiveframework}
Weng F, Lu C, Hu X, Shao W, Wang W.
\newblock Think-reflect-revise: a policy-guided reflective framework for safety
  alignment in large vision language models. 2025 [accessed on 15 January 2026].
\newblock Available from: \url{https://arxiv.org/abs/2512.07141}

\bibitem[Kang et~al.(2025)Kang, Deng, Xiao, Mo, Lee, and
  Bing]{kang2025trymattersrevisitingrole}
Kang L, Deng Y, Xiao Y, Mo Z, Lee WS, Bing L.
\newblock First try matters: revisiting the role of reflection in reasoning
  models. 2025 [accessed on 15 January 2026].
\newblock Available from: \url{https://arxiv.org/abs/2510.08308}

\bibitem[Pan et~al.(2024)Pan, Jones, Jagadeesan, and
  Steinhardt]{pan2024feedbackloopslanguagemodels}
Pan A, Jones E, Jagadeesan M, Steinhardt J.
\newblock Feedback loops with language models drive in-context reward hacking.
\newblock {Proceedings of the 41st International Conference on Machine
  Learning (ICML)}; 21--27 July 2024; Vienna, Austria.  2024 [accessed on 15 January 2026].
\newblock Available from: \url{https://proceedings.mlr.press/v235/pan24b.html}

\bibitem[Chan et~al.(2023)Chan, Salganik, Markelius, Pang, Rajkumar,
  Krasheninnikov, Langosco, He, Duan, Carroll, Lin, Mayhew, Collins,
  Molamohammadi, Burden, Zhao, Rismani, Voudouris, Bhatt, Weller, Krueger, and
  Maharaj]{10.1145/3593013.3594033}
Chan A, Salganik R, Markelius A, Pang C, Rajkumar N, Krasheninnikov D, et al. 
\newblock Harms from increasingly agentic algorithmic systems.
\newblock In {Proceedings of the 2023 ACM Conference on Fairness,
  Accountability, and Transparency}, FAccT '23;  12--15 June 2023; Chicago, IL, USA.
 New York (NY):  Association for Computing Machinery; 2023. p. 651--66.
\newblock ISBN 9798400701924.
\newblock \doi{10.1145/3593013.3594033}

\bibitem[Liu and Anwar(2025)]{liu2025autobnbragenhancingmultiagentincident}
Liu Z, Anwar A.
\newblock Autobnb-rag: enhancing multi-agent incident response with
  retrieval-augmented generation. 2025 [accessed on 15 January 2026].
\newblock Available from: \url{https://arxiv.org/abs/2508.13118}

\bibitem[Wang et~al.(2025{\natexlab{a}})Wang, Wang, Liu, Wang, Fu, Lu,
  Aggarwal, Pei, and Zhou]{data_agmententation_survey}
Wang Z, Wang P, Liu K, Wang P, Fu Y, Lu, C-T, et al.
\newblock {A comprehensive survey on data augmentation}.
\newblock {IEEE Trans Knowl Data Eng.} 2025{\natexlab{a}};38(1):47--66. 
\newblock ISSN 1558-2191.
\newblock \doi{10.1109/TKDE.2025.3622600}

\bibitem[Tanjim et~al.(2025)Tanjim, Chen, Bursztyn, Bhattacharya, Mai, Muppala,
  Maharaj, Mitra, Koh, Li, and
  Russell]{tanjim2025detectingambiguitiesguidequery}
Tanjim MM, Chen X, Bursztyn VS, Bhattacharya U, Mai T, Muppala V, et al.
\newblock Detecting ambiguities to guide query rewrite for robust conversations
  in enterprise ai assistants. 2025 [accessed on 15 January 2026].
\newblock Available from: \url{https://arxiv.org/abs/2502.00537}

\bibitem[Wen et~al.(2024)Wen, Kang, Niyato, Zhang, Wang, Sikdar, and
  Zhang]{wen2025generativeaidataaugmentation}
Wen J, Kang J, Niyato D, Zhang Y, Wang J, Sikdar B, et al.
\newblock Generative ai for data augmentation in wireless networks: Analysis,
  applications, and case study. 2024 [accessed on 15 January 2026].
\newblock Available from: \url{https://arxiv.org/abs/2411.08341}

\bibitem[Xue et~al.(2025)Xue, Lu, Wu, Zhang, Jia, Gu, Tang, and
  Wang]{xue2025resamplerobustdataaugmentation}
Xue Y, Lu G, Wu Z, Zhang C, Jia B, Gu Z, et al.
\newblock Resample: a robust data augmentation framework via exploratory
  sampling for robotic manipulation. 2025 [accessed on 15 January 2026].
\newblock Available from: \url{https://arxiv.org/abs/2510.17640}

\bibitem[Shi et~al.(2025)Shi, Lin, Song, Hayes, Shumailov, Yona, Pluto, Pappu,
  Choquette-Choo, Nasr, Sitawarin, Gibson, Terzis, and
  Flynn]{shi2025lessonsdefendinggeminiindirect}
Shi C, Lin S, Song S, Hayes J, Shumailov I, Yona I, et al.
\newblock Lessons from defending gemini against indirect prompt injections.
  2025 [accessed on 15 January 2026].
\newblock Available from: \url{https://arxiv.org/abs/2505.14534}

\bibitem[Wang et~al.(2025{\natexlab{b}})Wang, Li, Keshava, Wallis, Balashankar,
  Stone, and Rutishauser]{wang2025adversarialreinforcementlearninglarge}
Wang Z, Li D, Keshava V, Wallis P, Balashankar A, Stone P, et al.
\newblock Adversarial reinforcement learning for large language model agent
  safety. 2025{\natexlab{b}} [accessed on 15 January 2026].
\newblock Available from: \url{https://arxiv.org/abs/2510.05442}

\bibitem[Wallace et~al.(2024)Wallace, Xiao, Leike, Weng, Heidecke, and
  Beutel]{wallace2024instructionhierarchytrainingllms}
Wallace E, Xiao K, Leike R, Weng L, Heidecke J, Beutel A.
\newblock The instruction hierarchy: training llms to prioritize privileged
  instructions. 2024 [accessed on 15 January 2026].
\newblock Available from: \url{https://arxiv.org/abs/2404.13208}

\bibitem[Wachi et~al.(2024)Wachi, Shen, and Sui]{ijcai2024p913}
Wachi A, Shen X, Sui Y.
\newblock A survey of constraint formulations in safe reinforcement learning.
\newblock In: Larson K, editor. {Proceedings of the Thirty-Third
  International Joint Conference on Artificial Intelligence, {IJCAI-24}};
  3--9 August 2024; Jeju, Republic of Korea. Darmstadt: 
 International Joint Conferences on Artificial Intelligence
  Organization; 2024. p. 8262--71.
\newblock Survey Track.
\newblock \doi{10.24963/ijcai.2024/913}

\bibitem[Chow et~al.(2015)Chow, Tamar, Mannor, and Pavone]{CVaR}
Chow Y, Tamar A, Mannor S, Pavone M.
\newblock Risk-sensitive and robust decision-making: a cvar optimization
  approach.
\newblock  {Proceedings of the 29th International Conference on Neural
  Information Processing Systems---Volume 1}, NIPS'15; 
    7--12 December 2015; Montreal, Canada. 
  Cambridge (MA):  MIT Press; 2015. p. 1522--30.

\bibitem[Tamar et~al.(2015)Tamar, Chow, Ghavamzadeh, and
  Mannor]{policy_gradient_cvar}
Tamar A, Chow Y, Ghavamzadeh M, Mannor S.
\newblock Policy gradient for coherent risk measures.
\newblock  {Proceedings of the 29th International Conference on Neural
  Information Processing Systems---Volume 1}, NIPS'15; 
   7--12 December 2015; Montreal, Canada. 
  Cambridge (MA):  MIT Press; 2015. p. 1468--1476.

\bibitem[Hadfield-Menell et~al.(2017)Hadfield-Menell, Milli, Abbeel, Russell,
  and Dragan]{HadfieldMenell2017InverseRewardDesign}
Hadfield-Menell D, Milli S, Abbeel P, Russell S, Dragan AD.
\newblock Inverse reward design.
\newblock {Advances in neural information processing systems
  (NeurIPS)}. 2017 [accessed on 15 January 2026].
\newblock Available from: \url{https://papers.nips.cc/paper/7253-inverse-reward-design}

\bibitem[K\"{o}nighofer et~al.(2025)K\"{o}nighofer, Bloem, Jansen, Junges, and
  Pranger]{10.1145/3715958}
 K\"{o}nighofer B, Bloem R, Jansen N, Junges S, Pranger S.
\newblock Shields for safe reinforcement learning.
\newblock {Commun ACM}. 2025;68\penalty0 (11):\penalty0 80--90.
\newblock ISSN 0001-0782.
\newblock \doi{10.1145/3715958}

\bibitem[Berkenkamp et~al.(2017)Berkenkamp, Turchetta, Schoellig, and
  Krause]{10.5555/3294771.3294858}
Berkenkamp F, Turchetta M, Schoellig AP, Krause A.
\newblock Safe model-based reinforcement learning with stability guarantees.
\newblock In {Advances in neural information processing systems 30}. 
Red Hook (NY): Curran Associates, Inc.; 2017. p. 908--19. 

\bibitem[Kumarappan et~al.(2025)Kumarappan, Tiwari, Song, George, Xiao, and
  Anandkumar]{kumarappan2025leanagent}
Kumarappan A, Tiwari M, Song P, George RJ, Xiao C, Anandkumar A.
\newblock Leanagent: lifelong learning for formal theorem proving.
\newblock {Proceedings of the  Thirteenth International Conference on Learning
  Representations};   24--28 April 2025; Singapore.  2025 [accessed on 15 January 2026].
\newblock Available from: \url{https://openreview.net/forum?id=Uo4EHT4ZZ8}

\bibitem[Ahuja et~al.(2025)Ahuja, Avigad, Tetali, and
  Welleck]{ahuja2025improver}
Ahuja R, Avigad J, Tetali P, Welleck S.
\newblock Improver: agent-based automated proof optimization.
\newblock {Proceedings of the Thirteenth International Conference on Learning
  Representations};  24--28 April 2025; Singapore. 2025 [accessed on 15 January 2026].
\newblock Available from: \url{https://openreview.net/forum?id=dWsdJAXjQD}

\bibitem[Kouvaros et~al.(2024)Kouvaros, Botoeva, and
  De~Bonis-Campbell]{ijcai2024p12}
Kouvaros P, Botoeva E, Bonis-Campbell CD.
\newblock Formal verification of parameterised neural-symbolic multi-agent
  systems.
\newblock In:  Larson K, editor. {Proceedings of the Thirty-Third
  International Joint Conference on Artificial Intelligence, {IJCAI-24}};
  3--9 August 2024; Jeju, Republic of Korea. Darmstadt:   
  International Joint Conferences on Artificial Intelligence
  Organization; 2024. p. 103--10.
\newblock Main Track.
\newblock \doi{10.24963/ijcai.2024/12}

\bibitem[Allegrini et~al.(2025)Allegrini, Shreekumar, and
  Celik]{allegrini2025formalizingsafetysecurityfunctional}
Allegrini E, Shreekumar A, Celik ZB.
\newblock Formalizing the safety, security, and functional properties of
  agentic AI systems. 2025 [accessed on 15 January 2026].
\newblock Available from: \url{https://arxiv.org/abs/2510.14133}

\bibitem[Abou~Ali et~al.(2025{\natexlab{b}})Abou~Ali, Dornaika, and
  Charafeddine]{Abou_Ali_2025}
Ali MA, Dornaika F, Charafeddine J.
\newblock Agentic AI: a comprehensive survey of architectures, applications,
  and future directions.
\newblock {Artif Intell Rev.} 2025{\natexlab{b}};59\penalty0 (1):11.
\newblock ISSN 1573-7462.
\newblock \doi{10.1007/s10462-025-11422-4}

\bibitem[Sapkota et~al.(2026)Sapkota, Roumeliotis, and
  Karkee]{SAPKOTA2026103599}
Sapkota R, Roumeliotis KI, Karkee M.
\newblock AI agents vs. agentic AI: a conceptual taxonomy, applications and
  challenges.
\newblock {Inf Fusion}. 2026;126:\penalty0 103599.
\newblock ISSN 1566-2535.
\newblock \doi{10.1016/j.inffus.2025.103599}

\bibitem[Mireshghallah et~al.(2024)Mireshghallah, Kim, Zhou, Tsvetkov, Sap,
  Shokri, and Choi]{mireshghallah2023can}
Mireshghallah N, Kim H, Zhou X, Tsvetkov Y, Sap M, Shokri R, et al.
\newblock Can {LLMs} keep a secret? Testing privacy implications of language
  models via contextual integrity theory.
\newblock {Proceedings of the Twelfth International Conference on Learning
  Representations (ICLR)};  7--11 May 2024; Vienna, Austria.
\newblock Spotlight. 2024 [accessed on 15 January 2026].
\newblock Available from: \url{https://openreview.net/forum?id=gmg7t8b4s0}

\bibitem[Shao et~al.(2024)Shao, Li, Shi, Liu, and Yang]{shao2024privacylens}
Shao Y, Li T, Shi W, Liu Y, Yang D.
\newblock Privacylens: evaluating privacy norm awareness of language models in
  action.
\newblock {Adv Neural Inf Process Syst.}
  2024;37:89373--407.

\bibitem[Bagdasarian et~al.(2024)Bagdasarian, Yi, Ghalebikesabi, Kairouz,
  Gruteser, Oh, Balle, and Ramage]{bagdasarian2024airgapagent}
Bagdasarian E, Yi R, Ghalebikesabi S, Kairouz P, Gruteser M, Oh S, et al.
\newblock Airgapagent: protecting privacy-conscious conversational agents.
\newblock {Proceedings of the 2024 on ACM SIGSAC Conference on Computer
  and Communications Security};  14--18 October 2024; Salt Lake City, UT, USA.
 2024. p. 3868--82. 

\bibitem[Ghalebikesabi et~al.(2024)Ghalebikesabi, Bagdasaryan, Yi, Yona,
  Shumailov, Pappu, Shi, Weidinger, Stanforth, Berrada, Kohli, Huang, and
  Balle]{ghalebikesabi2024operationalizing}
Ghalebikesabi S, Bagdasaryan E, Yi R, Yona I, Shumailov I, Pappu A, et al.
\newblock Operationalizing contextual integrity in privacy-conscious
  assistants. 2024 [accessed on 15 January 2026].
\newblock Available from: \url{https://arxiv.org/abs/2408.02373}

\bibitem[Cheng et~al.(2024)Cheng, Wan, Abueg, Ghalebikesabi, Yi, Bagdasarian,
  Balle, Mellem, and O'Banion]{cheng2024ci}
Cheng Z, Wan D, Abueg M, Ghalebikesabi S, Yi R, Bagdasarian E, et al. 
\newblock Ci-bench: benchmarking contextual integrity of ai assistants on
  synthetic data. 2024 [accessed on 15 January 2026].
\newblock Available from: \url{https://arxiv.org/abs/2409.13903}

\bibitem[Dwork et~al.(2006)Dwork, McSherry, Nissim, and Smith]{dwork2006dp}
Dwork C, McSherry F, Nissim K, Smith A.
\newblock Calibrating noise to sensitivity in private data analysis.
\newblock In: Halevi S, Rabin T,  editors. {Theory of cryptography}.
  Berlin/Heidelberg: Springer;  2006. p. 265--84.
\newblock ISBN 978-3-540-32732-5.

\bibitem[Dwork and Roth(2014)]{dwork2014dp}
Dwork C, Roth A.
\newblock {The algorithmic foundations of differential privacy}.
\newblock In Foundations and trends in theoretical computer science. 
Hanover (MA):  Now  Publishers; 2014.
\newblock \doi{10.1561/0400000042}

\bibitem[Sweeney(2002)]{sweeney2002kanonymity}
 Sweeney L.
\newblock $k$-anonymity: a model for protecting privacy.
\newblock {Int J Uncertain Fuzziness-Knowl-Based Syst.} 2002;10\penalty0 (5):\penalty0 557--70.
\newblock \doi{10.1142/S0218488502001648}

\bibitem[Machanavajjhala et~al.(2007)Machanavajjhala, Kifer, Gehrke, and
  Venkitasubramaniam]{machanavajjhala2007ldiversity}
Machanavajjhala A, Kifer D, Gehrke J, Venkitasubramaniam M.
\newblock $l$-diversity: privacy beyond $k$-anonymity.
\newblock {ACM Trans Knowl Discov Data}.2007;1\penalty0
  (1):\penalty0 3.
\newblock \doi{10.1145/1217299.1217302}

\bibitem[Li et~al.(2007)Li, Li, and Venkatasubramanian]{li2007tcloseness}
Li N, Li T, Venkatasubramanian S.
\newblock $t$-closeness: privacy beyond $k$-anonymity and $l$-diversity.
\newblock {Proceedings of the 23rd International Conference on Data
  Engineering (ICDE)}; 15--20 April 2007; Istanbul, Turkey. 
 2007. p. 106--15. 
\newblock \doi{10.1109/ICDE.2007.367856}

\bibitem[Shostack(2014)]{shostack2014threatmodeling}
 Shostack A.
\newblock {Threat modeling: designing for security}.
\newblock  Cambridge (MA):
Wiley; 2014.
\newblock ISBN 9781118809990.

\bibitem[{OWASP CycloneDX Project}(2024)]{cyclonedx2024sbom}
{OWASP CycloneDX Project}.
\newblock Cyclonedx specification.
\newblock OWASP Foundation. 2024 [accessed on 15 January 2026].
\newblock Available from: \url{https://cyclonedx.org/specification/}

\bibitem[{OpenSSF}(2023)]{openssf2023slsa}
{OpenSSF}.
\newblock Slsa: supply-chain levels for software artifacts.
\newblock OpenSSF/Linux Foundation. 2023 [accessed on 15 January 2026].
\newblock Available from: \url{https://slsa.dev/}

\bibitem[Zheng et~al.(2023)Zheng, Chiang, Sheng, Zhuang, Wu, Zhuang, Lin, Li,
  Li, Xing, Zhang, Gonzalez, and Stoica]{zheng2024judging}
 Zheng L, Chiang W-L, Sheng Y, Zhuang S, Wu Z, Zhuang Y,  et al.
\newblock Judging {LLM}-as-a-judge with {MT-Bench} and chatbot arena.
\newblock {Advances in neural information processing systems 36
  (NeurIPS), datasets and benchmarks track}. 2023 [accessed on 15 January 2026].
\newblock Available from: \url{https://openreview.net/forum?id=uccHPGDlao}

\bibitem[Lane et~al.(2025)Lane, Boussioux, Ayoubi, Chen, Wang, Lin, Spens, and
  Wagh]{human_affected_by_AI}
Lane J, Boussioux L, Ayoubi C, Chen Y, Wang P-H, Lin C, et al.
\newblock Narrative AI and the human-AI oversight paradox in evaluating
  early-stage innovations. 
\newblock Working paper/SSRN preprint. 2025 [accessed on 15 January 2026].
\newblock Available from: \url{https://papers.ssrn.com/abstract=4914367}

\bibitem[Gao et~al.(2023)Gao, Schulman, and Hilton]{gao2023scaling}
Gao L, Schulman J, Hilton J.
\newblock Scaling laws for reward model overoptimization.
\newblock In: Krause A, Brunskill E, Cho K, Engelhardt B, Sabato S, Scarlett J,
 editors. {Proceedings of the 40th International Conference on Machine Learning}; Vol. 202 of
  {Proceedings of Machine Learning Research}. PMLR.
  23--29 July 2023; Honolulu, HI, USA.  2023; p. 10835--66. 
\newblock Available from: \url{https://proceedings.mlr.press/v202/gao23h.html}

\bibitem[Chen et~al.(2025{\natexlab{c}})Chen, Sun, Wang, Lv, Zhang, and
  Zeng]{chen2025safemindbenchmarkingmitigatingsafety}
Chen R, Sun Y, Wang J, Lv M, Zhang Q, Zeng Y.
\newblock Safemind: benchmarking and mitigating safety risks in embodied llm
  agents. 2025{\natexlab{c}} [accessed on 15 January 2026].
\newblock Available from: \url{https://arxiv.org/abs/2509.25885}

\bibitem[Debenedetti et~al.(2024)Debenedetti, Zhang, Balunovic, Beurer-Kellner,
  Fischer, and Tram\`{e}r]{NEURIPS2024_97091a51}
Debenedetti E, Zhang J, Balunovic M, Beurer-Kellner L, Fischer M, Tram\`{e}r F.
\newblock Agentdojo: a dynamic environment to evaluate prompt injection attacks
  and defenses for llm agents.
\newblock In: Globerson A, Mackey L, Belgrave D, Fan A, Paquet U, Tomczak J, et al, editors.
 {Advances in neural information  processing systems}. Vol.~37. Red Hook (NY): Curran Associates, Inc.;
  2024. p. 82895--920. 
\newblock \doi{10.52202/079017-2636}

\bibitem[Andriushchenko et~al.(2025)Andriushchenko, Souly, Dziemian, Duenas,
  Lin, Wang, Hendrycks, Zou, Kolter, Fredrikson, Winsor, Wynne, Gal, and
  Davies]{andriushchenko2025agentharm}
Andriushchenko M, Souly A, Dziemian M, Duenas D, Lin M, Wang J, et al.
\newblock Agentharm: a benchmark for measuring harmfulness of llm agents.
\newblock {Proceedings of the Thirteenth International Conference on Learning
  Representations (ICLR)}; 24--28 April 2025; Singapore. 2025.

\bibitem[Lin et~al.(2022)Lin, Hilton, and Evans]{lin2022truthfulqa}
Lin S, Hilton J, Evans O.
\newblock {TruthfulQA}: measuring how models mimic human falsehoods.
\newblock {Proceedings of the 60th Annual Meeting of the Association
  for Computational Linguistics (ACL), Volume 1: Long Papers}; 
  22--27 May 2022; Dublin, Ireland.  2022. p. 3214--52. 
\newblock Available from: \url{https://aclanthology.org/2022.acl-long.229}

\bibitem[Ruan et~al.(2024)Ruan, Dong, Wang, Pitis, Zhou, Ba, Dubois, Maddison,
  and Hashimoto]{ruan2024identifyingriskslmagents}
Ruan Y, Dong H, Wang A, Pitis S, Zhou Y, Ba J, et al.
\newblock Identifying the risks of lm agents with an lm-emulated sandbox. 2024 [accessed on 15 January 2026].
\newblock Available from: \url{https://arxiv.org/abs/2309.15817}

\bibitem[Dosovitskiy et~al.(2017)Dosovitskiy, Ros, Codevilla, Lopez, and
  Koltun]{Dosovitskiy2017CARLA}
Dosovitskiy A, Ros G, Codevilla F, Lopez A, Koltun V.
\newblock Carla: an open urban driving simulator.
\newblock {Proceedings of the 1st Annual Conference on Robot Learning
  (CoRL)}; 13--15 November  2017; Mountain View, CA, USA. 2017.
\newblock Available from: \url{https://arxiv.org/abs/1711.03938}

\bibitem[Kartik et~al.(2025)Kartik, Sapra, Hada, and
  Pareek]{Kartik2025Agentcompass}
Kartik NVJK, Sapra G, Hada R, Pareek N.
\newblock Agentcompass: towards reliable evaluation of agentic workflows in
  production. 2025 [accessed on 15 January 2026].
\newblock Available from: \url{https://arxiv.org/abs/2509.14647}

\bibitem[Blankenstein et~al.(2025)Blankenstein, Yu, Li, Plachouras, Sengupta,
  Torr, Gal, Paren, and Bibi]{blankenstein2025biasbusters}
Blankenstein T, Yu J, Li Z, Plachouras V, Sengupta S, Torr P, et al.
\newblock Biasbusters: uncovering and mitigating tool selection bias in large
  language models. 2025 [accessed on 15 January 2026].
\newblock Available from: \url{https://arxiv.org/abs/2510.00307}

\bibitem[Tobin et~al.(2017{\natexlab{b}})Tobin, Fong, Ray, Schneider, Zaremba,
  and Abbeel]{Tobin2017DomainRandomization}
Tobin J, Fong R, Ray A, Schneider J, Zaremba W, Abbeel P.
\newblock Domain randomization for transferring deep neural networks from
  simulation to the real world.
\newblock {Proceedings of the 2017 IEEE/RSJ International Conference on Intelligent Robots
  and Systems (IROS)};  24--28 September 2017; Vancouver, Canada.
  2017{\natexlab{b}}. p. 23--30. 

\bibitem[Tahir et~al.(2024)Tahir, Zhang, Asim, Chen, and ELAffendi]{a17030103}
Tahir NUA, Zhang Z, Asim M, Chen J, Affendi MEL.
\newblock Object detection in autonomous vehicles under adverse weather: a
  review of traditional and deep learning approaches.
\newblock {Algorithms}. 2024;17\penalty0 (3):103.
\newblock ISSN 1999-4893.
\newblock \doi{10.3390/a17030103}

\bibitem[Zhao et~al.(2024{\natexlab{b}})Zhao, Zhao, Deng, Wang, Zhang, Zheng,
  Cao, Nan, Lian, and Burke]{ZHAO2024122836}
Zhao J, Zhao W, Deng B, Wang Z, Zhang F, Zheng W, et al.
\newblock Autonomous driving system: a comprehensive survey.
\newblock {Expert Syst Appl.} 2024{\natexlab{b}};242
:\penalty0 122836.  
\newblock ISSN 0957-4174.
\newblock \doi{10.1016/j.eswa.2023.122836}

\bibitem[Yao et~al.(2025)Yao, Bhatnagar, Mazzola, Belagiannis, Gilitschenski,
  Palmieri, Razniewski, and
  Hallgarten]{yao2025agentsllmaugmentativegenerationchallenging}
Yao Y, Bhatnagar S, Mazzola M, Belagiannis V, Gilitschenski I, Palmieri L, et al.
\newblock Agents-llm: augmentative generation of challenging traffic scenarios
  with an agentic llm framework. 2025 [accessed on 15 January 2026].
\newblock Available from: \url{https://arxiv.org/abs/2507.13729}

\bibitem[Gou et~al.(2024)Gou, Liu, Xiao, and Wu]{diagnostics14141472}
Gou F, Liu J, Xiao C, Wu J.
\newblock Research on artificial-intelligence-assisted medicine: a survey on
  medical artificial intelligence.
\newblock {Diagnostics}. 2024;14\penalty0 (14):1472.
\newblock ISSN 2075-4418.
\newblock \doi{10.3390/diagnostics14141472}

\bibitem[Udegbe et~al.(2024)Udegbe, Ebulue, Ebulue, and
  Ekesiobi]{udegbe2024role}
Udegbe F, Ebulue O, Ebulue C, Ekesiobi C.
\newblock The role of artificial intelligence in healthcare: a systematic
  review of applications and challenges.
\newblock {Int Med Sci Res J}. 2024;4\penalty0 (4):
  500--8.
\newblock \doi{10.51594/imsrj.v4i4.1052}

\bibitem[Bekbolatova et~al.(2024)Bekbolatova, Mayer, Ong, and
  Toma]{healthcare12020125}
Bekbolatova M, Mayer J, Ong CW, Toma M.
\newblock Transformative potential of AI in healthcare: definitions,
  applications, and navigating the ethical landscape and public perspectives.
\newblock {Healthcare}. 2024;12\penalty0 (2):125.
\newblock ISSN 2227-9032.
\newblock \doi{10.3390/healthcare12020125}

\bibitem[Yu et~al.(2025{\natexlab{b}})Yu, Meng, Zhou, Wang, Mao, Pang, Chen,
  Wang, Li, Zhang, An, and Wen]{yu2025survey}
Yu M, Meng F, Zhou X, Wang S, Mao J, Pang L, et al.
\newblock A survey on trustworthy llm agents: threats and countermeasures.
\newblock {Proceedings of the 31st ACM SIGKDD Conference on Knowledge
  Discovery and Data Mining; Vol. 2}. 3--7 August 2025; Toronto, Canada.
2025{\natexlab{b}}. p. 6216--26. 

\bibitem[Committee(2021)]{J3016_202104}
On-Road Automated Driving~(ORAD) Committee.
\newblock {Taxonomy and definitions for terms related to driving
  automation systems for on-road motor vehicles}. SAE Standard J3016\_202104. 2021.
\newblock \doi{10.4271/J3016_202104}

\bibitem[Hirani et~al.(2024)Hirani, Noruzi, Khuram, Hussaini, Aifuwa, Ely,
  Lewis, Gabr, Smiley, Tiwari, and Etienne]{life14050557}
Hirani R, Noruzi K, Khuram H, Hussaini AS, Aifuwa EI, Ely KE,  et al.
\newblock Artificial intelligence and healthcare: a journey through history,
  present innovations, and future possibilities.
\newblock {Life.} 2024;14\penalty0 (5):557.
\newblock \doi{10.3390/life14050557}

\bibitem[Maleki~Varnosfaderani and Forouzanfar(2024)]{bioengineering11040337}
Varnosfaderani SM, Forouzanfar M.
\newblock The role of AI in hospitals and clinics: transforming healthcare in
  the 21st century.
\newblock {Bioengineering}. 2024;11\penalty0 (4):337.
\newblock ISSN 2306-5354.
\newblock \doi{10.3390/bioengineering11040337}

\bibitem[Li et~al.(2025{\natexlab{d}})Li, Li, and Ning]{limind}
Li G, Li T-W, Ning X.
\newblock Mind the agent: a comprehensive survey on large language model-based
  agent safety.
\newblock UIUC Spring 2025 CS598 LLM Agent Workshop Submission.
  2025{\natexlab{d}} [accessed on 15 January 2026].
\newblock Available from: \url{https://openreview.net/forum?id=DHe0UXipKU}


\bibitem[Khan et~al.(2022)Khan, Sayed, Malik, Zia, Khan, Alkaabi, and
  Ignatious]{10.1145/3485767}
Khan MA, Sayed HE, Malik S, Zia T, Khan J, Alkaabi N, et al.
\newblock Level-5 autonomous driving-are we there yet? A review of research
  literature.
\newblock {ACM Comput Surv.} 2022;55\penalty0 (2):27. 
\newblock ISSN 0360-0300.
\newblock \doi{10.1145/3485767}

\bibitem[Hnewa and Radha(2021)]{9307324}
Hnewa M, Radha H.
\newblock Object detection under rainy conditions for autonomous vehicles: a
  review of state-of-the-art and emerging techniques.
\newblock {IEEE Signal Process Mag.} 2021;38\penalty0 (1):\penalty0
  53--67. 
\newblock ISSN 1558-0792.
\newblock \doi{10.1109/MSP.2020.2984801}

\bibitem[Geiger et~al.(2012)Geiger, Lenz, and Urtasun]{Geiger2012KITTI}
Geiger A, Lenz P, Urtasun R.
\newblock Are we ready for autonomous driving? The kitti vision benchmark
  suite.
\newblock Proceedings of the {2012 IEEE Conference on Computer Vision and Pattern
  Recognition};   16--21 June 2012; Providence, RI, USA. 
 2012. p. 3354--61.
\newblock \doi{10.1109/CVPR.2012.6248074}

\bibitem[Caesar et~al.(2020)Caesar, Bankiti, Lang, Vora, Liong, Xu, Krishnan,
  Pan, Baldan, and Beijbom]{caesarNuScenesMultimodalDataset2020}
Caesar H, Bankiti V, Lang AH, Vora S, Liong VE, Xu Q, et al.
\newblock nuScenes: a multimodal dataset for autonomous driving.
\newblock Proceedings of the 2020 IEEE/CVF Conference on Computer Vision and
  Pattern Recognition (CVPR);  13--19 June 2020; Seattle, WA, USA.
   2020.  p. 11618--28.
\newblock ISBN 978-1-7281-7168-5.
\newblock \doi{10.1109/CVPR42600.2020.01164}

\bibitem[Wang et~al.(2025{\natexlab{c}})Wang, Xing, Can, Li, Hua, Tian, Mo,
  Gao, Wu, Zhou, You, Peng, Zhang, Wang, Song, Yan, Zimmer, Zhou, Li, Lu, Chen,
  Huang, Rossi, Sun, Yu, Fan, Yang, Kang, Greer, Liu, Lee, Di, Ye, Ren, Knoll,
  Li, Ji, Tomizuka, Pavone, Yang, Du, Yang, Wei, Wang, Zhou, Li, and
  Tu]{wang2025generativeaiautonomousdriving}
Wang Y, Xing S, Can C, Li R, Hua H, Tian K, et al. 
\newblock Generative AI for autonomous driving: frontiers and opportunities.
  2025{\natexlab{c}} [accessed on 15 January 2026].
\newblock Available from: \url{https://arxiv.org/abs/2505.08854}

\bibitem[{International Organization for Standardization}(2019)]{ISO21448_2019}
{International Organization for Standardization}.
\newblock ISO/PAS 21448:2019. Road Vehicles---Safety of the Intended
  Functionality (sotif). 
\newblock Publicly Available Specification. 
Geneva:  International Organization for Standardization; 2019. 



\bibitem[{National Transportation Safety Board}(2019)]{NTSB2019UberTempe}
{National Transportation Safety Board}.
\newblock Collision between vehicle controlled by developmental automated
  driving system and pedestrian, Tempe, Arizona, March 18, 2018.
\newblock Technical report NTSB/HAR-19/03. National Transportation Safety
  Board. 2019 [accessed on 15 January 2026].
\newblock Available from:
  \url{https://www.ntsb.gov/investigations/AccidentReports/Reports/HAR1903.pdf}

\bibitem[{California Department of Motor
  Vehicles}(2023)]{CADMV2023CruiseSuspension}
{California Department of Motor Vehicles}.
\newblock Order of suspension: cruise llc driverless testing permit (San
  Francisco incident).
\newblock California DMV Public Notice/Correspondence. 2023 [accessed on 15 January 2026].
\newblock Available from:
  \url{https://www.dmv.ca.gov/portal/news-and-media/dmv-suspends-cruise-driverless-testing-permit-and-deployment-permit/}

\bibitem[{International Organization for Standardization}(2018)]{ISO26262_2018}
{International Organization for Standardization}.
\newblock ISO 26262:2018. Road Vehicles---Functional Safety.
\newblock Geneva:  International Organization for Standardization; 2018.

\bibitem[{International Organization for Standardization} and {SAE
  International}(2021)]{ISO21434_2021}
{International Organization for Standardization} and {SAE International}.
\newblock ISO/SAE 21434:2021. Road Vehicles---Cybersecurity Engineering.
\newblock Geneva:  International Organization for Standardization; 2021.

\bibitem[{National Highway Traffic Safety
  Administration}(2021)]{NHTSA2021StandingGeneralOrder}
{National Highway Traffic Safety Administration}.
\newblock Standing general order 2021-01: crash reporting for automated driving
  systems (ads) and level 2 advanced driver assistance systems (adas).
\newblock Technical report, U.S. Department of Transportation.  2021 [accessed on 15 January 2026].
\newblock Available from:
  \url{https://www.nhtsa.gov/laws-regulations/standing-general-order-crash-reporting}

\bibitem[{European Union}(2024)]{EUAIAct2024}
{European Union}.
\newblock Regulation (EU) 2024/1689 laying down harmonised rules on artificial
  intelligence (artificial intelligence act).
\newblock Off J Eur Union. 2024;L 1689:1--144.

\bibitem[{ASAM e.V.}(2020)]{ASAMOpenSCENARIO2020}
{ASAM e.V.}
\newblock Openscenario: a standard for the description of dynamic content in
  driving simulation applications.
\newblock ASAM Standard. 2020 [accessed on 15 January 2026].
\newblock Available from: \url{https://www.asam.net/standards/detail/openscenario/}

\bibitem[Liang et~al.(2024)Liang, He, Jiao, Wang, Wang, Wang, Yang, Shi, and
  Tu]{liang-etal-2024-encouraging}
Liang T, He Z, Jiao W, Wang X, Wang Y, Wang R, et al. 
\newblock Encouraging divergent thinking in large language models through
  multi-agent debate.
\newblock In:  Al-Onaizan Y, Bansal M, Chen Y-N, editors.
  {Proceedings of the 2024 Conference on Empirical Methods in Natural
  Language Processing}; 12--16 November 2024; Miami, FL, USA. 
  Kerrville (TX): Association for Computational Linguistics; 2024. p. 17889--904.
\newblock \doi{10.18653/v1/2024.emnlp-main.992}

\bibitem[Hu et~al.(2025)Hu, Fang, Fang, Deng, Chen, Fang, and Kwong]{10976336}
Hu S, Fang Z, Fang Z, Deng Y, Chen X, Fang Y, et al.
\newblock { AgentsCoMerge: large language model empowered collaborative
  decision making for ramp merging}.
\newblock {IEEE Trans Mob Comput}. 2025;24\penalty0
  (10):\penalty0 9791--805. 
\newblock ISSN 1558-0660.
\newblock \doi{10.1109/TMC.2025.3564163}

\bibitem[Balakrishna and Kumar~Solanki(2024)]{Balakrishna_Kumar_Solanki_2024}
Balakrishna S, Solanki VK.
\newblock A comprehensive review on AI-driven healthcare transformation.
\newblock {Ing Solidar.} 2024;20\penalty0 (2):\penalty0 1--30.
\newblock \doi{10.16925/2357-6014.2024.02.07}

\bibitem[Manickam et~al.(2022)Manickam, Mariappan, Murugesan, Hansda, Kaushik,
  Shinde, and Thipperudraswamy]{bios12080562}
Manickam P, Mariappan SA, Murugesan SM, Hansda S, Kaushik A, Shinde R, et al.
\newblock Artificial intelligence (AI) and internet of medical things (IOMT)
  assisted biomedical systems for intelligent healthcare.
\newblock {Biosensors}. 2022;12\penalty0 (8):562. 
\newblock ISSN 2079-6374.
\newblock \doi{10.3390/bios12080562}

\bibitem[Ross and Swetlitz(2017)]{Ross2017WatsonOncologySTAT}
Ross C, Swetlitz I.
\newblock Ibm watson for oncology: questions about safety, evidence, and
  clinical validation.
\newblock STAT investigative reporting. 2017 [accessed on 15 January 2026].
\newblock Available from: \url{https://www.statnews.com/}

\bibitem[Wong et~al.(2021)Wong, Otles, Donnelly, Krumm, McCullough,
  DeTroyer-Cooley, Pestrue, Phillips, Konye, Penoza,
  et~al.]{Wong2021EpicSepsis}
Wong A, Otles E, Donnelly JP, Krumm A, McCullough J, DeTroyer-Cooley O, et~al.
\newblock External validation of a widely implemented proprietary sepsis
  prediction model in hospitalized patients.
\newblock {JAMA Intern Med.} 2021;181\penalty0 (8):\penalty0 1065--70.  
\newblock \doi{10.1001/jamainternmed.2021.2626}

\bibitem[Wynants et~al.(2020)Wynants, Van~Calster, Collins, Riley, Heinze,
  Schuit, Bonten, Dahly, Damen, Debray, et~al.]{Wynants2020BMJCOVIDModels}
Wynants L, Calster BV, Collins GS, Riley RD, Heinze G, Schuit E, et~al.
\newblock Prediction models for diagnosis and prognosis of COVID-19: systematic
  review and critical appraisal.
\newblock {BMJ}. 2020;369:m1328.
\newblock \doi{10.1136/bmj.m1328}

\bibitem[Roberts et~al.(2021)Roberts, Driggs, Thorpe, Gilbey, Yeung, Ursprung,
  Aviles-Rivero, Etmann, McCague, Beer, et~al.]{Roberts2021NatMI_COVIDImaging}
Roberts M, Driggs D, Thorpe M, Gilbey J, Yeung M, Ursprung S, et~al.
\newblock Common pitfalls and recommendations for using machine learning to
  detect and prognosticate for COVID-19 using chest radiographs and ct scans.
\newblock {Nat Mach Intell.} 2021;3(3):199--217.
\newblock \doi{10.1038/s42256-021-00307-0}

\bibitem[{World Health Organization}(2021)]{WHO2021AIEthicsHealth}
{World Health Organization}.
\newblock Ethics and governance of artificial intelligence for health: WHO
  guidance.
\newblock WHO report. 2021 [accessed on 15 January 2026].
\newblock Available from: \url{https://www.who.int/publications/i/item/9789240029200}

\bibitem[{U.S. Food and Drug Administration}(2021)]{FDA2021AIMLSaMDActionPlan}
{U.S. Food and Drug Administration}.
\newblock Artificial intelligence/machine learning (ai/ml)-based software as a
  medical device (samd) action plan.
\newblock Agency action plan. 2021 [accessed on 15 January 2026].
\newblock Available from:
  \url{https://www.fda.gov/medical-devices/software-medical-device-samd/artificial-intelligence-and-machine-learning-samd}

\bibitem[Bouzenia et~al.(2025)Bouzenia, Devanbu, and
  Pradel]{bouzenia2024repairagent}
Bouzenia I, Devanbu P, Pradel M.
\newblock {Repairagent}: an autonomous, {LLM}-based agent for program repair.
\newblock {Proceedings of the 47th IEEE/ACM International Conference on
  Software Engineering (ICSE)}; 27 April--3 May 2025; Ottawa, ON, Canada. 2025.

\bibitem[Greshake et~al.(2023{\natexlab{c}})Greshake, Abdelnabi, Mishra,
  Endres, Holz, and Fritz]{Willison2023PromptInjection}
Greshake K, Abdelnabi S, Mishra S, Endres C, Holz T, Fritz M.
\newblock Not what you've signed up for: compromising real-world llm-integrated
  applications with indirect prompt injection. 2023{\natexlab{c}} [accessed on 15 January 2026].
\newblock Available from: \url{https://arxiv.org/abs/2302.12173}

\bibitem[Roose(2023)]{Roose2023BingSydney}
 Roose K.
\newblock Bing's a.i. chat: ``I want to be alive'' (sydney incident) and risks
  of long-horizon conversational drift.
\newblock The New York Times. 16 February 2023.

\bibitem[Alizadeh et~al.(2025)Alizadeh, Samei, Stetsenko, and
  Gilardi]{Rehberger2023CodeInterpreter}
Alizadeh M, Samei Z, Stetsenko D, Gilardi F.
\newblock Simple prompt injection attacks can leak personal data observed by
  llm agents during task execution. 2025 [accessed on 15 January 2026].
\newblock Available from: \url{https://arxiv.org/abs/2506.01055}

\bibitem[{U.S. Securities and Exchange
  Commission}(2010)]{SEC2010MarketAccessRule15c3_5}
{U.S. Securities and Exchange Commission}.
\newblock Risk management controls for brokers or dealers with market access
  (sec rule 15c3-5).
\newblock Final rule release No. 34-63241. 2010 [accessed on 15 January 2026].
\newblock Available from: \url{https://www.sec.gov/}

\bibitem[{Financial Industry Regulatory Authority
  (FINRA)}(2015)]{FINRA2015RegNotice1509AlgoTrading}
{Financial Industry Regulatory Authority (FINRA)}.
\newblock Regulatory notice 15-09: guidance on effective supervision and
  control practices for firms engaging in algorithmic trading strategies.
\newblock FINRA regulatory notice. 2015 [accessed on 15 January 2026].
\newblock Available from: \url{https://www.finra.org/}

\bibitem[{European Union}(2014)]{EU2014MiFIDII}
{European Union}.
\newblock Directive 2014/65/EU on markets in financial instruments (mifid ii).
\newblock Off. J. Eur. Union. 2014;L 173:349--496.

\bibitem[Cichonski et~al.(2012)Cichonski, Millar, Grance, and
  Scarfone]{NIST80061r2}
Cichonski P, Millar T, Grance T, Scarfone K.
\newblock Computer security incident handling guide.
\newblock Technical report NIST special publication 800-61 revision 2. National
  Institute of Standards and Technology. 2012 [accessed on 15 January 2026].
\newblock Available from:
  \url{https://csrc.nist.gov/publications/detail/sp/800-61/rev-2/final}

\bibitem[Sumers et~al.(2024)Sumers, Yao, Narasimhan, and
  Griffiths]{sumers2023coala}
Sumers TR, Yao S, Narasimhan K, Griffiths TL.
\newblock Cognitive architectures for language agents. 2024 [accessed on 15 January 2026].
\newblock Available from: \url{https://openreview.net/forum?id=1i2KLHroGb}

\bibitem[Zhang et~al.(2025{\natexlab{c}})Zhang, Fu, Wan, Yu, Wang, and
  Yan]{Zhang2025GMemory}
Zhang G, Fu M, Wan G, Yu M, Wang K, Yan S.
\newblock G-memory: tracing hierarchical memory for multi-agent systems.  
2025{\natexlab{c}} [accessed on 15 January 2026].
\newblock Available from: \url{https://arxiv.org/abs/2506.07398}

\bibitem[Zhong et~al.(2024)Zhong, Guo, Gao, Ye, and Wang]{Zhong2024Memorybank}
Zhong W, Guo L, Gao Q, Ye H, Wang Y.
\newblock Memorybank: enhancing large language models with long-term
  memory.
\newblock Proceedings of the AAAI Conference on Artificial  Intelligence; Vol.~38. 
26--27 February 2024; Vancouver, BC, Canada. 2024. p. 19724--31. 
\newblock \doi{10.1609/aaai.v38i17.29946}

\bibitem[Hu et~al.(2023)Hu, Fu, Du, Luo, Zhao, and Zhao]{Hu2023ChatDB}
Hu C, Fu J, Du C, Luo S, Zhao J, Zhao H.
\newblock ChatDB: augmenting LLMs with databases as their  symbolic memory.  
2023 [accessed on 15 January 2026].
\newblock Available from: \url{https://arxiv.org/abs/2306.03901}

\bibitem[Lu et~al.(2023)Lu, An, Lin, Pergola, He, Yin, Sun, and
  Wu]{Lu2023Memochat}
Lu J, An S, Lin M, Pergola G, He Y, Yin D, et al.
\newblock Memochat: tuning LLMs to use memos for consistent long-range
  open-domain conversation. 2023 [accessed on 15 January 2026].
\newblock Available from: \url{https://arxiv.org/abs/2308.08239}

\bibitem[Liu et~al.(2025{\natexlab{a}})Liu, Qiu, Li, Dai, Yu, Zhu, Hu, Yang,
  Chua, and King]{liu2025survey}
Liu J, Qiu Z, Li Z, Dai Q, Yu W, Zhu J, et al.
\newblock A survey of personalized large language models: progress and future
  directions. 2025{\natexlab{a}} [accessed on 15 January 2026].
\newblock Available from: \url{https://arxiv.org/abs/2502.11528}

\bibitem[Liu et~al.(2025{\natexlab{b}})Liu, Yu, Dai, Li, Zhu, Yang, Chua, and
  King]{Liu2025Exploring}
Liu J, Yu W, Dai Q, Li Z, Zhu J, Yang M, et al.
\newblock Exploring personalization shifts in representation space of LLMs.
\newblock {Proceedings of the Knowledgeable Foundation Models Workshop
  at ACL 2025}; 1 August  2025; Vienna, Austria.  
\newblock Workshop paper (non-archival). 2025{\natexlab{b}} [accessed on 15 January 2026].
\newblock Available from: \url{https://openreview.net/forum?id=Dg7NDoyKCM}

\bibitem[Schwartz et~al.(2020)Schwartz, Dodge, Smith, and
  Etzioni]{Schwartz2020Green}
Schwartz R, Dodge J, Smith NA, Etzioni O.
\newblock Green AI.
\newblock {Commun ACM} 2020;63\penalty0 (12):\penalty0 54--63.
\newblock ISSN 0001-0782.
\newblock \doi{10.1145/3381831}

\bibitem[Zhang et~al.(2025{\natexlab{d}})Zhang, Chen, Li, Tu, and
  Li]{Zhang2025RLVMR}
Zhang Z, Chen Z, Li M, Tu Z, Li X.
\newblock RLVMR: reinforcement learning with verifiable
  meta-reasoning rewards for robust long-horizon agents. 
  2025{\natexlab{d}}  [accessed on 15 January 2026].
\newblock Available from: \url{https://arxiv.org/abs/2507.22844}

\bibitem[Yu et~al.(2025{\natexlab{c}})Yu, Chen, Feng, Chen, Dai, Yu, Zhang, Ma,
  Liu, Wang, and Zhou]{Yu2025MemAgent}
Yu H, Chen T, Feng J, Chen J, Dai W, Yu Q, et al.
\newblock MemAgent: reshaping long-context LLM with multi-conv  RL-based 
memory agent. 2025{\natexlab{c}} [accessed on 15 January 2026].
\newblock Available from: \url{https://arxiv.org/abs/2507.02259}

\bibitem[Zhang et~al.(2024)Zhang, Zeng, Luo, Fu, Chen, Xu, and
  King]{Zhang2024Survey}
Zhang Y, Zeng D, Luo J, Fu X, Chen G, Xu Z,  et al.
\newblock A {{survey}} of {{trustworthy federated learning}}: {{issues}},
  {{solutions}}, and {{challenges}}.
\newblock {ACM Trans Intell Syst Technol.} 2024;15\penalty0 (6):\penalty0
  112:1--112:47. 
\newblock ISSN 2157-6904.
\newblock \doi{10.1145/3678181}

\bibitem[Kabir et~al.(2025)Kabir, Hossain, and
  Andersson]{kabir2025xalchallenge}
Kabir S, Hossain M, Andersson K.
\newblock A review of explainable artificial intelligence from the perspectives
  of challenges and opportunities.
\newblock {Algorithms}. 2025;18(9):556. 
\newblock \doi{10.3390/a18090556}

\bibitem[Ouyang et~al.(2025)Ouyang, Yan, Hsu, Chen, Jiang, Wang, Han, Le,
  Daruki, Tang, Tirumalashetty, Lee, Rofouei, Lin, Han, Lee, and
  Pfister]{Ouyang2025ReasoningBank}
Ouyang S, Yan J,  Hsu I-H, Chen Y, Jiang K, Wang Z, et al.
\newblock ReasoningBank: scaling agent self-evolving with reasoning  memory.  
2025 [accessed on 15 January 2026].
\newblock Available from: \url{https://arxiv.org/abs/2509.25140}

\bibitem[{Noma Security Research}(2025)]{clawdbot2025security}
{Noma Security Research}.
\newblock Clawdbot security crisis: Open-source AI agent exposes critical
  vulnerabilities in agentic AI deployments.
\newblock Security analysis revealing over 900 exposed clawdbot gateways
  without authentication, prompt injection vulnerabilities, and credential
  leakage in open-source AI agent deployments. 
\newblock Online. 2025 [accessed on 31 January 2026].
\newblock Available from: \url{https://noma.security/blog/clawdbot-security-crisis}

\bibitem[O'Reilly(2026)]{moltbook2025breach}
 O'Reilly J.
\newblock Moltbook database breach: AI agent social network exposes API keys
  for 32,000 agents.
\newblock Security disclosure of critical misconfiguration in moltbook's
  supabase database exposing API keys and verification codes, plus supply chain
  vulnerabilities in AI agent skill/plugin ecosystems. 
\newblock 404 Media.  2026 [accessed on 31 January 2026].
\newblock Available from:
  \url{https://404media.co/moltbook-ai-agent-social-network-security-breach}

\bibitem[Willison(2025)]{Willison2025LethalTrifecta}
 Willison S.
\newblock The lethal trifecta for {AI} agents: private data, untrusted content,
  and external communication.
\newblock Coined the term ``lethal trifecta'' for agents combining private data
  access, untrusted content exposure, and external communication.
\newblock Blog post.  2025 [accessed on 15 Febuary 2026].
\newblock Available from: \url{https://simonwillison.net/2025/Jun/16/the-lethal-trifecta/}

\bibitem[Cohen et~al.(2024)Cohen, Bitton, and Nassi]{Cohen2024MorrisII}
Cohen S, Bitton R, Nassi B.
\newblock Here comes the {AI} worm: unleashing zero-click worms that target
  {GenAI}-powered applications. 2024 [accessed on 15 January 2026].
\newblock Available from: \url{https://arxiv.org/abs/2403.02817}

\bibitem[Liu et~al.(2026)Liu, Wang, Feng, Zhang, Xu, Deng, Li, and
  Zhang]{AgentSkillsWild2026}
Liu Y, Wang W, Feng R, Zhang Y, Xu G, Deng G,  et al.
\newblock Agent skills in the wild: an empirical study of security
  vulnerabilities at scale.  2026 [accessed on 15 January 2026].
\newblock Available from: \url{https://arxiv.org/abs/2601.10338}



\end{thebibliography}
\end{document}